\documentclass[journal]{IEEEtran}
\ifCLASSINFOpdf
  \usepackage[pdftex]{graphicx}
  \DeclareGraphicsExtensions{.pdf,.jpeg,.png}
\else
\fi
\usepackage{amsmath, amsthm}
\usepackage{algpseudocode}
\usepackage[ruled,vlined,linesnumbered]{algorithm2e}

\SetKwInput{KwIn}{Input}
\SetKwInput{KwOut}{Output}
\SetKw{KwRet}{return}
\SetKwInput{KwInit}{Initialize}
\SetKwInput{KwPre}{Precompute}
\SetKwComment{tcp}{\footnotesize\itshape// }{}

\newcommand{\Enc}{\textsc{Encode}}

\usepackage{booktabs,tabularx,seqsplit}
\usepackage{makecell}

\newcolumntype{Y}{>{\raggedright\arraybackslash}X}

\usepackage{array}

\usepackage{booktabs,siunitx,multirow,threeparttable}
\usepackage{dblfloatfix}   
\usepackage{placeins} 
\newcommand{\std}[1]{\num[exponent-mode=scientific]{#1}}

\newcommand{\tauany}{\tau_{\mathrm{ANY}}}

\usepackage{subcaption}
\usepackage{comment}

\usepackage{url}

\usepackage[colorlinks=true, allcolors=blue]{hyperref}

\usepackage{amsfonts}
\usepackage{amssymb}
\usepackage{array}

\usepackage[utf8]{inputenc}

\makeatletter
\newcommand{\GenerateLoFWithoutPrinting}{%
  \begingroup
  \let\clearpage\relax
  \let\newpage\relax
  \let\pagebreak\relax
  \setbox0=\vbox{\listoffigures}%
  \endgroup
}
\makeatother

\begin{document}

\GenerateLoFWithoutPrinting

%
\title{Generating High-quality Privacy-preserving Synthetic Data}
%
%
%
\author{
David Yavo,
Richard Khoury,
Christophe Pere,
and Sadoune Ait~Kaci~Azzou%
\thanks{David Yavo, Richard Khoury, and Christophe Pere are with the
Department of Computer Science and Software Engineering,
Université Laval, Québec City, QC, Canada
(e-mail: david.yavo.1@ulaval.ca; richard.khoury@ift.ulaval.ca;
christophe.pere.1@ulaval.ca).}%
\thanks{Sadoune Ait~Kaci~Azzou is with Caisses Desjardins du Québec,
Lévis, QC, Canada
(e-mail: sadoune.aitkaciazzou@gmail.com).}%
\thanks{David Yavo is the corresponding author.}%
\thanks{David Yavo, Richard Khoury, Christophe Pere, and Sadoune Ait~Kaci~Azzou
contributed equally to this work.}
}

\maketitle


\begin{abstract}
Synthetic tabular data can enable sharing and analysis of sensitive records, but it is difficult to deploy in practice because data curators must balance distributional fidelity, downstream utility, and privacy protection. We study a simple, model-agnostic post-processing framework that can be applied on top of any synthetic data generator to improve this trade-off. First, a \emph{mode-patching} step repairs categories that are missing or severely underrepresented in the synthetic data, while largely preserving other learned dependencies. Second, a $k$-nearest-neighbor filter replaces synthetic records that lie too close to real data points, enforcing a minimum distance from each real record to any synthetic one. We instantiate this framework for two neural generative models for tabular data---a feed-forward generator and a variational autoencoder---and evaluate it on three public datasets (credit card transactions, cardiovascular health, and census-based income). We measure marginal and joint distributional similarity, the performance of models trained on synthetic data and tested on real data, and several empirical privacy indicators including nearest-neighbor distances and attribute-inference attacks. With moderate thresholds ($\approx 0.2$--$0.35$), our post-processing reduces a standard divergence between real and synthetic category distributions by up to $36\%$ and improves a combined measure of pairwise dependence preservation by $10$--$14\%$, while keeping downstream predictive performance within about $\pm 1\%$ of the unprocessed baseline. At the same time, distance-based privacy indicators improve and the success of attribute-inference attacks is largely unchanged. Our results provide practical guidance for choosing thresholds and applying post-hoc repairs to enhance the quality and empirical privacy of synthetic tabular data, while complementing approaches that offer formal differential privacy guarantees.
\end{abstract}


%
\IEEEpeerreviewmaketitle
%
%
%
%

\section{Introduction}

\IEEEPARstart{R}{eliable} access to tabular data is essential for training and validating machine‑learning systems, yet legal and organizational constraints often preclude sharing real records \cite{kotal2022privetab}. Synthetic tabular data offers a viable alternative when it (i) resembles the real distribution; (ii) supports downstream analysis comparably to the original; and (iii) respects privacy \cite{liu2024scaling} \cite{yang2024structured}. Early systems (e.g., Bayesian‑network and copula‑based synthesizers) made transparent assumptions but struggled with complex mixed‑type dependencies \cite{elidan2010copula} \cite{feldman2024nonparametric}. Deep generative models—GAN/VAE variants such as CTGAN and TVAE—now dominate, yet they can suffer from mode collapse (notably on rare categories) and may emit samples dangerously close to training records \cite{yadav2024rigorous}. Meanwhile, diffusion models and row-serialization approaches using large language models are emerging as promising alternatives \cite{li2025diffusion} \cite{fang2024large} \cite{barr2025generative}. At the same time, mechanisms with hard privacy guarantees (e.g., DP‑SGD, PATE) often incur nontrivial fidelity/utility costs \cite{adams2025fidelity}\cite{hernandez2025comprehensive}. Evaluation practices increasingly emphasize fidelity–utility–privacy triads, but many studies still rely on narrow metric sets that underweight multivariate structure and adversarial risk \cite{hernandez2025comprehensive}.

Against this backdrop, we study \emph{post‑processing} controls that apply to any trained tabular generator and target two persistent pain points: categorical mode collapse and proximity‑based disclosure risk. Our goal is not to propose a new generator, but to introduce a model‑agnostic pipeline that can be slotted after CTGAN, TVAE, or future models and assessed under a unified, multi‑criteria protocol.

\textbf{Contributions.} We formalize a pipeline with three complementary evaluation layers—\emph{fidelity} (univariate and multivariate resemblance), \emph{utility} (TSTR across a diversified classifier suite), and \emph{privacy} (distance‑based metrics and adversarial proxies)—and introduce two architecture‑orthogonal components:
\begin{itemize}
  \item \textbf{Layer‑frozen mode‑patching.} We detect dropped categorical modes by cross‑tabulating real vs.\ synthetic frequency tables and flagging any real category with zero synthetic count. We then \emph{freeze} the feature‑extracting (lower) layers of the trained generator and fine‑tune only upper layers on rows carrying the missing category, producing a temporary model whose samples for that category are merged into the synthetic pool. The base generator remains unchanged; only the sample set is updated. Empirically, freezing roughly 60–80\% of early layers stabilizes adaptation while avoiding catastrophic forgetting (Algorithm~2; Figs.~3–4). This targeted repair restores categorical support without retraining from scratch or degrading previously covered modes.
  \item \textbf{HEOM–kNN \(\boldsymbol{\varepsilon_{\text{ANY}}}\) privacy filter.} We embed mixed‑type rows using the Heterogeneous Euclidean Overlap Metric (HEOM)—min–max scaling for numerics; one‑hot for categoricals scaled by \(1/\sqrt{2}\)—compute each real record’s 2‑NN radius \(r_i\), and deem a synthetic row unsafe if it falls inside any real radius. The \emph{ANY‑risk} is the fraction of unsafe rows. A rejection‑with‑replacement loop resamples the worst points until the empirical risk \(\widehat{\varepsilon}_{\text{ANY}}\) falls below a user‑set threshold \(\tau_{\text{ANY}}\) (Algorithms~3–4). This is a post‑hoc, distance‑based heuristic; it is \emph{not} a formal \((\varepsilon,\delta)\)‑DP mechanism and carries no DP accounting.
\end{itemize}

\textbf{Evaluation design.} We study three mixed‑type, real‑world benchmarks—\emph{Credit Default}, \emph{Adult}, and \emph{Cardio}—each with numeric and categorical variables and a binary outcome. For each dataset we train CTGAN and TVAE, then apply mode‑patching and the \(\varepsilon_{\text{ANY}}\) filter as post‑processing. We evaluate the results separately for fidelity, utility and privacy. 

\emph{Fidelity.} Univariate resemblance is measured via Jensen–Shannon divergence for categorical marginals and quantile‑based discrepancies (and effect sizes) for continuous variables. Multivariate structure is assessed with Pearson’s \(r\) (numeric–numeric), Cramér’s \(V\) (categorical–categorical), and the correlation ratio \(\eta^2\) (mixed), summarized by Frobenius norms and rank‑order correlations of the resulting dependence matrices. We also report joint‑distribution coverage and support recovery to track combinatorial patterns in high‑cardinality categoricals.

\emph{Utility.} Following the TSTR protocol, we train a diversified suite of eight classifiers on synthetic data and evaluate on real data, comparing against TRTR baselines. We report accuracy, AUC, F1, and related metrics averaged across models.

\emph{Privacy.} We compile complementary indicators: Distance to Closest Record (DCR) distributions, \emph{RPR} (fraction of synthetic records closer to real than to synthetic neighbors), Correct Attribution Probability (CAP), and attribute‑inference attacks (AIA; attacker accuracy or \(R^2\) as leakage proxies). These proxies are informative but not guarantees.

\textbf{Summary of findings.} Across datasets and models, \emph{moderate} filtering thresholds (\(\tau_{\text{ANY}}\approx 0.2\text{–}0.3\)) often strike the best balance: categorical divergence drops substantially on several tasks (e.g., 20–50\% Jensen–Shannon reductions) with multivariate structure preserved or modestly improved, while TSTR utility remains within \(\sim 1\%\) of the unfiltered baseline. In some cases, mild filtering acts as a regularizer that slightly improves generalization. By contrast, overly tight thresholds (\(\tau_{\text{ANY}}\!\ll\!0.1\)) can inflate tails or induce missing-pattern artifacts that hurt resemblance. Taken together, the results indicate a practical “sweet spot” where categorical support and numeric alignment improve without sacrificing predictive performance.

By decoupling fidelity restoration and proximity control from the core generator, our proposed post‑processing can complement current tabular generative methods. We release an open‑source evaluation suite that integrates our metrics and filtering to support reproducible, multi‑criteria benchmarking of synthetic tabular data generators.\footnote{\url{https://github.com/dayav/pipeline_synthetic_tabular_data}}

\section{Related Work}
\label{sec:background}
\label{RelatedWorks}
\label{Definitions}

\subsection{Generative Models for Tabular Data and Synthetic Data Generation}
\label{sec:generative_models}

Two major families of synthetic data generators dominate the field: (i) \emph{traditional/statistical synthesizers} such as Bayesian networks, copula-based models, and column-wise regressors (e.g., those implemented in \texttt{synthpop}), and (ii) \emph{deep generative models} (GANs/VAEs and successors) \cite{eigenschink2023deep} \cite{stoian2025survey}. 
Oversampling methods (SMOTE and variants) are often used to balance minority classes, but they do not model the full joint distribution \cite{engelmann2021conditional} \cite{bellinger2018manifold}\cite{kachan2025simplicial}. 
Deep models---particularly CTGAN and TVAE---adapt image-era generators to mixed continuous/categorical tables.


Generative Adversarial Networks (GANs) \cite{goodfellow2014generative} train two networks in a minimax game. The discriminator $D$ learns to distinguish real vs. fake data, while the generator $G$ tries to fool $D$. The standard objective is:

\begin{equation}
  \min_G \max_D \; 
  \mathbb{E}_{x \sim p_{\text{data}}} [\log D(x)] 
  + 
  \mathbb{E}_{z \sim p_z} [\log(1 - D(G(z)))].
\end{equation} \cite{goodfellow2014generative}

where $D(x)$ is the probability assigned to $x$ being real data, and $G(z)$ is the synthetic sample generated from noise $z$. 
CTGAN tailors this to tabular data with conditional sampling and training-by-sampling for categoricals.


On the other hand, Variational Autoencoders (VAEs) \cite{kingma2013auto} consist of an encoder $q_{\theta}(z|x)$ that maps input $x$ to latent code $z$, and a decoder $p_{\phi}(x|z)$ that reconstructs $x$. VAEs train by maximizing the Evidence Lower Bound (ELBO) on the log-likelihood:

\begin{equation}
  \ell_i(\theta, \phi)
  = -\mathbb{E}_{z \sim q_\phi(z|x_i)} [\log p_\phi(x_i \mid z)]
  + \mathrm{KL}\!\left(q_\theta(z \mid x_i) \,\|\, p(z)\right).
\end{equation} \cite{kingma2013auto}

A negative ELBO form is used in training; the above is a maximization form for clarity.
TVAE specializes VAEs for tabular variables with appropriate encoders/decoders.

Classical methods are transparent and fast but miss complex dependencies; 
modern CTGAN/TVAE variants capture richer mixed-type structure and are \emph{de-facto} baselines \cite{miletic2024challenges} \cite{yadav2024rigorous} \cite{jiang2025synthetic}.

Building on these foundations, recent years have witnessed an explosion of methods for generating synthetic tabular data, ranging from classical statistical approaches to modern deep generative models \cite{shi2025comprehensive} \cite{stoian2025survey} \cite{hassan2023deep} \cite{challagundla2025synthetic}. Traditional methods include probabilistic graphical models (e.g., Bayesian networks, copulas) and rule-based synthesizers. For instance, the DataSynthesizer system \cite{data_synthesizer} employs Bayesian networks to sample from learned conditional distributions, while the \texttt{synthpop} package \cite{nowok2016synthpop} generates data by sequentially modeling each column through regression or classification.

Over the past decade, deep generative models have become the dominant generative approach. Early adaptations of GANs and VAEs, originally developed for image synthesis, were repurposed for tabular data. TableGAN \cite{park2018data} extended the GAN framework with an auxiliary classifier to better capture feature–label dependencies, using convolution layers to model tabular correlations. MedGAN \cite{choi2017generating} was an early GAN-based approach for electronic health records, combining an autoencoder and a GAN to generate high-dimensional binary count vectors, pioneering deep learning for non-image data synthesis. The landmark CTGAN model by Xu \textit{et al.} (2019) \cite{xu2019modeling} introduced conditional sampling and training-by-sampling strategies for categorical columns, significantly outperforming naïve GANs on mixed-type data. The authors also proposed TVAE, showing that a variational autoencoder can achieve comparable fidelity on tabular data.

Subsequent models expanded on these ideas. CTAB-GAN \cite{zhao2021ctab} integrated auxiliary classification loss and novel encodings to handle skewed continuous distributions and imbalanced categoricals, while CTAB-GAN+ \cite{zhao2024ctab} further improved realism and incorporated differential privacy. Beyond GANs and VAEs, flow-based and diffusion models have recently emerged \cite{li2025diffusion}\cite{shi2025comprehensive}. Diffusion models (e.g., TabDDPM, TabDiffusor) generate data by iteratively denoising samples from pure noise and have shown strong potential in overcoming GAN limitations such as mode collapse. For example, TabDDPM \cite{kotelnikov2023tabddpm} and TAB-DIFF \cite{shi2024tabdiff} apply multimodal diffusion to mixed continuous–categorical data, achieving state-of-the-art fidelity. Another emerging line of work employs large language models (LLMs) to synthesize tabular data by serializing rows as text. Although maintaining statistical consistency remains challenging, these methods leverage semantic priors and domain knowledge \cite{dong2024large}.

In summary, deep generative models now dominate tabular data synthesis thanks to their flexibility. CTGAN and TVAE have become de facto baselines, while diffusion and transformer-based approaches continue to push fidelity closer to real data \cite{stoian2025survey}. Our study builds upon these advances: rather than introducing a new generator, we develop a post-processing filter that enhances privacy. This makes our contribution orthogonal to model design: any improvements in generative fidelity can be complemented by our privacy mechanism. Indeed, our results show that even well-trained CTGAN and TVAE outputs benefit from such post-hoc filtering.

\subsection{Mode Collapse and Mode Patching}
\label{sec:mode_collapse_}

Mode collapse occurs when the generator distribution $P_G$ fails to cover part of the true data distribution. Formally, there exists an event $A$ such that $P_{\text{data}}(A) > 0$ but $P_G(A) \approx 0$. In tabular data, this most frequently appears as \emph{categorical mode dropping}, where certain real categories never occur in the synthetic samples. Let $X = (X_{\text{num}}, X_{\text{cat}})$ and let the categorical support of the real data be $M_R = \prod_{j=1}^{q} C_j$. The generator collapses if
\[
M_G = \{\, c : P_G(X_{\text{cat}} = c) > 0 \,\} \subsetneq M_R,
\]
meaning that one or more valid categorical combinations receive zero synthetic probability.

To mitigate this, prior work has introduced several diversity-promoting mechanisms. PacGAN \cite{lin2018pacgan} increases the discriminator’s sensitivity to lack of diversity by feeding it small packs of samples jointly, making it easier to detect collapsed modes \cite{shi2025comprehensive}. AdaGAN \cite{tolstikhin2017adagan} uses a boosting formulation in which each new generator focuses on regions that previous ones under-modeled. MSGAN \cite{mao2019mode} adds a mode-seeking regularizer that explicitly encourages different latent inputs to produce distinguishable outputs.

In tabular settings, mode collapse usually manifests as complete disappearance of minority categorical levels \cite{shafqat2022hybrid}. Conditional GANs alleviate this by conditioning on class labels \cite{dong2022sa}, but when categories are extremely imbalanced, even conditional sampling may collapse to the majority values. Techniques such as oversampling rare levels or integrating additional diversity losses have been explored, and CTGAN itself incorporates oversampling heuristics for minority categories and PacGAN-style discrimination to reduce collapse \cite{xu2019modeling}.

Our work extends this line of research by introducing a \emph{layer-frozen mode-patching} strategy. After the base generator is trained, we detect which categorical levels are missing in the synthetic data and selectively fine-tune only the upper generator layers—freezing the earlier, feature-extracting layers—to restore these missing modes. This targeted fine-tuning recovers categorical support while preserving the rest of the learned distribution, providing an efficient and non-destructive remedy for categorical mode dropping.

\subsection{Privacy Guarantees and Privacy-Preserving Synthesis}
\label{sec:def-privacy}

In synthetic tabular data, several notions are used to reason about privacy \cite{hyrup2024sharing}\cite{palacios2025contrastive}. Among them, \emph{Differential Privacy (DP)} is the only formal, provable guarantee \cite{schlegel2025generating} \cite{hassan2023deep}; the others—such as $\varepsilon$-identifiability and distance-based measures like DCR—serve as practical proxies for assessing whether synthetic records lie too close to real individuals.

A randomized mechanism $K$ is $(\varepsilon,\delta)$-DP if for any adjacent datasets $D,\tilde{D}$ and any measurable output set $\mathcal{S}$,
\begin{equation}
\Pr[K(D)\in\mathcal{S}] \le e^{\varepsilon}\Pr[K(\tilde{D})\in\mathcal{S}] + \delta.
\end{equation}
Training-time approaches such as DP-SGD \cite{abadi2016deep} (gradient clipping plus noise) and the PATE framework \cite{papernot2016semi} provide DP guarantees for synthetic generators (e.g., DP-CTGAN, PATE-CTGAN).

Beyond DP, proximity-based criteria are widely used in practice to ensure synthetic records are not near-duplicates of real individuals. One such measure is \emph{$\varepsilon$-identifiability}, which requires that only an $\varepsilon$ fraction of synthetic samples lie closer to some real record than that record’s nearest real neighbor. Using $r_i$ (real-to-real 2-NN distance) and $\hat{r}_i$ (real-to-synthetic), the identifiability index is
\begin{equation}
I(D,\hat{D}) = \frac{1}{N}\sum_{i=1}^{N}\mathbf{1}[\hat{r}_i < r_i] < \varepsilon.
\end{equation}
This serves as a practical safeguard against near-copies, though it is not a formal privacy guarantee.

A closely related family of proxies is based on the \emph{Distance to Closest Record} (DCR). For each synthetic record $s$, its distances to the nearest real points in the training and holdout subsets are
\begin{equation}
\mathrm{DCR}_{\text{train}}(s) = \min_{x\in D_{\text{train}}} d(s,x), \qquad
\mathrm{DCR}_{\text{test}}(s) = \min_{x\in D_{\text{test}}} d(s,x).
\end{equation}
These distances form DCR curves that reveal whether synthetic points lie abnormally close to training data. A convenient scalar summary is the \emph{Relative Proximity Ratio (RPR)}, which reports the fraction of total nearest-neighbor proximity mass assigned to the training data:
\begin{equation}
\mathrm{RPR} =
\frac{\sum \mathrm{DCR}_{\mathrm{train}}}
     {\sum \mathrm{DCR}_{\mathrm{train}} + \sum \mathrm{DCR}_{\mathrm{test}}}
\times 100.
\end{equation}
An RPR of 50\% indicates that synthetic points are, on average, equally close to training and unseen holdout data. Larger values suggest overfitting or memorization. In our experiments we also report $\Delta\mathrm{RPR}$ (percentage-point change) to measure movement toward or away from the ideal 50\% balance.

These notions connect directly to practical privacy-preserving synthesis. While DP offers the only formal guarantee, many works instead rely on distance-based screening. Rejection sampling based on nearest-neighbor radii is widely used: synthetic samples falling within an $\varepsilon$-ball of any real record are discarded to prevent trivial re-identification. Our second main contribution follows this line: a HEOM–$k$NN $\epsilon_{\mathrm{ANY}}$ filter that removes synthetic points violating a heterogeneous per-record radius. Here, HEOM is a mixed-type Euclidean
distance for numeric and categorical attributes, and
$\varepsilon_{\mathrm{ANY}}$ is an empirical risk that counts the fraction of synthetic records falling inside any real record’s radius; both are formalized in section~\ref{subsec:heom-any}. This approach, related to prior $\varepsilon$-identifiability protection schemes \cite{lautrup2025syntheval}, provides an operational privacy boundary even when formal DP is not applied.

\subsection{Evaluation Frameworks}


The \textbf{resemblance evaluation} of a synthetic dataset aims to assess the similarities between the synthetic data and the original dataset. This evaluation serves as a direct proxy to determine the efficacy of the generative model in synthesizing the data. Typically, this assessment involves a statistical evaluation of the synthetic dataset to ascertain whether it possesses the same statistical properties as the original dataset.


In the context of synthetic tabular data, the \textbf{utility evaluation} is designed to determine whether synthetic datasets can match the utility of real data for specific machine learning (ML) tasks. Specifically, it assesses the efficacy of synthetic tabular data as a substitute for real data across various analytical and machine learning applications. This process is crucial in validating whether the synthetic data can support the same insights and decisions as the original data.

A common approach for evaluating the utility of synthetic data involves comparing the performance of two models: one trained on synthetic data and tested on real data (Train on Synthetic Test on Real, TSTR), and another trained and tested on real data (Train on Real Test on Real, TRTR). A high-utility synthetic dataset would have TSTR results that are equal or near-equal to the TRTR results.

Another complementary approach evaluates whether the feature importance structure induced by synthetic data mirrors that of real data. Giles et al. \cite{giles2022faking} use the Ranked Bias Overlap (RBO) to compare feature-importance rankings—specifically, SHAP-based importance scores—between models trained on real versus synthetic data. RBO quantifies the similarity between the ranked feature-importance lists, providing a sensitive measure of whether synthetic data preserves the relative predictive contributions of individual features.

As synthetic data methods proliferate, rigorous evaluation has become essential. A high-quality synthetic dataset must (1) reproduce real distributions (fidelity), (2) support meaningful analysis (utility), and (3) protect sensitive information (privacy). Several benchmarking initiatives address these criteria. SDGym \cite{sdgym} established an early benchmark for fidelity and machine-learning utility by comparing synthetic and real data across multiple tasks—though early versions neglected privacy, prompting later extensions.

More comprehensive frameworks have since emerged, such as SynthEval \cite{lautrup2025syntheval}, Anonymeter \cite{giomi2022unified}, and SynthCity \cite{griffiths2019synthcity}. SynthEval computes a wide range of fidelity metrics (e.g., univariate divergences via Jensen–Shannon or Wasserstein distances, bivariate correlations such as Pearson’s $r$, Cramér’s $V$, and correlation ratio $\eta^2$), as well as joint-distribution coverage. Utility is typically measured through the Train-on-Synthetic, Test-on-Real (TSTR) protocol \cite{shi2025comprehensive}: if a model trained on synthetic data performs comparably to one trained on real data, the synthetic dataset is deemed useful. Additional metrics such as feature-importance rank correlation (e.g., permutation importance or Rank-Biased Overlap \cite{sarica2022introducing}) quantify analytical consistency.

Privacy evaluations now routinely include simulated attacks. Membership inference tests whether a classifier can distinguish between training and synthetic samples \cite{shi2025comprehensive}, while attribute inference evaluates the ability to predict sensitive variables from partial knowledge. Distance-based measures such as the Distance to Closest Record (DCR) \cite{min2025can} complement these attacks by assessing whether any synthetic record lies too close to a real one. A large average DCR indicates a low risk of memorization. Practitioners also examine exact matches and nearest-neighbor ratios to detect potential re-identification. The Correct Attribution Probability (CAP) measure proposed by Elliot and Taub (2018) \cite{taub2018differential} further formalized the risk of an adversary correctly inferring a target attribute from synthetic data.

Overall, evaluation practice is converging toward multi-criteria assessment—balancing fidelity, utility, and privacy. Our work aligns with this evolution by introducing two complementary contributions: (i) a layer-frozen mode-patching method that restores categorical diversity without retraining from scratch, and (ii) a HEOM–$k$NN $\epsilon_{\mathrm{ANY}}$ rejection filter that enforces a practical privacy boundary while maintaining statistical fidelity. These concepts and metrics are formalized above and further developed in the sections that follow to lay the groundwork for our methodology.

\section{Data Generation Pipeline}
\label{Pipeline}

\begin{figure*}[!t]
\centering
\includegraphics[width=6.2in]{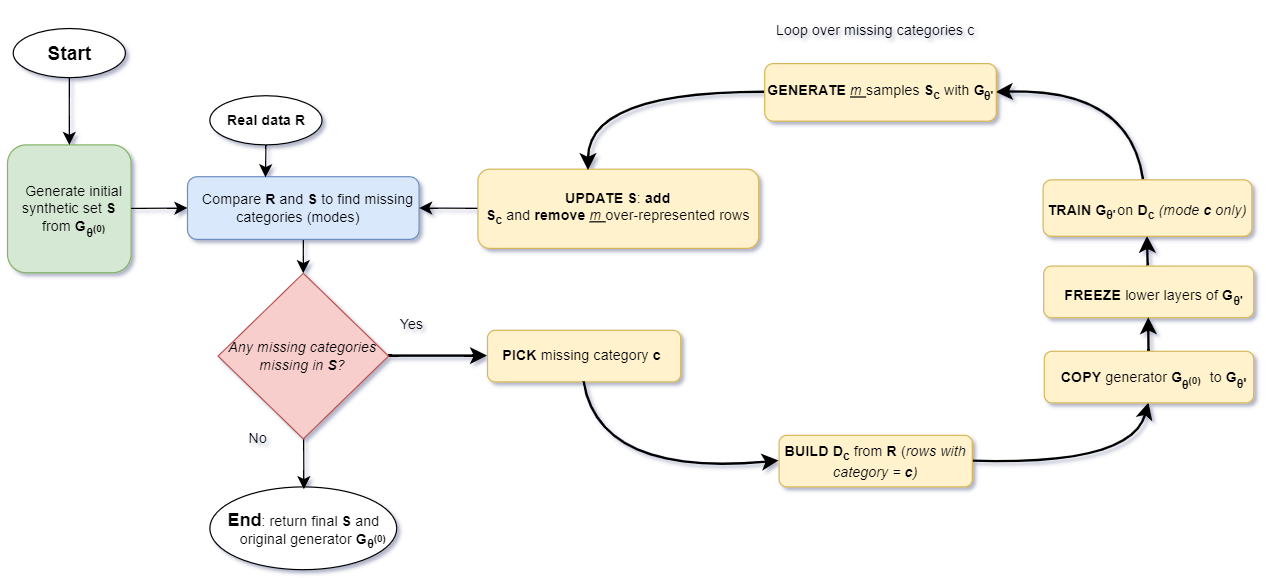}
\caption{\textbf{Iterative layer-frozen mode-patching procedure}. Starting from an initial synthetic set $S$ drawn from the fixed generator $G_\theta^{(0)}$, the algorithm repeatedly detects categorical modes present in the real data $R$ but missing in $S$, fine-tunes a copy of $G_\theta^{(0)}$ on the corresponding slice $D_c$ with its lower layers frozen, and replaces over-represented synthetic rows with samples $S_c$ from the adapted generator. The loop stops once no real category is missing in $S$.}
\label{fig_sim}
\end{figure*}

We adopt a model-agnostic, post-hoc pipeline that can be placed after any trained tabular generator to balance three goals, namely faithfulness to the source data, downstream task utility, and practical privacy protection, without modifying the base model. The pipeline targets two persistent failure modes: categorical mode dropping and proximity-based leakage. An overview of the iterative training/sampling procedure appears in Fig.~\ref{fig_sim}.

First, a layer-frozen ``mode-patching'' stage identifies dropped categories via real--vs.--synthetic crosstabs and restores support by fine-tuning only the late generator layers on the flagged slices, keeping earlier layers (and the base model) fixed to avoid catastrophic forgetting~\cite{park2023train,lin2018pacgan,xu2019modeling,shi2025comprehensive}. Algorithms~\ref{alg:detection} and \ref{alg:patch} formalize the detector and the patch loop without modifying the underlying generator. Second, a HEOM--$k$NN privacy filter embeds mixed-type rows using min--max scaling for numerics and one-hot categoricals scaled by $1/\sqrt{2}$, sets per-record radii from real 2-NN distances, and runs a rejection-with-replacement sampler until the empirical ANY-risk $\widehat{\varepsilon}_{\mathrm{ANY}}$ (formally defined in Section~\ref{subsec:heom-any}) falls below a user-set threshold $\tau_{\mathrm{ANY}}$~\cite{santos2020distance,jiang2025privacy}. The subsections that follow present, in order: the mode-gap filling algorithm; the HEOM--$k$NN filter and its empirical risk criterion; and the evaluation protocol---univariate/bivariate fidelity (Jensen--Shannon distance, Cohen's $d$, Pearson's $r$, Cramér's $V$), utility via TSTR vs.\ TRTR, and privacy checks spanning attribute inference, singling out, linkability, and distance to the closest record.

\subsection{Mode‑gap filling methodology}
\label{sec:mode-gap-filling}

As stated previously, mode collapse is a common issue in synthetic data generation. In this work, we propose to overcome it by fine-tuning our synthetic data generator to these specific missing values. 

Let $X\!=\!(X_\mathrm{num},X_\mathrm{cat})$ be a row sampled from the empirical
distribution $\widehat P_R$ of the real data with $q$ categorical columns
$X_{c_j}\!\in\!\mathcal C_j$.  A mode is any vector
$c=(c_1,\ldots,c_q)\in\mathcal M_R:=\prod_{j=1}^{q}\mathcal C_j$ with
$\widehat P_R(X_\mathrm{cat}=c)>0$.  A generator $G_\theta$ exhibits mode
collapse if the set
$\mathcal M_{G_\theta}:=\{c\mid P_{G_\theta}(X_\mathrm{cat}=c)>0\}$ is a proper
subset of $\mathcal M_R$.

The first step of our methodology is to detect if there are any missing modes, and what their values are. Algorithm~\ref{alg:detection} implements the {\tt get\_mode\_collapse}
procedure used in practice. For each categorical
column $c$, it computes a contingency table between real and synthetic data; any real category with zero synthetic count is flagged.  The detector therefore returns the set $\mathcal M^{(k)}=\{c\in\mathcal M_R\mid\widehat P_{S^{(k)}}(X_\mathrm{cat}=c)=0\}$.

\begin{algorithm}[!t]
    \caption{Categorical mode‑collapse detector}
    \label{alg:detection}
    \KwIn{real set $R$, synthetic set $S$, categorical columns $\mathcal{C}$}
    \KwOut{dictionary $\textit{Missing}$: column $\to$ list of unseen categories}

    $\textit{Missing} \gets \{\}$\;
    \For(\tcp*[f]{iterate over categorical columns}){$c \in \mathcal{C}$}{
        $\textit{tab} \gets \textsc{crosstab}\bigl(R[c],\,S[c]\bigr)$\;
        $\textit{Missing}[c] \gets 
            \{\,k \mid k \in R[c]\;\wedge\; \textit{tab}[k]=0\,\}$\;
    }
    \KwRet{$\textit{Missing}$}\;
\end{algorithm}

\begin{figure}[!t]
    \centering
    \includegraphics[height=0.25\textheight]{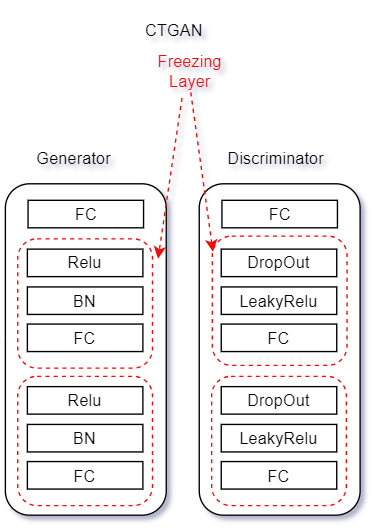}
    \caption{\textbf{Schematic of the CTGAN architecture highlighting frozen layers (in red) during fine-tuning}. The generator’s lower layers (up to the freezing point) remain fixed while only the later layers are retrained to produce a specific rare class. A similar freezing approach can be applied to the discriminator’s feature layers to maintain stability.}
    \label{figure:ctgan_freeze}
\end{figure}

\begin{figure}[!t]
    \centering
    \includegraphics[height=0.2\textheight]{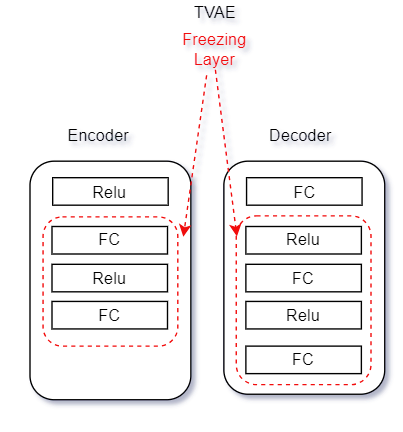}
        \caption{\textbf{Schematic of the TVAE (tabular VAE) architecture with a freezing point between the encoder and decoder}. In fine-tuning, the encoder and the first part of the decoder are frozen (red), and only the last layers of the decoder are adjusted. This focuses the adaptation on generating a missing category without disturbing the overall latent structure.}
        \label{figure:tvae_freeze}
\end{figure}

Next, we fine-tune a synthetic data generator to generate data points in each of the missing modes. 
Empirical studies on transfer learning for GANs show that freezing $60$–$80\%$ of the lowest layers yields the best trade‑off between stability and adaptation \cite{park2023train}.  Figures$\sim$\ref{figure:ctgan_freeze}–\ref{figure:tvae_freeze} illustrate the split points for CTGAN and TVAE used throughout this work.  During patching, parameters below the dashed line are held fixed, so the fine‑tune updates operate in a low‑rank subspace of the original weight manifold, mitigating catastrophic forgetting.

Algorithm~\ref{alg:patch} formalizes the complete procedure. 

\begin{algorithm}[!t]
\caption{Iterative layer‑frozen mode‑patching}
\label{alg:patch}
\KwIn{trained generator $G_{\theta^{(0)}}$, real data $R$, batch size $B$,
threshold $\varepsilon$, sample budget $m{=}10$.}
\KwOut{synthetic set $S$, (unchanged) generator $G_{\theta^{(0)}}$}
$S\gets G_{\theta^{(0)}}(P_Z^{B})$\tcp*[r]{initial synthetic batch}
$k\gets 0$\;
\Repeat{$\mathcal M^{(k)}=\varnothing$}{
  $\mathcal M^{(k)}\gets$ {\sc get\_mode\_collapse}$(R,S)$\;
  \ForEach{missing mode $c\in\mathcal M^{(k)}$}{
    $D_c\!\gets\!\big\{x\!\in\!R: x_\mathrm{cat}=c\big\}$\;
    \If{$|D_c|<m$}{augment $D_c$ with $m-|D_c|$
      random rows from $R$ (oversampling).}
    $\theta'\!\gets\!\theta^{(0)}$\tcp*[r]{\#1 deep copy}
    \textsc{FreezeLayers}($\theta'$ below split layer)\tcp*[r]{\#2}
    \textsc{Train}($G_{\theta'}$ on $D_c$)\;
    $S_c\!\gets\!G_{\theta'}(P_Z^{m})$\;
    $S\!\gets\!(S\cup S_c)\setminus$ $m$ rows that
        carried the most common value in column of $c$\;
  }
  $k\gets k+1$\;
}
\Return{$S, G_{\theta^{(0)}}$}\;
\end{algorithm}

Freezing the majority of layers during generator transfer was recently advocated as
a robust remedy against collapse in image GANs \cite{park2023train}.  Our
algorithm extends that idea to the {\em tabular} setting and couples it with an
iterative re‑weighting loop inspired by packing‑based diversity promotion
\cite{lin2018pacgan}.  Unlike one‑shot class‑conditional sampling in CTGAN
\cite{xu2019modeling}, our method guaranties explicit coverage for {\em every}
categorical combination, aligning with the desiderata identified in the latest
survey on synthetic tabular data generation \cite{shi2025comprehensive}.

\subsection{Privacy Filtering via HEOM--kNN Radius ({$\widehat{\varepsilon}_{\mathrm{ANY}}$} with threshold {$\tau_{\mathrm{ANY}}$})}
\label{subsec:heom-any}

The goal of this portion of the pipeline is to reduce the risk that a model-generated record is unusually close to an individual in the training data. It does this by enforcing a nearest-neighbor radius criterion measured in a mixed-type metric space.

We embed mixed-type records into a Euclidean space using the Heterogeneous Euclidean Overlap Metric (HEOM) \cite{santos2020distance}.
Let $\Enc:\mathcal{X}\!\to\!\mathbb{R}^d$ denote the encoder that applies: (i) min--max normalization to each numeric feature into $[0,1]$; and (ii) one-hot encoding for each categorical feature, scaled by $1/\sqrt{2}$ so that two distinct categories have unit Euclidean separation.
Distances are then

\begin{equation}
d(x,y) \;=\; \bigl\| \Enc(x) - \Enc(y) \bigr\|_{2}, \qquad x,y \in \mathcal{X}.
\end{equation}

For each real record $x_i$, define its \emph{privacy radius}

\begin{equation}
r_i \;=\; \min_{k \neq i} \bigl\| \Enc(x_i) - \Enc(x_k) \bigr\|_{2}.
\end{equation}

i.e., the 2-NN distance among the reals \cite{jiang2025privacy}. A synthetic record $s$ is \emph{unsafe} if it falls within the radius of any real:

\begin{equation}
\exists\, i \in [N] \;:\; d(s, x_i) < r_i.
\end{equation}

Equivalently, define the \emph{squared margin}

\begin{equation}
M(s) \;=\; 
\min_{i \in [N]} \Bigl( \bigl\| \Enc(s) - \Enc(x_i) \bigr\|_{2}^{2} - r_i^{2} \Bigr).
\end{equation}

Then $s$ is unsafe iff $M(s)<0$. For a synthetic set $S=\{s_j\}_{j=1}^n$, the (empirical) ANY-risk is

\begin{equation}
\widehat{\varepsilon}_{\mathrm{ANY}}(S)
\;=\;
\frac{1}{n} \sum_{j=1}^{n} \mathbf{1}\!\left[ M(s_j) < 0 \right].
\end{equation}

We then generate an initial pool $S$ from $G$ and compute $M(s_j)$ for all $s_j\in S$.
If $\widehat{\varepsilon}_{\mathrm{ANY}}(S)<\tau_{\mathrm{ANY}}$ we stop. Otherwise, we iteratively replace the \emph{worst} record $s_{j^\star}$ with a fresh draw $s'\sim G$ whenever this strictly improves privacy:

\begin{equation}
\begin{aligned}
&\bigl(M(s_{j^\star}) < 0 \;\wedge\; M(s') \ge 0\bigr)
\\[2pt]
&\qquad\text{or}\;\;
\bigl(M(s_{j^\star}) \ge 0 \;\wedge\; M(s') > M(s_{j^\star})\bigr).
\end{aligned}
\end{equation}

The loop continues until $\widehat{\varepsilon}_{\mathrm{ANY}}(S)<\tau_{\mathrm{ANY}}$ or no further improvement is obtained (feasibility depends on $G$). For efficiency, we precompute $r_i$ and use nearest-neighbor indexing in the encoded space. Let $R_{\max}:=\max_i r_i$; then a radius-$R_{\max}$ query suffices to detect violations, since if all reals lie beyond $R_{\max}$ from a candidate $z$, $M(z)\ge \|z-x_{\text{NN}}\|^2-R_{\max}^2\ge 0$ where $x_{\text{NN}}$ is the (1-NN) real neighbor.

\begin{algorithm}[t]
\small
\DontPrintSemicolon
\caption{HEOM--kNN \texorpdfstring{$\widehat{\varepsilon}_{\mathrm{ANY}}$}{epsilon\_ANY-hat} Rejection-with-Replacement (target $\tau_{\mathrm{ANY}}$)}
\label{alg:heom-any}
\KwIn{Real data $D_{\mathrm{real}}$ with numeric $\mathcal{N}$ and categorical $\mathcal{C}$; generator $G$; threshold $\tau_{\mathrm{ANY}}$ (implementation name \texttt{min\_eps}); sample size $n$.}
\KwOut{Synthetic dataset $S$ with $\widehat{\varepsilon}_{\mathrm{ANY}}(S)<\tau_{\mathrm{ANY}}$ (if feasible).}

\BlankLine
\textbf{Encode \& index.}
Fit min--max on $\mathcal{N}$; one--hot $\mathcal{C}$ with scale $1/\sqrt{2}$ to obtain encoder $\Enc$.\;
$X_r \leftarrow \Enc(D_{\mathrm{real}})$.\;
Compute radii $r_i$ as 2-NN distances of $x_i\in X_r$; set $R_{\max}\leftarrow \max_i r_i$.\;
Build 1-NN and radius-$R_{\max}$ indexes over $X_r$.\;

\BlankLine
\textbf{Initialize.}
Draw $n$ samples $S\sim G$ and set $X_s\leftarrow \Enc(S)$.\;
For each $x\in X_s$, compute
$\displaystyle M(x)=\min_{i}\bigl(\|x-x_i\|_2^2-r_i^2\bigr)$ using the radius-$R_{\max}$ index; if no neighbor is found within $R_{\max}$, use $M(x)\leftarrow \|x-x_{\text{NN}}\|_2^2-R_{\max}^2$.\;
$\displaystyle \widehat{\varepsilon}_{\mathrm{ANY}}(X_s)\leftarrow \frac{1}{n}\sum_{x\in X_s}\mathbf{1}[M(x)<0]$.\;

\BlankLine
\textbf{Iterate.}
\While{$\widehat{\varepsilon}_{\mathrm{ANY}}(X_s) \ge \tau_{\mathrm{ANY}}$}{
  Select the worst record $j^\star \leftarrow \arg\min_j M(X_s[j])$.\;
  Propose $s'\sim G$, set $x'\leftarrow \Enc(s')$, and compute $M(x')$ as above.\;
  \If(\tcp*[f]{accept if it fixes a violation or increases a safe margin}){$\big(M(X_s[j^\star])<0 \wedge M(x')\ge 0\big)\ \lor\ \big(M(X_s[j^\star])\ge 0 \wedge M(x')>M(X_s[j^\star])\big)$}{
    Replace $S[j^\star]\leftarrow s'$, $X_s[j^\star]\leftarrow x'$, and $M(X_s[j^\star])\leftarrow M(x')$.\;
    Update $\displaystyle \widehat{\varepsilon}_{\mathrm{ANY}}(X_s)\leftarrow \frac{1}{n}\sum_{x\in X_s}\mathbf{1}[M(x)<0]$.\;
  }
}
\Return{$S$}
\end{algorithm}

\subsection{Fidelity Evaluation}
\label{subsec:fidelity-evaluation}
The objective of this first evaluation step is to confirm that the synthetic data faithfully represents the statistical characteristics of the original dataset, thereby ensuring its validity for subsequent applications. 
Following similar previous works \cite{platzer2021holdout}\cite{basri2025useful}\cite{barr2025zero}, we focus primarily on univariate and bivariate fidelity assessments. In other words, we analyze the distribution of individual variables (univariate) as well as the relationships between pairs of variables (bivariate).


For the \textbf{univariate analysis}, we opted not to employ p-values as an evaluation metric. This decision is informed by the sensitivity of p-values to sample size; larger datasets tend to produce more statistically-significant outcomes, even for effects that are practically negligible \cite{cohen2013statistical}. Given this consideration, it is inappropriate to employ this method in contexts where the dataset exceeds 5,000 observations. In contrast, the use of effect size measures facilitates a robust comparison of univariate distributions. This approach allows for a more meaningful assessment of the magnitude of differences, irrespective of sample size \cite{aarts2014importance}.

For these reasons, within our framework, we employ both Jensen-Shannon (JS) distance and Cohen's $d$ measure. 
For categorical variables we report the \emph{Jensen--Shannon distance}, defined as the square root of the Jensen--Shannon divergence:

\begin{equation}
JS\!\left(p_R(v), p_S(v)\right)
= \tfrac{1}{2}\, KL\!\left(p_R \,\middle\|\, m\right)
+ \tfrac{1}{2}\, KL\!\left(p_S \,\middle\|\, m\right),
\end{equation}

\begin{equation}
m \;=\; \tfrac{1}{2} \left( p_R + p_S \right).
\end{equation}

where $KL(\cdot\|\cdot)$ denotes the Kullback--Leibler divergence. 
Taking the square root yields a true metric bounded in $[0,1]$, often called the \emph{Jensen--Shannon distance}, which improves interpretability by making the scale approximately linear with respect to distributional dissimilarity. 
Lower values indicate higher resemblance between real and synthetic marginals.

For numeric variables we report the absolute Cohen’s~$d$,

\begin{equation}
\lvert d(v) \rvert
= \frac{\lvert \mu_S(v) - \mu_R(v) \rvert}{s_p(v)}.
\end{equation}

\begin{equation}
s_p(v)
= \sqrt{\frac{(n_R - 1)s_R^{2} + (n_S - 1)s_S^{2}}{n_R + n_S - 2}}.
\end{equation}

using the pooled standard deviation $s_p(v)$; again, lower is better.


The \textbf{bivariate evaluation} assesses whether pairwise relationships between features in the real dataset are preserved in the synthetic dataset. We quantify this by computing pairwise association measures for every pair of variables in the real dataset and in the synthetic dataset, and comparing the resulting dependence matrices. Preserving these associations is a strong indicator that the synthetic data reproduces the multivariate structure of the original dataset.

Since numerical, categorical, and mixed feature pairs do not share a single universal dependence measure, we use the measure best suited to each type: Pearson’s correlation coefficient $\rho$ for numerical pairs, Cramer’s $V$ for categorical pairs, and the correlation ratio $\eta^{2}$ for numerical–categorical pairs. For each measure $f \in \{\rho, V, \eta^{2}\}$, we compute a dependence matrix on the real data, $\mathbf{C}^{(f)}_{\mathrm{real}}$, and an analogous matrix on the synthetic data, $\mathbf{C}^{(f)}_{\mathrm{syn}}$.

To summarize the discrepancy between the real and synthetic dependence structures, we report two complementary metrics. First, we use the Frobenius norm of the difference between the real and synthetic dependence matrices:
\begin{equation}
D^{(f)}_{\mathrm{F}}
= \bigl\| \mathbf{C}^{(f)}_{\mathrm{real}} - \mathbf{C}^{(f)}_{\mathrm{syn}} \bigr\|_{F}
= \sqrt{ \sum_{i=1}^{d_f} \sum_{j=1}^{d_f}
\bigl( C^{(f)}_{\mathrm{real},ij} - C^{(f)}_{\mathrm{syn},ij} \bigr)^{2} },
\end{equation}
where $\|\cdot\|_{F}$ denotes the Frobenius norm and $d_f$ is the dimension of the matrix for family $f$.

Second, to capture similarity in the \emph{ordering} of pairwise associations, we compute the Spearman rank correlation between the off-diagonal entries of the real and synthetic matrices. Let $\mathrm{vec}_{\mathrm{off}}(\mathbf{A})$ denote the vector obtained by stacking all off-diagonal elements of a square matrix $\mathbf{A}$. For each family $f$ we define
\begin{equation}
r^{(f)}_{\mathrm{S}}
= \rho_{\mathrm{S}}\!\left(
\mathrm{vec}_{\mathrm{off}}\!\bigl(\mathbf{C}^{(f)}_{\mathrm{real}}\bigr),
\mathrm{vec}_{\mathrm{off}}\!\bigl(\mathbf{C}^{(f)}_{\mathrm{syn}}\bigr)
\right),
\end{equation}
where $\rho_{\mathrm{S}}(\cdot,\cdot)$ denotes Spearman's rank correlation coefficient.
Lower values of $D^{(f)}_{\mathrm{F}}$ and higher values of $r^{(f)}_{\mathrm{S}}$ indicate better preservation of bivariate structure in the synthetic data.

\subsection{Utility Evaluation}

The second evaluation step aims to verify that the synthetic dataset can be used to train machine learning (ML) algorithms that generalize well to real data. Following prior work on Train-on-Synthetic, Test-on-Real (TSTR) evaluation, we compare predictive performance between models trained on synthetic versus real data, and we complement this with an analysis of feature-importance consistency derived from XGBoost.

Let $D_{\text{train}}$ and $D_{\text{test}}$ denote the real training and test sets, respectively, and let $\hat{D}_{\text{train}}$ be a synthetic training set generated to mimic $D_{\text{train}}$. We consider a suite of $K$ downstream classifiers (here $K=8$). For each base learner $a_k$ ($k=1,\dots,K$), we train two models:
\[
f_k^{\mathrm{TRTR}} = a_k(D_{\text{train}}),
\qquad
f_k^{\mathrm{TSTR}} = a_k(\hat{D}_{\text{train}}),
\]
corresponding to the Train-on-Real/Test-on-Real (TRTR) and Train-on-Synthetic/Test-on-Real (TSTR) scenarios, respectively.

Let $\mathcal{M}$ denote the set of bounded predictive metrics we report,
\[
\mathcal{M} = \{\text{accuracy},\ \text{balanced accuracy},\ \text{weighted }F_{1},\ \text{ROC AUC}\},
\]
all of which take values in $[0,1]$ with higher being better. For any $M \in \mathcal{M}$, we define the performance of classifier $k$ under each protocol as
\begin{equation}
M_k^{\mathrm{TRTR}} = M\bigl(f_k^{\mathrm{TRTR}}, D_{\text{test}}\bigr),
\qquad
M_k^{\mathrm{TSTR}} = M\bigl(f_k^{\mathrm{TSTR}}, D_{\text{test}}\bigr).
\label{eq:metric_tstr_trtr}
\end{equation}
We summarize performance across the classifier suite by the cross-model means
\begin{equation}
\bar{M}^{\mathrm{TRTR}} = \frac{1}{K}\sum_{k=1}^{K} M_k^{\mathrm{TRTR}},
\qquad
\bar{M}^{\mathrm{TSTR}} = \frac{1}{K}\sum_{k=1}^{K} M_k^{\mathrm{TSTR}}.
\label{eq:metric_means}
\end{equation}

The \emph{utility gap} for metric $M$ is then defined as the absolute difference between these averages,
\begin{equation}
\Delta_M = \bigl|\bar{M}^{\mathrm{TRTR}} - \bar{M}^{\mathrm{TSTR}}\bigr|,
\label{eq:utility_gap_M}
\end{equation}
which is bounded in $[0,1]$ (and reported as an absolute percentage in the experiments). A synthetic dataset with $\Delta_M$ close to zero for all $M \in \mathcal{M}$ is considered high-utility, as it yields predictive performance comparable to that of models trained on real data.

In particular, for the (weighted) $F_{1}$ score we write
\begin{equation}
\Delta_{\mathrm{F1}} = \bigl|\bar{F}_{1}^{\mathrm{TRTR}} - \bar{F}_{1}^{\mathrm{TSTR}}\bigr|,
\label{eq:utility_gap_F1}
\end{equation}
which specializes the utility gap in \eqref{eq:utility_gap_M} to $M = F_{1}$. The $F_{1}$-based utility gap used later on for hyperparameter optimization corresponds to \eqref{eq:utility_gap_F1} in the single-classifier setting.

To probe whether synthetic data preserves the \emph{structure} of feature contributions, we compute permutation feature importance (PFI) for XGBoost models trained on real versus synthetic data. Let $f$ denote a trained classifier (either $f^{\mathrm{TRTR}}_k$ or $f^{\mathrm{TSTR}}_k$) and let $D_{\text{test}}^{\pi(j)}$ be the test set obtained by independently permuting the $j$-th feature across samples while leaving all other features and labels fixed. Using the same performance metric $M \in \mathcal{M}$, the permutation feature importance of feature $j \in \{1,\dots,d\}$ for model $f$ is defined as
\begin{equation}
\operatorname{PFI}_j(f; M)
= M\bigl(f, D_{\text{test}}\bigr)
 - M\bigl(f, D_{\text{test}}^{\pi(j)}\bigr).
\label{eq:pfi_def}
\end{equation}
Since $M$ is a bounded, higher-is-better metric, larger values of $\operatorname{PFI}_j(f; M)$ indicate features whose perturbation causes the greatest degradation in predictive performance. For each model (real-trained vs.\ synthetic-trained), this yields a vector of scores
\[
\bigl(\operatorname{PFI}_1(f; M),\dots,\operatorname{PFI}_d(f; M)\bigr),
\]
which we sort in descending order to obtain feature-importance rankings.

Let $L^{(\mathrm{real})}$ and $L^{(\mathrm{syn})}$ denote the ranked lists of feature indices induced by PFI for XGBoost trained on $D_{\text{train}}$ and $\hat{D}_{\text{train}}$, respectively. For a depth $d$, let $L^{(\mathrm{real})}_{1:d}$ and $L^{(\mathrm{syn})}_{1:d}$ be the top-$d$ prefixes of these lists. The fraction of overlap at depth $d$ is
\begin{equation}
A_d = \frac{1}{d}\,\bigl|\,L^{(\mathrm{real})}_{1:d} \cap L^{(\mathrm{syn})}_{1:d}\bigr|,
\label{eq:rbo_overlap}
\end{equation}
and the Rank-Biased Overlap (RBO) \cite{webber2010similarity} between the two rankings with top-weighting parameter $p \in [0,1]$ is given by
\begin{equation}
\mathrm{RBO}\bigl(L^{(\mathrm{real})}, L^{(\mathrm{syn})}; p\bigr)
= (1-p)\sum_{d=1}^{\infty} p^{d-1} A_d.
\label{eq:rbo_def}
\end{equation}
In practice, the sum in \eqref{eq:rbo_def} is truncated at $d=d_{\max}$, the number of features of interest. The RBO score lies in $[0,1]$, with values near~1 indicating that the synthetic-trained and real-trained models induce highly similar PFI rankings. Thus, \eqref{eq:pfi_def}--\eqref{eq:rbo_def} provide an analytical complement to the TSTR/TRTR utility gaps in \eqref{eq:utility_gap_M}--\eqref{eq:utility_gap_F1}, quantifying how faithfully synthetic data preserves the relative predictive contributions of individual features.

\subsection{Privacy Evaluation}

The final evaluation step in our pipeline aims to verify that the synthetic data points are sufficiently different from the real data points so that information about any individual is unlikely to leak through the synthetic dataset. Our privacy analysis combines the distance-based proxies introduced in Section~\ref{sec:def-privacy} with four empirical assessments: (i) Attribute Inference Attacks (AIA) and Correct Attribution Probability (CAP), (ii) Singling Out, (iii) Linkability, and (iv) Distance to Closest Record (DCR) analyses based on the Relative Proximity Ratio (RPR). Each of these metrics probes a complementary aspect of disclosure risk in synthetic data.

The \textbf{Attribute Inference Attack (AIA)} evaluation addresses the risk that an adversary, given partial information about an individual, can infer sensitive attributes more accurately because of access to the synthetic data. For each target attribute $Y$ and a set of quasi-identifiers $X_Q$, we train an attacker model $f_{\mathrm{AIA}}$ on synthetic samples and evaluate it on held-out real records $\{(x_Q^{(i)}, y^{(i)})\}_{i=1}^N$, producing predictions $\hat{y}^{(i)} = f_{\mathrm{AIA}}(x_Q^{(i)})$. For categorical targets, attack success is measured by top-1 accuracy
\begin{equation}
\mathrm{AIA}_{\mathrm{acc}} = \frac{1}{N} \sum_{i=1}^{N} \mathbf{1}\bigl[\hat{y}^{(i)} = y^{(i)}\bigr],
\end{equation}
where $\mathbf{1}[\cdot]$ denotes the indicator function. For continuous targets, we report the coefficient of determination
\begin{equation}
R^2_{\mathrm{AIA}} = 1 - \frac{\sum_{i=1}^{N} \bigl(y^{(i)} - \hat{y}^{(i)}\bigr)^2}{\sum_{i=1}^{N} \bigl(y^{(i)} - \bar{y}\bigr)^2}, 
\qquad 
\bar{y} = \frac{1}{N} \sum_{i=1}^{N} y^{(i)}.
\end{equation}
Higher values of $\mathrm{AIA}_{\mathrm{acc}}$ or $R^2_{\mathrm{AIA}}$ indicate stronger attribute leakage. Complementing these attack-based scores, we also compute the Correct Attribution Probability (CAP) for each confidential categorical attribute $Y_j$ following Elliot and Taub (2018). Let $p_i$ denote the posterior probability (estimated from the synthetic data) that an intruder assigns to the true category $y_j^{(i)}$ of record $i$ given its quasi-identifiers $x_Q^{(i)}$; with $N_j$ denoting the number of real records with non-missing $Y_j$, the CAP for attribute $j$ is
\begin{equation}
\mathrm{CAP}_j = \frac{1}{N_j} \sum_{i=1}^{N_j} p_i.
\end{equation}
The SDV implementation returns the corresponding protection score $\mathrm{cap\_prot}_j = 1 - \mathrm{CAP}_j$, which we report as a percentage and summarize by the median across attributes.

The final privacy assessment is the \textbf{Closest Distance Record (DCR)} evaluation, which quantifies how close synthetic records lie to the training data compared to unseen hold-out data. For each synthetic record $s$, we compute its Distance to Closest Record in the training and test (hold-out) sets, $\mathrm{DCR}_{\mathrm{train}}(s)$ and $\mathrm{DCR}_{\mathrm{test}}(s)$, as defined in Section~III-C by equations~(5)--(6). Aggregating these distances across all synthetic records yields the Relative Proximity Ratio (RPR) in equation~(7), which measures the proportion of overall nearest-neighbor proximity mass that falls toward the training data rather than the test data; values near $50\%$ indicate that training and test sets are, on average, equally close to the synthetic samples, whereas values substantially above $50\%$ suggest possible memorization of training records.

\section{Experimental Results}
\label{Experiments}


Unless stated otherwise, differences on bounded predictive metrics (accuracy, balanced accuracy, weighted $F_1$, ROC~AUC) are reported as \emph{absolute percentage (\%)} differences. Relative changes on divergence or distance metrics (e.g., JS divergence, Cohen's~$d$) are also reported in \emph{percent (\%)}. Metric directionality: lower is better for JS, Cohen's~$d$, Frobenius norms, and log-loss; higher is better for AUC, accuracy, balanced accuracy, weighted $F_1$, and Spearman rank correlation. We consistently write the privacy threshold as $\tau$ and refer to the Default of Credit Card Clients dataset as \emph{Credit Card (UCI)} (abbrev.\ \emph{Credit}), the Cardiovascular dataset as \emph{Cardio}, and \emph{Adult} remains unchanged.

\subsection{Datasets}
We evaluate models on three widely-used, mixed-type tabular benchmarks spanning finance, health, and socioeconomics. Each dataset provides a binary prediction target and a combination of real-valued and discrete (nominal/ordinal) features, making them suitable for assessing resemblance metrics  (e.g., Pearson’s~$\rho$, Cramér’s~$V$, correlation ratio~$\eta^2$), TSTR utility, and privacy risks (e.g., attribute inference) under post-hoc sampling control (HEOM--$k$NN $\varepsilon_{\mathrm{ANY}}$ filtering) and mode-collapse mitigation.

\begin{table}[t]
\caption{Benchmarks used in this study. $N$: instances; $d$: features (excluding target);
$d_{\text{num}}$: numerical (real-valued); $d_{\text{cat}}$: categorical/ordinal. Counts follow our
modeling convention of treating ordinal-coded variables as categorical.}
\label{tab:datasets}
\centering
\setlength{\tabcolsep}{3.2pt}      
\renewcommand{\arraystretch}{1.05}  
\scriptsize                        
\begin{tabularx}{\columnwidth}{@{}lrrrrl}
\toprule
Dataset & $N$ & $d$ & $d_{\text{num}}$ & $d_{\text{cat}}$ & Task \\
\midrule
Default of Credit Card Clients (UCI) & 30{,}000 & 23 & 14 & 9 & Binary classification \\
Cardiovascular Disease (Kaggle)      & 70{,}000 & 11 & 5  & 6 & Binary classification \\
Adult (Census Income\; UCI)           & 48{,}842 & 14 & 6  & 8 & Binary classification \\
\bottomrule
\end{tabularx}
\vspace{-1mm}
\end{table}

The \textbf{Default of Credit Card Clients}\footnote{\href{https://doi.org/10.24432/C55S3H}{UCI Default of Credit Card Clients (DOI: 10.24432/C55S3H)}.} dataset from UCI contains $N{=}30{,}000$ credit card clients with $d{=}23$ features and a binary target indicating default in the next month. Features include demographic variables (\texttt{SEX}, \texttt{EDUCATION}, \texttt{MARRIAGE}, \texttt{AGE}); credit limit (\texttt{LIMIT\_BAL}); six months of repayment status (\texttt{PAY\_0}, \texttt{PAY\_2}–\texttt{PAY\_6}; ordinal); monthly bill amounts (\texttt{BILL\_AMT1}–\texttt{BILL\_AMT6}) and payments (\texttt{PAY\_AMT1}–\texttt{PAY\_AMT6}). The repeated monthly structure induces strong intra-group correlations and heavy-tailed monetary variables, making it an informative stress-test for resemblance metrics and for privacy evaluation in a financially sensitive domain. In the rest of this paper we will refer to this dataset as \textit{Credit} for simplicity.

The \textbf{Cardiovascular Disease}\footnote{\url{https://www.kaggle.com/datasets/sulianova/cardiovascular-disease-dataset}.} dataset from Kaggle is a health dataset comprising $N{=}70{,}000$ patient records with $d{=}11$ features and a binary target \texttt{cardio}. Numerical features are age (in days), height, weight, systolic and diastolic blood pressure (\texttt{ap\_hi}, \texttt{ap\_lo}); categorical/ordinal features include \texttt{gender}, \texttt{cholesterol} (3 levels), \texttt{gluc} (3 levels), and lifestyle indicators (\texttt{smoke}, \texttt{alco}, \texttt{active}). The mixture of clinical measurements and discretized risk indicators yields heterogeneous marginals and nontrivial dependencies characteristic of biomedical registries, and the medical context is directly relevant to privacy analysis. In the rest of this paper we will refer to this dataset as \textit{Cardio} for simplicity.

The \textbf{Adult Census Income}\footnote{\href{https://doi.org/10.24432/C5XW20}{UCI Adult (Census Income) (DOI: 10.24432/C5XW20)}.} dataset from UCI contains $N{=}48{,}842$ records with $d{=}14$ features extracted from the 1994 US Census; the target is whether income exceeds \$50K/year. Features include six numerical attributes (\texttt{age}, \texttt{fnlwgt}, \texttt{education-num}, \texttt{capital-gain}, \texttt{capital-loss}, \texttt{hours-per-week}) and eight categorical attributes (e.g., \texttt{workclass}, \texttt{education}, \texttt{marital-status}, \texttt{occupation}, \texttt{relationship}, \texttt{race}, \texttt{sex}, \texttt{native-country}). Its mix of high-cardinality categoricals and skewed continuous variables (e.g., capital gains/losses), combined with widely reported use as a fairness and privacy benchmark, makes it an appropriate testbed for both fidelity and privacy metrics. In the rest of this paper we will refer to this dataset as \textit{Adult} for simplicity.

We selected these datasets because, in addition to covering three different domains, they exhibit a number of interesting features. Notably, they: (i) exhibit heterogeneous feature types and scales (continuous, ordinal, nominal); (ii) present correlation structure (e.g., repeated monthly panels, socioeconomic groupings) that challenges generative models; (iii) involve sensitive attributes and outcomes (finance, health, income) that are pertinent for privacy evaluation; and (iv) are widely used and well-documented public benchmarks, facilitating reproducibility and comparability.

\subsection{Hyperparameter Optimization for Synthetic Data Generators }

In this first experiment, our objective is to find the optimal hyper-parameters to with which to fine-tune our two deep generative models (CTGAN and TVAE) in order to produce a high-quality synthetic tabular dataset that closely mimics the real dataset it is based on. Our downstream objective is that a classifier trained on our synthetic data should perform almost as well as one trained on the real data. 
Our metric of success will be to minimize the performance gap of a classifier trained using synthetic and real data.
Achieving a small gap in predictive performance (especially F1 score) indicates that the synthetic data preserves the important statistical patterns of the real data \cite{santangelo2025good}.


To optimize the synthetic data generator for each dataset, we perform Bayesian hyperparameter optimization using Optuna \cite{akiba2019optuna}. The hyperparameter search space is defined to cover key model settings for CTGAN and TVAE, aiming to identify configurations that yield the best utility. Notably, we tune parameters that control the model’s complexity and learning dynamics:

\begin{itemize}

\item \textbf{Network dimensions:} Embedding size for the latent representation (tested values: 128, 256, 512), as well as the hidden layer sizes of the generator and discriminator networks. For CTGAN, we explore different two-layer architectures (e.g. [256,256], [512,512] or mixed [256,512] for generator; and [256,256] or [512,512] for discriminator). For TVAE, analogous parameters are the encoder \texttt{compress\_dims} and decoder \texttt{decompress\_dims} layers (with similar layer-size options). These control model capacity to capture tabular patterns.

\item \textbf{Learning rates:} For CTGAN’s adversarial training, separate generator and discriminator learning rates are sampled (log-uniformly in the range $1\times10^{–5}$ to $1\times10^{-3}$). This allows the optimization to find a stable training regime for the GAN. (TVAE uses a single VAE optimizer; we kept its default learning schedule and focused on regularization instead, e.g. a fixed L2 weight decay of $1\times10^{–5}$.)

\item \textbf{Training parameters:} We vary the batch size (100, 500, or 1000) and the number of training epochs (ranging from 800 up to 1500) for both CTGAN and TVAE. A higher epoch count gives the model more opportunity to fit the data, while batch size can affect training stability and mode coverage. Other advanced settings (such as CTGAN’s PacGAN grouping of 10 samples to mitigate mode collapse \cite{xu2019modeling}) are kept constant as per best practices from the literature.

\end{itemize}

We use Optuna’s Tree-structured Parzen Estimator (TPE) sampler \cite{akiba2019optuna} to efficiently navigate this search space. For each trial, the objective function trains the generative model with a candidate hyperparameter set and runs the full pipeline (generation + evaluation). The objective value returned is the absolute F1 score difference between the real-trained and synthetic-trained classifiers when tested on real data, as shown in Equation \ref{eq:diffF1}. 
This captures the utility gap and needs to be minimized: an ideal synthetic dataset would achieve a score of 0, indicating its utility matches real data. The optimization is set to minimize this gap. By iteratively sampling and evaluating 50 trials per experiment, the Bayesian optimizer hones in on hyperparameter combinations that minimize the F1 difference, thereby maximizing synthetic data utility. This approach aligns with recent evaluation recommendations that emphasize reducing the performance disparity between models trained on synthetic vs. real data \cite{santangelo2025good}.

\begin{equation}
\text{UtilityGap} = \left| F1_{(\text{train real},\, \text{test real})} 
- F1_{(\text{train synthetic},\, \text{test real})} \right|
\label{eq:diffF1}
\end{equation}


Tables \ref{tab:ctgan-results} and \ref{tab:tvae-results} report the \emph{best} hyperparameters (i.e., those minimizing the utility gap over 50 Optuna trials) for each model-dataset pair. F1 values are omitted for brevity; the configurations below correspond to the minimum observed gap in each experiment.

\begin{table}[t]
\centering
\scriptsize
\setlength{\tabcolsep}{1.8pt}
\renewcommand{\arraystretch}{0.9}
\caption{Optimal CTGAN hyperparameters per dataset. 
“LR (G/D)” lists generator/discriminator learning rates. 
“Hidden dims (G/D)” lists hidden layer widths.}
\label{tab:ctgan-results}
\begin{tabularx}{\columnwidth}{@{}l c cc cc c c@{}}
\toprule
\multirow{2}{*}{\textbf{Dataset}} &
\multirow{2}{*}{\makecell{\textbf{Emb.}\\\textbf{dim}}} &
\multicolumn{2}{c}{\textbf{LR}} &
\multicolumn{2}{c}{\makecell{\textbf{Hidden}\\\textbf{dims}}} &
\multirow{2}{*}{\textbf{Batch}} &
\multirow{2}{*}{\textbf{Epochs}} \\
\cmidrule(lr){3-4}\cmidrule(l){5-6}
& & \textbf{G} & \textbf{D} & \textbf{G} & \textbf{D} & & \\
\midrule
Credit Card & 128 & \num{5.9e-5}  & \num{1.98e-4} & 256–256 & 512–512 & 500 & 1500 \\
Cardio      & 512 & \num{1.74e-4} & \num{3.73e-4} & 256–256 & 256–256 & 100 & 1000 \\
Adult       & 128 & \num{4.1e-5}  & \num{1.52e-4} & 256–512 & 512–512 & 500 & 1000 \\
\bottomrule
\end{tabularx}
\end{table}

\begin{table}[t]
\centering
\scriptsize
\setlength{\tabcolsep}{3pt}
\renewcommand{\arraystretch}{0.9}
\caption{Optimal TVAE hyperparameters per dataset. 
``compress\_dims'' and ``decompress\_dims'' denote encoder and decoder layer sizes.}
\label{tab:tvae-results}
\begin{tabularx}{\linewidth}{@{}l c c c c c@{}}
\toprule
\textbf{Dataset} & \textbf{Embedding} & \textbf{Compressions} & \textbf{Decompressions} & \textbf{Batch} & \textbf{Epochs} \\
\midrule
Credit Card & 256 & {[}512, 512{]} & {[}512, 512{]} & 1000 & 1000 \\
Cardio      & 256 & {[}256, 512{]} & {[}256, 512{]} & 1000 & 1500 \\
Adult       & 128 & {[}256, 512{]} & {[}256, 512{]} & 1000 & 1000 \\
\bottomrule
\end{tabularx}
\end{table}

Across datasets, CTGAN configurations commonly favor a higher discriminator learning rate than generator learning rate, and deeper discriminators {[}512, 512{]} for Adult and Credit Card. Longer training (1500 epochs) was selected for CTGAN on Credit Card and TVAE on Cardio, while TVAE consistently preferred a large batch size (1000). These settings reflect capacity and optimization regimes that yielded the smallest utility gap under our search space and training budget.


\subsection{Baseline Utility and Local Geometry}
\label{sec:baseline_utility}

\begin{figure*}[t]
\centering
\includegraphics[width=0.6\textwidth]{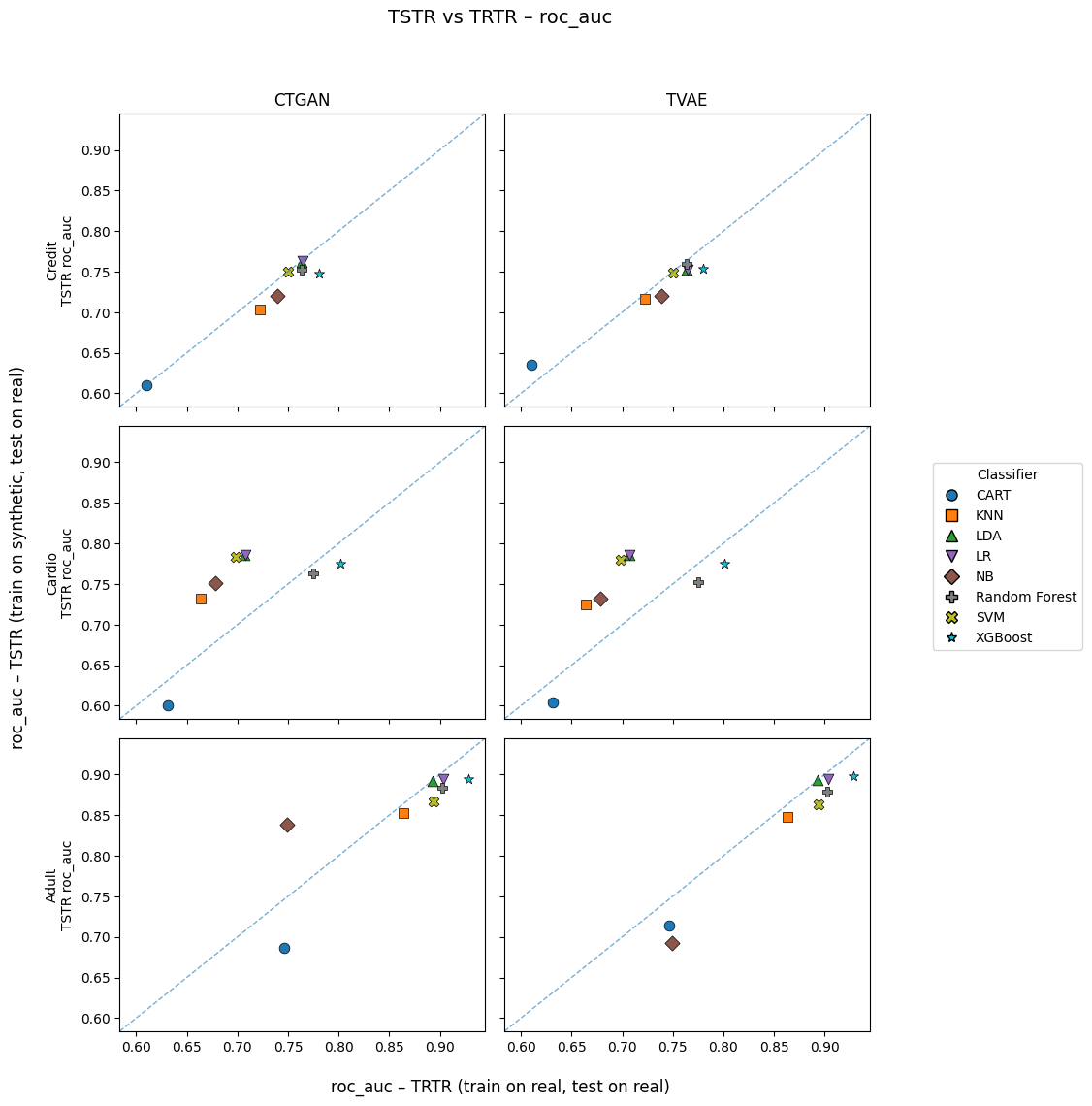}
\caption{\textbf{Baseline downstream utility of the unfiltered CTGAN and TVAE samples, evaluated by ROC--AUC under the TRTR and TSTR protocols}. Columns correspond to generators (left: CTGAN, right: TVAE) and rows to datasets (top: Credit, middle: Cardio, bottom: Adult). In each panel, every marker denotes one classifier from the suite (see legend). The horizontal axis shows ROC--AUC when the model is trained and tested on real data (TRTR), while the vertical axis shows ROC--AUC when the same model is trained on synthetic data and tested on real data (TSTR). The dashed diagonal marks parity between TSTR and TRTR; points below (above) this line indicate loss (gain) in ROC--AUC when training on synthetic instead of real data.}
\label{fig:baseline_utility}
\end{figure*}

We first evaluate the downstream classification utility of the baseline CTGAN and TVAE samples using the TRTR and TSTR protocols from Section~III-D. For each dataset we train a suite of eight classifiers (CART, KNN, LDA, Logistic Regression, Na\"ive Bayes, Random Forest, SVM, XGBoost) either on real training data and test on real data (TRTR) or on synthetic training data and test on real data (TSTR). Figure~\ref{fig:baseline_utility} summarizes ROC--AUC on the common real test set: each point corresponds to one classifier and the dashed line marks parity between TSTR and TRTR performance. Across all datasets and both generators, the vast majority of points lie close to the diagonal, indicating that models trained on synthetic data achieve ROC--AUC that is generally comparable to---and occasionally slightly higher than---models trained directly on real records.

\paragraph{Credit.}

On the Credit Default task, both CTGAN and TVAE produce synthetic data that yield ROC--AUC values close to the TRTR baseline. For CTGAN, LDA, Logistic Regression, and SVM lie almost on the diagonal: their TSTR ROC--AUC differs from TRTR by only a few percent. KNN, Na\"ive Bayes, Random Forest, and XGBoost show somewhat larger gaps (TSTR AUC typically lower by up to \(\approx 3\text{–}4\%\)), but still remain in the \(0.70\text{–}0.78\) range. TVAE exhibits a very similar pattern; for some models (e.g., CART), TSTR AUC is slightly \emph{higher} than TRTR, while for tree ensembles and Na\"ive Bayes it is slightly lower. Overall, synthetic credit samples from both generators support robust discrimination with modest utility loss relative to real-trained models.

\paragraph{Adult.}

The Adult income task has the highest baseline ROC--AUC, with TRTR models typically achieving values between about $0.75$ and $0.93$ across classifiers. When trained on CTGAN samples, downstream models largely preserve this level of performance: points cluster tightly around the parity line, with deviations usually below \(\approx 2\text{–}3\%\). Logistic Regression and LDA are particularly stable, achieving nearly identical ROC--AUC under TSTR and TRTR, indicating that the main linear decision structure of the Adult dataset is well captured by the generator. TVAE follows the same qualitative trend; the TSTR ROC--AUC for XGBoost and SVM is a few percent below the TRTR baseline, while simpler models remain very close. Importantly, the ranking of classifiers (XGBoost and Logistic Regression at the top; CART at the bottom) is preserved for both generators, suggesting that synthetic Adult data induce similar relative decision boundaries as real data.

\paragraph{Cardio.}

For the Cardio dataset, TRTR ROC--AUC scores are more moderate overall (roughly $0.63$--$0.80$), but the effect of synthetic training differs from the previous datasets. Under both CTGAN and TVAE, several classifiers---most notably KNN, LDA, Logistic Regression, Na\"ive Bayes, and SVM---achieve \emph{higher} ROC--AUC when trained on synthetic data than when trained on the real training set, with gains on the order of \(5\text{–}8\%\) for some linear models. In contrast, tree-based methods (CART, Random Forest, XGBoost) experience small decreases in AUC relative to TRTR.

To better understand this behavior and motivated by prior work that evaluates generative models through nearest-neighbour distances and local density estimates~\cite{alaa2022faithful,kansal2023evaluating,zhou2023exploration,li2023improving,yuan2025generative,figueira2022survey}, we quantify the ``smoothness'' of each synthetic dataset by computing the mean and variance of a simple $k$-nearest-neighbor (kNN) density proxy. After standardizing the numeric attributes, we approximate the local density of each point by the inverse of its mean distance to the $k=20$ nearest \emph{real} neighbors. Table~\ref{tab:local_density} reports these statistics for real data and for the corresponding CTGAN/TVAE samples. For Cardio, the mean local density of CTGAN/TVAE samples remains on the same scale as that of the real data, but the variance of local density is reduced by roughly a factor of five. In other words, synthetic Cardio records inhabit neighborhoods that are less heterogeneous: extremely dense clusters and very sparse outliers are strongly attenuated. For Credit and Adult we also observe a reduction in density variance, but the shrinkage is much milder (around a factor of two to three), so the smoothing effect is most pronounced on Cardio.

\begin{table}[t]
\centering
\caption{$k$-NN local-density statistics on numeric features. For each dataset and source, we report the mean and variance of the density proxy $1 / \bar{d}_k$, where $\bar{d}_k$ is the mean Euclidean distance to the $k=20$ nearest real neighbors in standardized numeric space.}
\label{tab:local_density}
\begin{tabular}{llcc}
\hline
Dataset & Source & Mean density & Variance \\
\hline
Cardio & Real   & 7.55 & 30.70 \\
       & CTGAN & 6.21 &  6.26 \\
       & TVAE  & 6.29 &  6.24 \\
Credit & Real   & 3.33 &  7.51 \\
       & CTGAN & 2.34 &  4.25 \\
       & TVAE  & 2.12 &  2.40 \\
Adult  & Real   & 6.14 & 27.38 \\
       & CTGAN & 4.84 & 18.04 \\
       & TVAE  & 4.94 & 18.57 \\
\hline
\end{tabular}
\end{table}

This geometric picture is consistent with the utility pattern in Figure~\ref{fig:baseline_utility}. By smoothing sharp local irregularities in the real Cardio distribution while approximately preserving global density levels, the generators provide synthetic training sets that behave like a regularized or mildly augmented version of the original data: margin- and distance-based classifiers can fit simpler decision boundaries that generalize better on the held-out real test set, leading to TSTR ROC--AUC that exceeds TRTR for several models. At the same time, the strong reduction in local-density variability suggests that rare or extreme numeric patterns may be under-represented in the synthetic Cardio samples. We therefore interpret these results as evidence of partial mode compression in the numeric feature space---a form of smoothing that is beneficial for average classification performance on this task, but that may still compromise coverage of very low-density regions, such as subtle minority subpopulations or borderline cases, which can be penalizing in applications where faithfully capturing the full range of outcomes or risk profiles is critical.

To sumamrize, across all dataset--generator pairs, the ROC--AUC points in Figure~\ref{fig:baseline_utility} concentrate near the identity line, with no catastrophic failures under TSTR. CTGAN and TVAE thus provide baseline synthetic datasets whose ROC--AUC on held-out real data is generally comparable to, and in the case of Cardio sometimes noticeably better than, that of models trained directly on real records. The kNN-based local-density analysis in Table~\ref{tab:local_density} indicates that this high utility can coexist with substantial smoothing of the real data geometry, especially on Cardio, where the variance of local density is strongly reduced. Subsequent sections examine how the proposed post-processing steps modify this baseline trade-off between fidelity, utility, and the preservation of fine-grained structure relevant for privacy and downstream decision-making.

\subsection{Effect of Rejection-with-Replacement Filtering on Univariate Resemblance}

\begin{figure*}[t]
  \centering
  \subfloat[Credit–CTGAN]{%
    \includegraphics[width=0.32\linewidth]{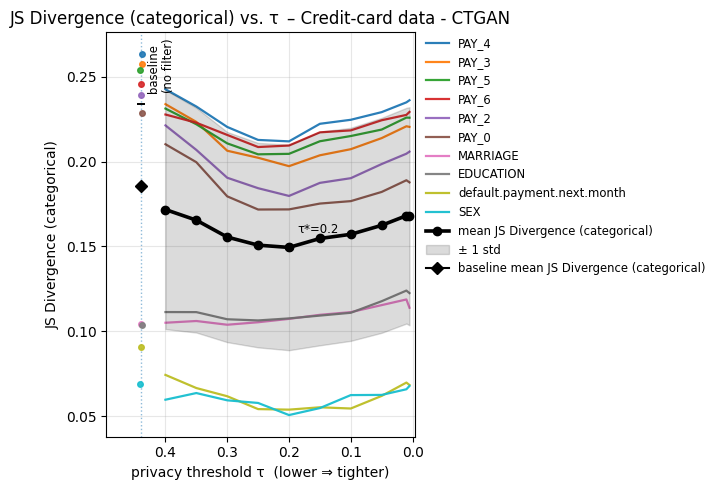}}
  \hfill
  \subfloat[Credit–TVAE]{%
    \includegraphics[width=0.32\linewidth]{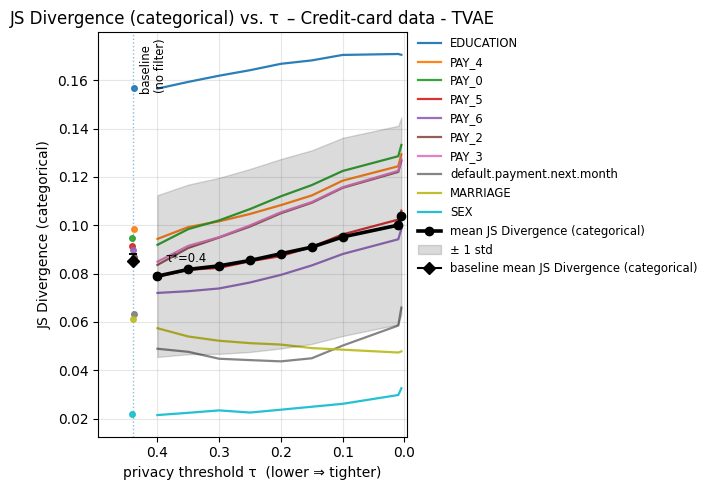}}
  \hfill
  \subfloat[Adult–CTGAN]{%
    \includegraphics[width=0.32\linewidth]{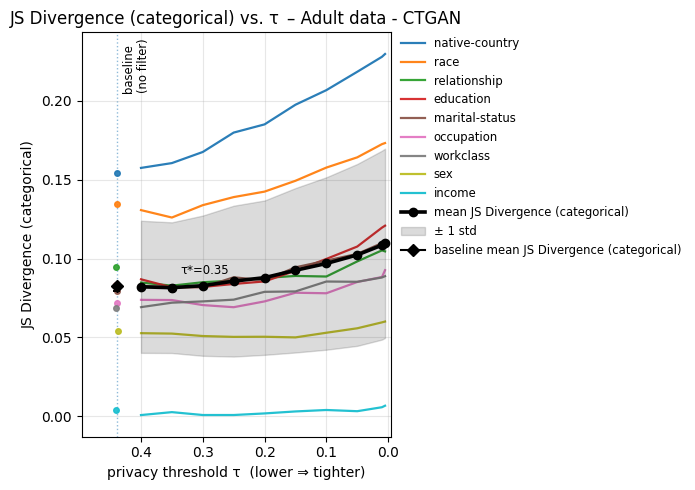}}

  \vspace{1em} 

  \subfloat[Adult–TVAE]{%
    \includegraphics[width=0.32\linewidth]{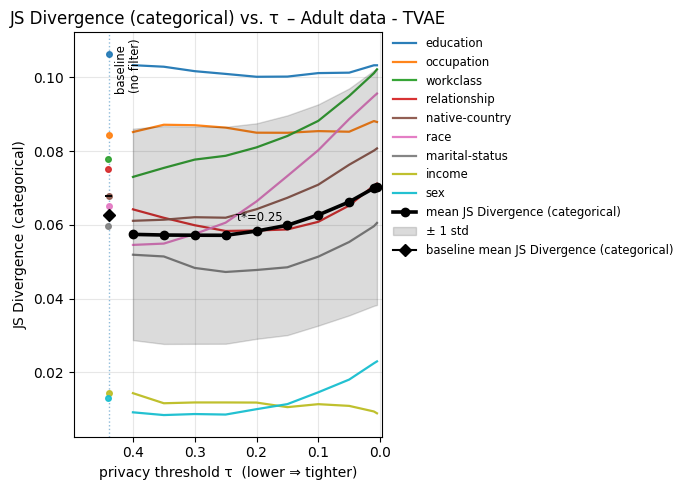}}
  \hfill
  \subfloat[Cardio–CTGAN]{%
    \includegraphics[width=0.32\linewidth]{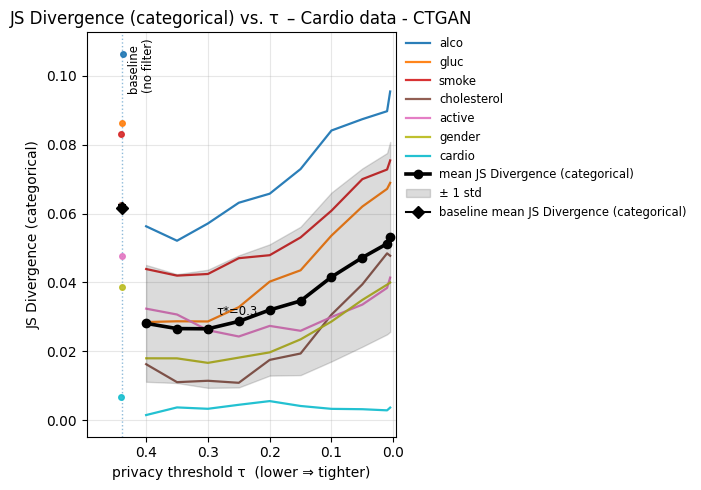}}
  \hfill
  \subfloat[Cardio–TVAE]{%
    \includegraphics[width=0.32\linewidth]{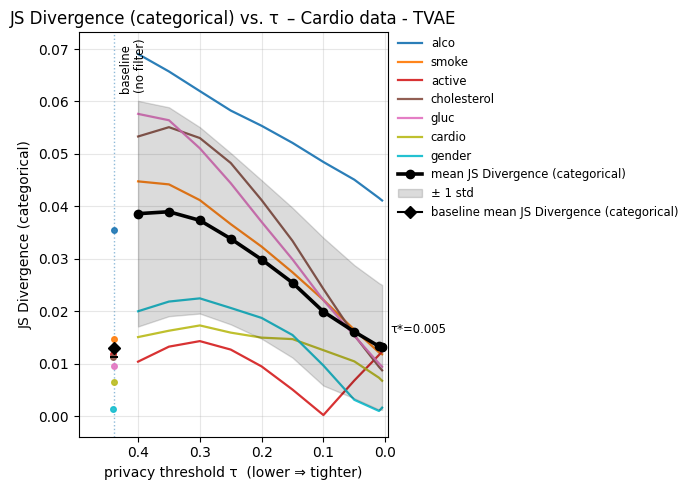}}

 \caption{\textbf{Privacy–utility trade‑off under HEOM--kNN $\widehat{\varepsilon}_{\mathrm{ANY}}$ filtering.}
For each dataset–generator pair we post‑process $G$’s samples with Alg.~\ref{alg:heom-any}, varying the target bound $\tau\equiv\tau_{\mathrm{ANY}}$ (smaller $\Rightarrow$ tighter privacy). The $y$‑axis reports the Jensen–Shannon (JS) divergence between real and synthetic categorical marginals. Thin colored curves are per‑attribute JS; the thick black curve is the mean across attributes; the gray band is $\pm 1$\,s.d.; the dotted vertical line and black diamond mark the unfiltered baseline (no rejection). The annotated $\tau^\star$ in each panel is the value that minimizes the mean JS for that setting. Panels (A–F), following the order in the figure: \textit{Credit}–CTGAN ($\tau^\star\!\approx\!0.2$), \textit{Credit}–TVAE ($\tau^\star\!\approx\!0.4$), \textit{Adult}–CTGAN ($\tau^\star\!\approx\!0.35$), \textit{Adult}–TVAE ($\tau^\star\!\approx\!0.25$), \textit{Cardio}–CTGAN ($\tau^\star\!\approx\!0.3$), and \textit{Cardio}–TVAE ($\tau^\star\!\approx\!5\times10^{-3}$).
Overall, enforcing a tighter $\tau$ can either improve (e.g., \textit{Cardio}–TVAE) or degrade (e.g., \textit{Credit}–TVAE) categorical utility, depending on the generator and dataset.}

  \label{fig:js-distance}
\end{figure*}

\begin{figure*}[t]
  \centering
  \subfloat[Credit–CTGAN]{%
    \includegraphics[width=0.32\linewidth]{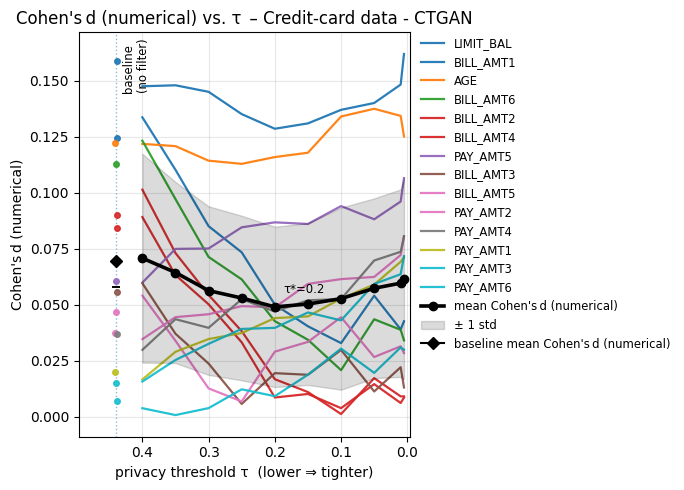}}
  \hfill
  \subfloat[Credit–TVAE]{%
    \includegraphics[width=0.32\linewidth]{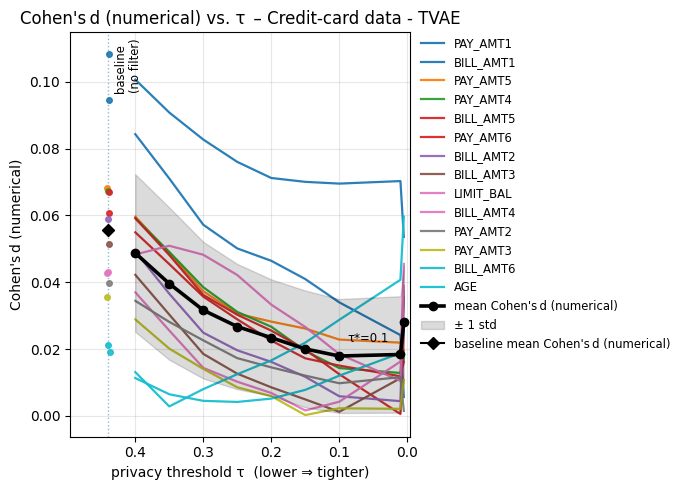}}
  \hfill
  \subfloat[Adult–CTGAN]{%
    \includegraphics[width=0.32\linewidth]{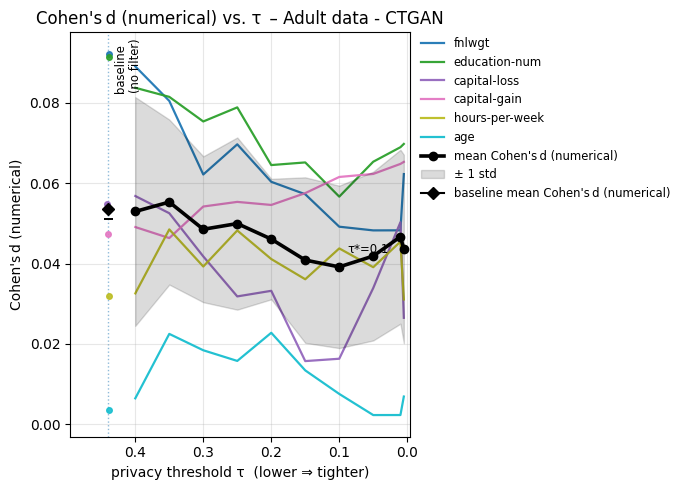}}

  \vspace{1em} 

  \subfloat[Adult–TVAE]{%
    \includegraphics[width=0.32\linewidth]{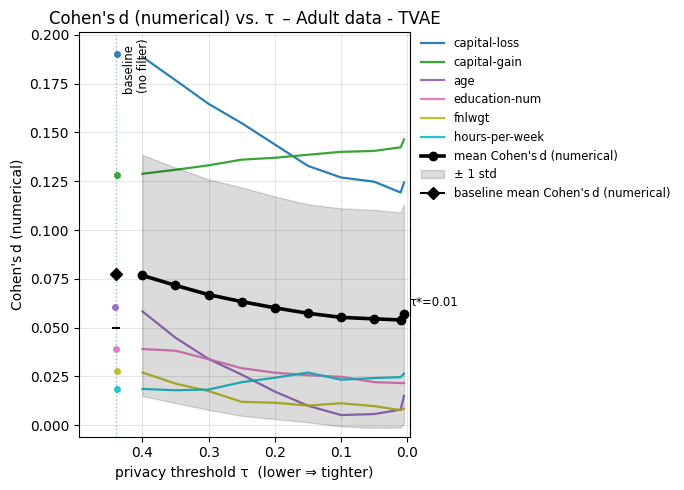}}
  \hfill
  \subfloat[Cardio–CTGAN]{%
    \includegraphics[width=0.32\linewidth]{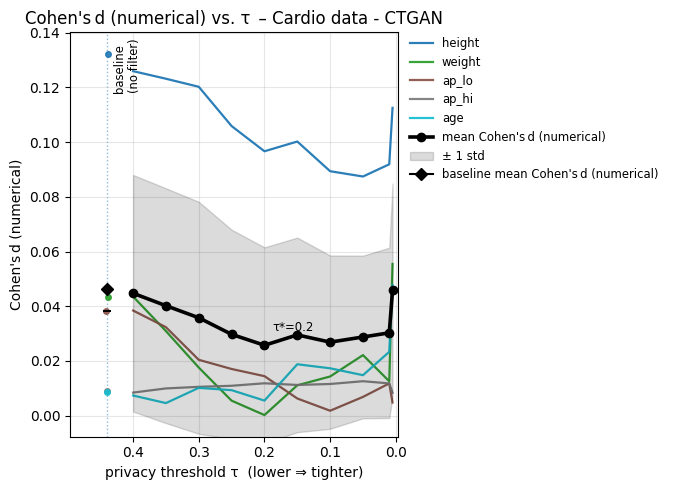}}
  \hfill
  \subfloat[Cardio–TVAE]{%
    \includegraphics[width=0.32\linewidth]{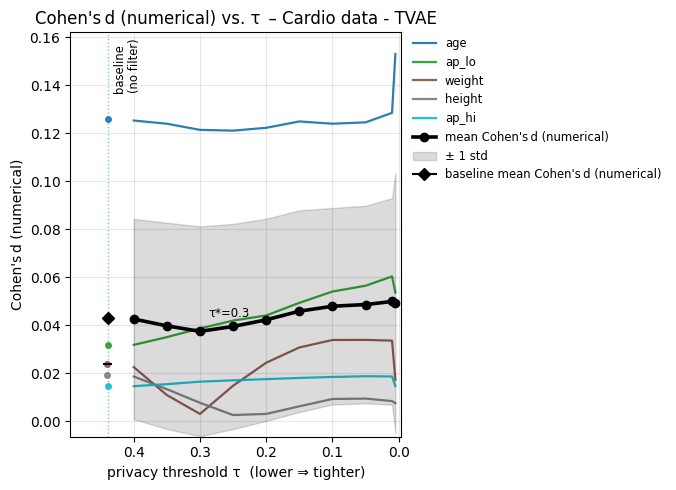}}

\caption{\textbf{Numerical feature fidelity under HEOM--kNN \texorpdfstring{$\widehat{\varepsilon}_{\mathrm{ANY}}$}{eps\_ANY} filtering.}
For each dataset–generator pair we post‑process $G$’s samples with Alg.~\ref{alg:heom-any}, sweeping the target bound $\tau\equiv\tau_{\mathrm{ANY}}$ (smaller $\Rightarrow$ tighter privacy).
The $y$‑axis reports the absolute Cohen’s $d$ (standardized mean difference) between real and synthetic marginals for each numerical attribute (lower is better).
Thin colored curves are per‑attribute Cohen's~$d$; the thick black curve is the mean across attributes; the gray band shows $\pm 1$\,s.d.; the dotted vertical line and black diamond mark the unfiltered baseline (no rejection). 
The annotated $\tau^\star$ in each panel is the value that minimizes the mean Cohen's~$d$.
Panels (A–F), following the order in the figure: \textit{Credit}–CTGAN ($\tau^\star\!\approx\!0.2$), \textit{Credit}–TVAE ($\tau^\star\!\approx\!0.1$), \textit{Adult}–CTGAN ($\tau^\star\!\approx\!0.2$), \textit{Adult}–TVAE ($\tau^\star\!\approx\!0.01$), \textit{Cardio}–CTGAN ($\tau^\star\!\approx\!0.2$), and \textit{Cardio}–TVAE ($\tau^\star\!\approx\!0.3$).
Overall, moderate filtering typically reduces average effect size, whereas overly tight thresholds can increase it, illustrating the privacy–utility trade‑off.}

  \label{fig:cohen-distance}
\end{figure*}

We study the ANY-radius filter (Algorithm~\ref{alg:heom-any}) as a post-sampling mechanism applied to fixed, pre-trained generators (CTGAN/TVAE). We ask whether stricter thresholds (smaller $\tau_{\mathrm{ANY}}$) can improve distance-based privacy proxies while preserving statistical fidelity. For each feature, we compare the marginal distributions between the real ($R$) and synthetic ($S$) data. we report the \emph{Jensen--Shannon distance} for the categorical variables and the absolute Cohen’s~$d$ for the numerical varibales. Unless noted otherwise, we express changes as percentages relative to the unfiltered generator baseline (negative = improvement). When feasible under the filter, the achieved risk satisfies $\widehat{\varepsilon}_{\mathrm{ANY}}(S) < \tau_{\mathrm{ANY}}$.

Across dataset–model pairs, the effect depends on both dataset and generator (Figure \ref{fig:js-distance}, Figure \ref{fig:cohen-distance}). For CTGAN, moderate strictness typically improves univariate fidelity: on \emph{Credit} and \emph{Cardio}, JS decreases by $\approx$20–57\% and Cohen's~$d$ by $\approx$20–44\% up to $\tau_{\mathrm{ANY}}\!\approx\!0.2$–0.3, then degrades under tighter thresholds. On \emph{Adult}, gains are modest with best $\tau_{\mathrm{ANY}}$ around 0.30–0.35. For TVAE, \emph{Credit} and \emph{Adult} exhibit a more monotone privacy–utility trade‑off: looser privacy (larger $\tau_{\mathrm{ANY}}$) yields best fidelity; stricter thresholds degrade both JS and Cohen's~$d$. \emph{Cardio} is an exception: numeric Cohen's~$d$ improves at $\tau_{\mathrm{ANY}}\!\approx\!0.1$ ($-24\%$) while categorical JS is worst near 0.35 but recovers at very strict settings.

Feature‑level observations: low‑cardinality or balanced categoricals (e.g., \texttt{sex}, \texttt{income}) remain stable across threshold levels, whereas high‑cardinality or skewed attributes (e.g., \texttt{race}, \texttt{native-country}, \texttt{PAY\_k}) are more sensitive. Heavy‑tailed numerics such as \texttt{BILL\_AMT*} (\emph{Credit}) and \texttt{capital-gain}/\texttt{loss} (\emph{Adult}) are most volatile under tight filtering.

\paragraph*{Hypothesis - Univariate fidelity.}
Across the credit, Adult, and Cardio tasks, the HEOM–kNN post-filter (threshold $\tau$) acts as a density-aware de-duplication step: it preferentially rejects candidates lying near dominant categorical levels and numeric centers. Tightening $\tau$ reduces near-duplicates and center shifts but systematically displaces probability into sparser tails, which can inflate categorical divergence when the base generator is already well calibrated. In practice, moderate thresholds help when the base model produces inflated centers or near-duplicates (typical for \textit{CTGAN}), whereas conservative filtering—if any—is preferable when the base marginals are already close to the real data (typical for \textit{TVAE}); overly tight targets generally re-inflate tails and increase categorical JS.

\noindent\textit{Model- and dataset-specific pattern.}
For \textbf{CTGAN}, moderate filtering curbs near-duplicates without distorting tails (credit sweet spot at $\tau\!\approx\!0.25$; Cardio stabilizes around $\tau\!\approx\!0.30$–$0.35$), while pushing $\tau$ too low re-weights mass toward minority/rare categories and slightly harms lower-tail coverage on Adult with little gain for already-stable numeric centers. For \textbf{TVAE}, whose marginals are well calibrated, filtering mostly pushes mass away from dense regions and increases categorical drift (credit: \emph{EDUCATION}; Adult: \emph{education}, \emph{workclass}, \emph{relationship}); looser acceptance (\(\tau\!\approx\!0.30\!-\!0.40\) on Adult) preserves strengths—\textbf{lowest categorical JS}, near-zero center shifts, and smaller tail deltas—whereas very tight targets trade modest numeric improvements for \textbf{higher categorical JS} (Cardio) and heavier tail deltas (Adult, capital-loss). \emph{Takeaway:} tune $\tau$ conservatively for TVAE (often mild or none), and use a moderate $\tau$ for CTGAN; avoid very tight targets in all cases.


\begin{figure*}[t]
  \centering
  \begin{subfigure}[t]{0.9\textwidth}
    \includegraphics[width=\linewidth]{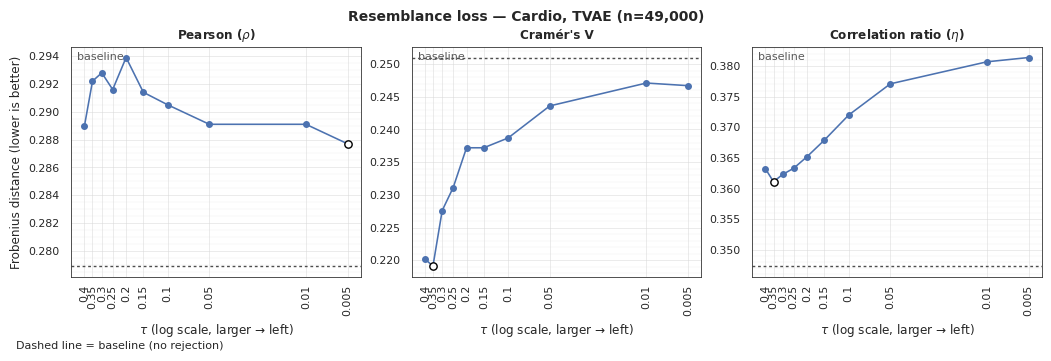}
    \caption{Cardio—TVAE}
    \label{fig:cardio-mv:tvae}
  \end{subfigure}\hfill
  \begin{subfigure}[t]{0.9\textwidth}
    \includegraphics[width=\linewidth]{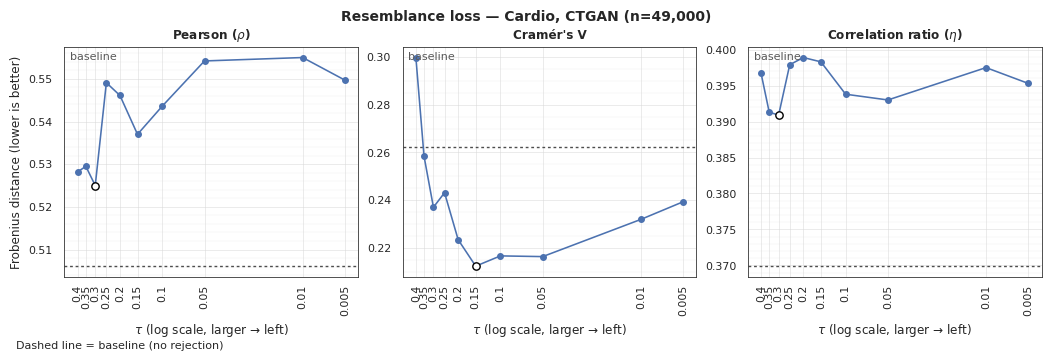}
    \caption{Cardio—CTGAN}
    \label{fig:cardio-mv:ctgan}
  \end{subfigure}
  \caption{\textbf{Cardio: Multivariate resemblance loss under $\tau_{\mathrm{ANY}}$ filtering.}
  Each panel (left$\rightarrow$right) summarizes \emph{family-specific dependence matrices} for numeric (Pearson’s $\rho$), categorical (Cramér’s $V$), and mixed (correlation ratio $\eta^2$) features.
  Curves report Frobenius distance between real and synthetic matrices (lower is better) across thresholds of $\tau_{\mathrm{ANY}}$ (x‑axis on a log scale with larger $\tau$ toward the left).}
  \label{fig:cardio-mv}
\end{figure*}

\begin{figure*}[t]
  \centering
  \begin{subfigure}[t]{0.9\textwidth}
    \includegraphics[width=\linewidth]{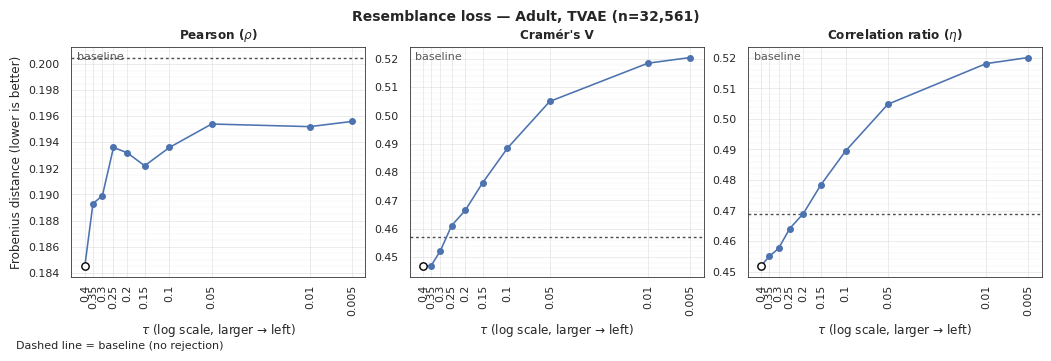}
    \caption{Adult—TVAE}
    \label{fig:adult-mv:tvae}
  \end{subfigure}\hfill
  \begin{subfigure}[t]{0.9\textwidth}
    \includegraphics[width=\linewidth]{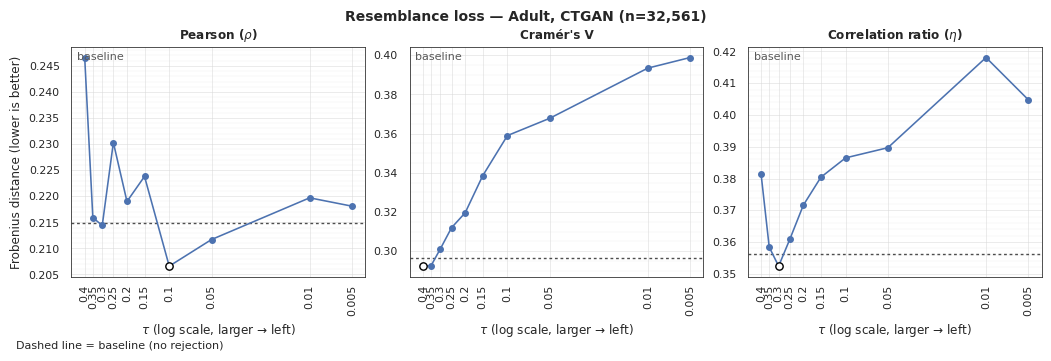}
    \caption{Adult—CTGAN}
    \label{fig:adult-mv:ctgan}
  \end{subfigure}
  \caption{\textbf{Adult: Multivariate resemblance loss under $\tau_{\mathrm{ANY}}$ filtering.} Organization and conventions match Fig.~\ref{fig:cardio-mv}.}
  \label{fig:adult-mv}
\end{figure*}

\begin{figure*}[t]
  \centering
  \begin{subfigure}[t]{0.9\textwidth}
    \includegraphics[width=\linewidth]{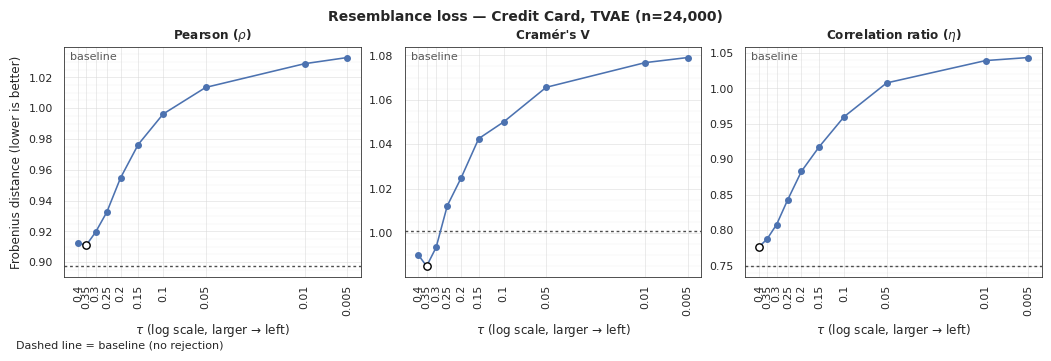}
    \caption{Credit—TVAE}
    \label{fig:credit-mv:tvae}
  \end{subfigure}\hfill
  \begin{subfigure}[t]{0.9\textwidth}
    \includegraphics[width=\linewidth]{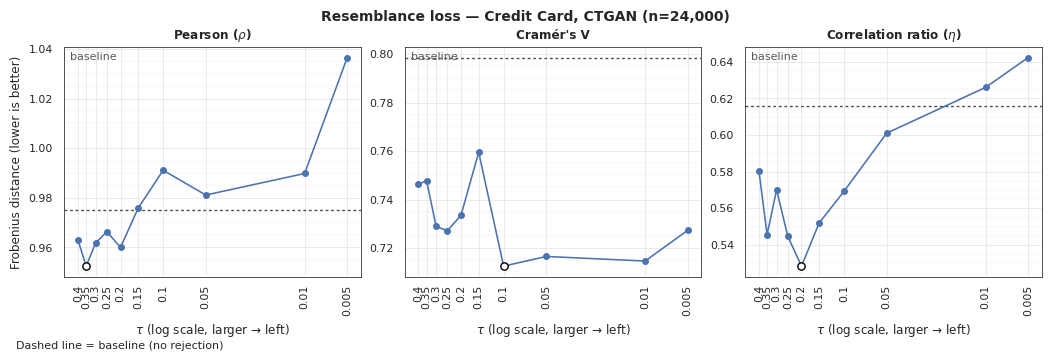}
    \caption{Credit—CTGAN}
    \label{fig:credit-mv:ctgan}
  \end{subfigure}
  \caption{\textbf{Credit: Multivariate resemblance loss under $\tau_{\mathrm{ANY}}$ filtering.} Organization and conventions match Fig.~\ref{fig:cardio-mv}.}
  \label{fig:credit-mv}
\end{figure*}

\subsection{Multivariate Resemblance under $\tau_{\mathrm{ANY}}$ Filtering}

We next examine structural resemblance using family-specific dependence matrices for numeric (Pearson's $\rho$), categorical (Cramér's $V$), and mixed (correlation ratio $\eta^2$) features (see Figs.~\ref{fig:cardio-mv}--\ref{fig:credit-mv}). We report magnitude via Frobenius norm (lower is better) and structure via Spearman rank correlation (higher is better). When feasible under the filter, the achieved risk satisfies $\widehat{\varepsilon}_{\mathrm{ANY}}(S) < \tau_{\mathrm{ANY}}$. Best-by-family results across datasets and models are summarized in Table~\ref{tab:best-by-family}.

On \emph{Credit}, CTGAN benefited modestly from filtering (e.g., numeric Frobenius reduced by $\approx$2.3\% at $\tau_{\mathrm{ANY}}{=}0.35$, categorical $V$ and mixed $\eta^2$ improved by $\approx$10.8\% and $\approx$14.2\% at $\tau_{\mathrm{ANY}}{=}0.10$ and $0.20$; Fig.~\ref{fig:credit-mv}). TVAE’s multivariate metrics were largely flat or best at baseline, with small categorical gains at loose thresholds. On \emph{Adult}, both models achieved small Frobenius improvements; TVAE showed a larger numeric reduction ($\approx$8\% at $\tau_{\mathrm{ANY}}{=}0.40$) and a notable rank‑order boost at stricter settings (e.g., numeric Spearman increased by +0.200 at $\tau_{\mathrm{ANY}}{=}0.10$; Fig.~\ref{fig:adult-mv}), indicating better ordering of correlation strengths even when the magnitude fit peaked at a looser threshold. On \emph{Cardio}, filtering primarily helped categorical structure ($V$ Frobenius: CTGAN $\approx$–19\% at $\tau_{\mathrm{ANY}}{=}0.15$; TVAE $\approx$–12.7\% at $\tau_{\mathrm{ANY}}{=}0.35$; Fig.~\ref{fig:cardio-mv}), while numeric/mixed fidelity was best at baseline.

Overall, moderate $\tau_{\mathrm{ANY}}$ ($\approx$0.2–0.4) often improved categorical structure and, in some cases, numeric structure, but no single threshold optimized all metrics across all blocks. Rank‑based and magnitude‑based criteria can peak at different $\tau_{\mathrm{ANY}}$ values, suggesting that filtering may prune outlier relationships (improving ranks) while slightly shifting correlation magnitudes.

\begin{table*}[t]
\centering
\scriptsize
\setlength{\tabcolsep}{3.5pt}
\renewcommand{\arraystretch}{1.1}
\caption{Best-by-family summary across datasets and models. 
Left: Frobenius norm of the difference between family-specific dependence matrices estimated on real and synthetic data, $\|C_{\text{real}}^{(f)}-C_{\text{synth}}^{(f)}\|_F$ (lower is better). 
Right: Spearman rank correlation between the vectorized entries of the same matrices (higher is better). 
Definitions: $\Delta_{\text{Frob}}=\big(\text{Best}-\text{Baseline}\big)/\text{Baseline}\times100\%$ (negative indicates a reduction); 
$\Delta_{\text{Spearman}}=\text{Best}-\text{Baseline}$. 
Families: $\rho$ (Pearson), $V$ (Cram\'er's $V$), $\eta^2$ (correlation ratio). 
If multiple $\tau_{\mathrm{ANY}}$ attain the same best value, all are listed; “baseline” denotes the unfiltered setting.}
\label{tab:best-by-family}

\begin{minipage}[t]{0.49\textwidth}
\centering
\textbf{Frobenius (lower is better)}\\[2pt]
\begin{tabular}{@{} l l l c c c @{}}
\toprule
Dataset & Model & Family &
\makecell{Baseline\\(Frob)} &
\makecell{Best Frob\\($\tau_{\mathrm{ANY}}$)} &
\makecell{Rel.\ $\Delta_{\text{Frob}}$} \\
\midrule
Adult & CTGAN & $\rho$   & 0.215 & \textbf{0.207} (0.10) & $-3.82\%$ \\
Adult & CTGAN & $V$      & 0.297 & \textbf{0.292} (0.40) & $-1.45\%$ \\
Adult & CTGAN & $\eta^2$ & 0.356 & \textbf{0.352} (0.30) & $-1.04\%$ \\
Adult & TVAE  & $\rho$   & 0.201 & \textbf{0.184} (0.40) & $-7.98\%$ \\
Adult & TVAE  & $V$      & 0.457 & \textbf{0.447} (0.40) & $-2.28\%$ \\
Adult & TVAE  & $\eta^2$ & 0.469 & \textbf{0.452} (0.40) & $-3.67\%$ \\
\midrule
Cardio & CTGAN & $\rho$   & \textbf{0.506} &  0.525 (0.3) & $+3.75\%$ \\
Cardio & CTGAN & $V$      & 0.262 & \textbf{0.212} (0.15) & $-19.0\%$ \\
Cardio & CTGAN & $\eta^2$ & \textbf{0.370} &  0.391 (0.3) & $+5.67\%$ \\
Cardio & TVAE  & $\rho$   & \textbf{0.279} & 0.288 (0.005) & $+3.16\%$ \\
Cardio & TVAE  & $V$      & 0.251 & \textbf{0.219} (0.35) & $-12.7\%$ \\
Cardio & TVAE  & $\eta^2$ & \textbf{0.347} & 0.361 (0.35) & $+3.97\%$ \\
\midrule
Credit Card & CTGAN & $\rho$   & 0.975 & \textbf{0.952} (0.35) & $-2.32\%$ \\
Credit Card & CTGAN & $V$      & 0.799 & \textbf{0.713} (0.10) & $-10.8\%$ \\
Credit Card & CTGAN & $\eta^2$ & 0.616 & \textbf{0.528} (0.20) & $-14.2\%$ \\
Credit Card & TVAE  & $\rho$   & \textbf{0.897} &  0.913 (0.40) & $+1.78\%$ \\
Credit Card & TVAE  & $V$      & 1.000 & \textbf{0.990} (0.40) & $+1.09\%$ \\
Credit Card & TVAE  & $\eta^2$ & \textbf{0.749} &  0.776 (0.40) & $+3.60\%$ \\
\bottomrule
\end{tabular}
\end{minipage}\hfill
\begin{minipage}[t]{0.49\textwidth}
\centering
\textbf{Spearman (higher is better)}\\[2pt]
\begin{tabular}{@{} l l l c c S[table-format=+1.2] @{}}
\toprule
Dataset & Model & Family &
\makecell{Baseline\\Spearman} &
\makecell{Best Spearman\\($\tau_{\mathrm{ANY}}$)} &
$\Delta_{\text{Spearman}}$ \\
\midrule
Adult & CTGAN & $\rho$   & 0.810 & \textbf{0.786} (0.30, 0.005) & -0.0250 \\
Adult & CTGAN & $V$      & 0.913 & \textbf{0.928} (0.35) & +0.0147 \\
Adult & CTGAN & $\eta^2$ & 0.910 & \textbf{0.923} (0.30) & +0.0127 \\
Adult & TVAE  & $\rho$   & 0.436 & \textbf{0.636} (0.10) & +0.200 \\
Adult & TVAE  & $V$      & 0.866 & \textbf{0.897} (0.35) & +0.0306 \\
Adult & TVAE  & $\eta^2$ & 0.849 & \textbf{0.892} (0.15) & +0.0431 \\
\midrule
Cardio & CTGAN & $\rho$   & \textbf{0.418} & 0.406 (0.30, 0.15) & -0.0287 \\
Cardio & CTGAN & $V$      & \textbf{0.810} &  0.774 (0.20) & -0.0360 \\
Cardio & CTGAN & $\eta^2$ & 0.834 & \textbf{0.854} (0.35) & +0.0188 \\
Cardio & TVAE  & $\rho$   & \textbf{0.552} &  0.539 (0.40- 0.25) & -0.0130 \\
Cardio & TVAE  & $V$      & \textbf{0.887} &  0.868 (0.01, 0.005) & -0.0190 \\
Cardio & TVAE  & $\eta^2$ & 0.729 & \textbf{0.741} (0.40) & +0.0123 \\
\midrule
Credit Card & CTGAN & $\rho$   & 0.794 & \textbf{0.808} (0.01) & +0.014 \\
Credit Card & CTGAN & $V$      & 0.941 & \textbf{0.945} (0.40) & +0.0047 \\
Credit Card & CTGAN & $\eta^2$ & 0.937 & \textbf{0.942} (0.10) & +0.0052 \\
Credit Card & TVAE  & $\rho$   & 0.843 & \textbf{0.875} (0.01) & +0.032 \\
Credit Card & TVAE  & $V$      & \textbf{0.889} &  0.883 (0.30) & -0.0060 \\
Credit Card & TVAE  & $\eta^2$ & 0.894 & \textbf{0.895} (0.40) & +0.001\\
\bottomrule
\end{tabular}
\end{minipage}
\end{table*}

Having characterized distributional resemblance, we next assess whether these filtering-induced changes carry over to downstream predictive performance.

\subsection{Downstream Utility under HEOM--$k$NN $\tau_{\mathrm{ANY}}$ Filtering}

For each dataset–generator block (Credit, Adult, Cardio $\times$ CTGAN/TVAE), we train eight classifiers on synthetic data filtered at a grid of $\tau_{\mathrm{ANY}}$ values and evaluate on real data. We summarize cross-classifier means for accuracy, balanced accuracy, weighted $F_1$, and ROC~AUC (bounded metrics; higher is better; units: \%), as well as log-loss (lower is better). Whenever feasible under the filter, the achieved risk satisfies $\widehat{\varepsilon}_{\mathrm{ANY}}(S) < \tau_{\mathrm{ANY}}$.

Across our $\tau_{\mathrm{ANY}}$ grid, cross-classifier means on bounded metrics remained within \emph{$\approx 1\text{–}1.5\%$} of the unfiltered baseline in each block (often under $1\%$), and we did not observe a directional drift in log-loss (Table~\ref{tab:heom-any-xgboost}). Relative to REAL~TRTR, \emph{Credit} was near parity, \emph{Adult} trailed by \(\approx 1.3\text{–}2.6\%\), and \emph{Cardio} exceeded by \(\approx 2\text{–}3\%\); we interpret the latter as a regularization-like effect of synthetic data rather than a causal improvement. Filtered TVAE on \emph{Adult} showed small average gains (up to \(\approx 1\%\)), while \emph{Cardio}/TVAE exhibited a slight mean decrease at strict thresholds (small $\tau_{\mathrm{ANY}}$, \(\le 1\%\)). Classifier “leaderboards” (e.g., XGBoost on \emph{Adult}) were stable across $\tau_{\mathrm{ANY}}$.

In summary, in our runs the ANY filter maintained bounded predictive metrics close to the unfiltered baseline while enabling privacy‑proxy improvements (Sec.~\ref{sec:privacy}). We do not claim zero‑cost privacy in general.

\begin{table}[t]
\centering\small
\caption{\textbf{HEOM--$k$NN $\tau_{\mathrm{ANY}}$ filtering (XGBoost only):}
performance for a single classifier. Each cell reports mean $\pm$ std for XGBoost over 10 repeated evaluations.
REAL and Baseline denote real-data training (TRTR) and unfiltered synthetic training (TSTR), respectively.
Min/Max$_{\tau_{\mathrm{ANY}}}$ are taken over $\tau_{\mathrm{ANY}}$ values using the mean score.}
\label{tab:heom-any-xgboost}
\setlength{\tabcolsep}{4pt}
\resizebox{\columnwidth}{!}{%
\begin{tabular}{lllcccc}
\toprule
Dataset & Generator & Metric & REAL & Baseline & Min$_{\tau_{\mathrm{ANY}}}$ & Max$_{\tau_{\mathrm{ANY}}}$ \\
\midrule
\multirow{5}{*}{Credit} & \multirow{5}{*}{CTGAN}
& accuracy            & 0.822 $\pm$ \std{0.001179} & 0.814 $\pm$ \std{0.00133}  & 0.809 $\pm$ \std{0.001025} & 0.816 $\pm$ \std{0.001404} \\
& & balanced\_accuracy & 0.660 $\pm$ \std{0.00195}  & 0.636 $\pm$ \std{0.002481} & 0.619 $\pm$ \std{0.00166}  & 0.644 $\pm$ \std{0.002878} \\
& & $F_1$ (weighted)   & 0.802 $\pm$ \std{0.001356} & 0.789 $\pm$ \std{0.001695} & 0.779 $\pm$ \std{0.001111} & 0.793 $\pm$ \std{0.002016} \\
& & ROC~AUC            & 0.779 $\pm$ \std{0.00108}  & 0.749 $\pm$ \std{0.00117}  & 0.750 $\pm$ \std{0.0015}   & 0.760 $\pm$ \std{0.000856} \\
& & log-loss           & 0.429 $\pm$ \std{0.000829} & 0.453 $\pm$ \std{0.000817} & 0.447 $\pm$ \std{0.000579} & 0.453 $\pm$ \std{0.001013} \\
\midrule
\multirow{5}{*}{Credit} & \multirow{5}{*}{TVAE}
& accuracy            & 0.822 $\pm$ \std{0.001179} & 0.804 $\pm$ \std{0.001878} & 0.802 $\pm$ \std{0.001313} & 0.810 $\pm$ \std{0.00137} \\
& & balanced\_accuracy & 0.660 $\pm$ \std{0.00195}  & 0.661 $\pm$ \std{0.002974} & 0.660 $\pm$ \std{0.001842} & 0.665 $\pm$ \std{0.001904} \\
& & $F_1$ (weighted)   & 0.802 $\pm$ \std{0.001356} & 0.791 $\pm$ \std{0.00197}  & 0.790 $\pm$ \std{0.001188} & 0.795 $\pm$ \std{0.000696} \\
& & ROC~AUC            & 0.779 $\pm$ \std{0.00108}  & 0.755 $\pm$ \std{0.000717} & 0.746 $\pm$ \std{0.000585} & 0.755 $\pm$ \std{0.000663} \\
& & log-loss           & 0.429 $\pm$ \std{0.000829} & 0.499 $\pm$ \std{0.00121}  & 0.501 $\pm$ \std{0.001061} & 0.516 $\pm$ \std{0.001361} \\
\midrule
\multirow{5}{*}{Adult} & \multirow{5}{*}{CTGAN}
& accuracy            & 0.874 $\pm$ \std{0.000546} & 0.843 $\pm$ \std{0.000816} & 0.840 $\pm$ \std{0.000592} & 0.845 $\pm$ \std{0.000642} \\
& & balanced\_accuracy & 0.799 $\pm$ \std{0.000763} & 0.735 $\pm$ \std{0.001223} & 0.726 $\pm$ \std{0.0021}   & 0.735 $\pm$ \std{0.001403} \\
& & $F_1$ (weighted)   & 0.870 $\pm$ \std{0.000548} & 0.834 $\pm$ \std{0.000798} & 0.830 $\pm$ \std{0.000637} & 0.834 $\pm$ \std{0.000801} \\
& & ROC~AUC            & 0.928 $\pm$ \std{0.000216} & 0.894 $\pm$ \std{0.000424} & 0.891 $\pm$ \std{0.00038}  & 0.895 $\pm$ \std{0.000368} \\
& & log-loss           & 0.275 $\pm$ \std{0.000371} & 0.336 $\pm$ \std{0.000416} & 0.336 $\pm$ \std{0.000587} & 0.341 $\pm$ \std{0.000436} \\
\midrule
\multirow{5}{*}{Adult} & \multirow{5}{*}{TVAE}
& accuracy            & 0.874 $\pm$ \std{0.000546} & 0.849 $\pm$ \std{0.000784} & 0.843 $\pm$ \std{0.000842} & 0.848 $\pm$ \std{0.000491} \\
& & balanced\_accuracy & 0.799 $\pm$ \std{0.000763} & 0.759 $\pm$ \std{0.001183} & 0.745 $\pm$ \std{0.001764} & 0.752 $\pm$ \std{0.001382} \\
& & $F_1$ (weighted)   & 0.870 $\pm$ \std{0.000548} & 0.843 $\pm$ \std{0.000792} & 0.836 $\pm$ \std{0.000968} & 0.841 $\pm$ \std{0.000505} \\
& & ROC~AUC            & 0.928 $\pm$ \std{0.000216} & 0.898 $\pm$ \std{0.000355} & 0.892 $\pm$ \std{0.000486} & 0.898 $\pm$ \std{0.000236} \\
& & log-loss           & 0.275 $\pm$ \std{0.000371} & 0.333 $\pm$ \std{0.000488} & 0.332 $\pm$ \std{0.000432} & 0.340 $\pm$ \std{0.000703} \\
\midrule
\multirow{5}{*}{Cardio} & \multirow{5}{*}{CTGAN}
& accuracy            & 0.739 $\pm$ \std{0.000797} & 0.720 $\pm$ \std{0.001092} & 0.714 $\pm$ \std{0.001016} & 0.722 $\pm$ \std{0.000791} \\
& & balanced\_accuracy & 0.739 $\pm$ \std{0.000797} & 0.720 $\pm$ \std{0.001092} & 0.714 $\pm$ \std{0.001015} & 0.722 $\pm$ \std{0.00079}  \\
& & $F_1$ (weighted)   & 0.739 $\pm$ \std{0.000804} & 0.719 $\pm$ \std{0.001105} & 0.713 $\pm$ \std{0.001003} & 0.721 $\pm$ \std{0.000785} \\
& & ROC~AUC            & 0.801 $\pm$ \std{0.000173} & 0.774 $\pm$ \std{0.000599} & 0.765 $\pm$ \std{0.00098}  & 0.776 $\pm$ \std{0.000383} \\
& & log-loss           & 0.543 $\pm$ \std{0.000205} & 0.574 $\pm$ \std{0.000528} & 0.572 $\pm$ \std{0.000488} & 0.582 $\pm$ \std{0.000919} \\
\midrule
\multirow{5}{*}{Cardio} & \multirow{5}{*}{TVAE}
& accuracy            & 0.739 $\pm$ \std{0.000797} & 0.713 $\pm$ \std{0.00063}  & 0.713 $\pm$ \std{0.00118}  & 0.719 $\pm$ \std{0.001273} \\
& & balanced\_accuracy & 0.739 $\pm$ \std{0.000797} & 0.713 $\pm$ \std{0.000632} & 0.713 $\pm$ \std{0.00118}  & 0.719 $\pm$ \std{0.001273} \\
& & $F_1$ (weighted)   & 0.739 $\pm$ \std{0.000804} & 0.711 $\pm$ \std{0.000689} & 0.712 $\pm$ \std{0.00119}  & 0.719 $\pm$ \std{0.00129}  \\
& & ROC~AUC            & 0.801 $\pm$ \std{0.000173} & 0.775 $\pm$ \std{0.00036}  & 0.772 $\pm$ \std{0.000561} & 0.779 $\pm$ \std{0.000455} \\
& & log-loss           & 0.543 $\pm$ \std{0.000205} & 0.576 $\pm$ \std{0.000254} & 0.570 $\pm$ \std{0.000476} & 0.577 $\pm$ \std{0.000218} \\
\bottomrule
\end{tabular}}
\end{table}

\subsection{Privacy Evaluation under HEOM--$k$NN $\tau_{\mathrm{ANY}}$ Filtering}
\label{sec:privacy}

\begin{table*}[t] 
\centering 
\scriptsize 
\caption{Privacy evaluation across models and datasets under progressively stricter HEOM--$k$NN~$\tau_{\mathrm{ANY}}$ filtering (Baseline~$\rightarrow$~$\tau_{\mathrm{ANY}}{=}0.005$). All percentage differences are reported in \%. Rank correlations ($\rho$) are computed over $\tau_{\mathrm{ANY}}$ values excluding the baseline. Higher $\Delta RPR$ and CAP protection indicate improved distance-based and disclosure-risk privacy, respectively, while lower AIA accuracy or $R^2$ reflects reduced inference success.} 
\label{tab:cross-summary} 

\begin{tabular}{llcccccc} 
\toprule 

Dataset & Gen. & $\Delta RPR$ (\%) & $\rho_{RPR}$ & Median \texttt{cap\_prot} (\%) & $\rho_{CAP}$ & AIA Acc (Base~$\rightarrow$~0.005; $\Delta$\%) & AIA Reg $R^2$ (Base~$\rightarrow$~0.005; $\Delta$) \\ 
\midrule 

Credit Card & CTGAN & +3.19 & --0.99 & 0.530 (+2.1\%) & --0.84 & 48.9\%~$\rightarrow$~50.7\% (+1.8\%) & --0.002~$\rightarrow$~--0.006 (--0.004) \\ 
Credit Card & TVAE & +3.65 & --1.00 & 0.529 (+1.9\%) & --0.98 & 44.6\%~$\rightarrow$~37.8\% (--6.8\%) & --0.007~$\rightarrow$~--0.020 (--0.013) \\ 
Adult & CTGAN & +1.90 & --1.00 & 0.271 (+7.5\%) & --0.96 & 81.7\%~$\rightarrow$~80.5\% (--1.2\%) & 0.051~$\rightarrow$~0.032 (--0.019) \\ 
Adult & TVAE & +3.18 & --1.00 & 0.265 (+10.4\%) & --1.00 & 84.2\%~$\rightarrow$~82.2\% (--2.0\%) & 0.037~$\rightarrow$~0.011 (--0.026) \\ 
Cardio & CTGAN & +4.48 & --0.99 & 0.169 (--3.0\%) & --0.96 & 75.5\%~$\rightarrow$~73.2\% (--2.3\%) & 0.001~$\rightarrow$~--0.002 (--0.003) \\
Cardio & TVAE & +4.54 & --1.00 & 0.195 (--3.0\%) & --1.00 & 78.6\%~$\rightarrow$~74.8\% (--3.8\%) & 0.000~$\rightarrow$~0.001 (+0.001) \\ 

\bottomrule
\end{tabular}
\end{table*}

We evaluate the threshold $\tau_{\mathrm{ANY}}$ from near-unfiltered to strict (0.4 $\downarrow$ 0.005) using: (i) distinct $L$-diversity $L_{\mathrm{distinct}}$ (higher is better), (ii) per-feature CAP protection (\texttt{cap\_protection}; higher is better), (iii) DCR relative proximity ratio $RPR$ (closer to 50\% indicates closely matched values of $DCR_{\text{train}}$ and $DCR_{\text{test}}$), and (iv) attribute inference attacks (AIA; lower attacker success is better). These are \emph{proxies} rather than formal guarantees. When feasible under the filter, the achieved risk satisfies $\widehat{\varepsilon}_{\mathrm{ANY}}(S) < \tau_{\mathrm{ANY}}$.

As $\tau_{\mathrm{ANY}}$ decreases, $RPR$ increased in every block in our runs (monotone trends; Spearman $\rho \approx -0.99$ on average), indicating that stricter filtering pushes synthetic samples farther from training records (Table~\ref{tab:heom-any-xgboost}). Median CAP protection generally improved modestly at strict $\tau_{\mathrm{ANY}}$ ($\approx{+}0.010$ to ${+}0.025$ absolute in four of six blocks), with a slight net decrease on \emph{Cardio}. By contrast, the exploratory $L$-diversity scores were unstable across $\tau_{\mathrm{ANY}}$ (sometimes far exceeding the real data’s diversity, other times dropping), reflecting sensitivity to which specific records are filtered. AIA outcomes remained within run-to-run variation relative to baseline across the threshold grid; we did not observe a measurable reduction in AIA success.

The filter reliably increases distance-based privacy proxies and can modestly improve CAP protection, while $L$-diversity may vary non-monotonically. Under our setting, AIA was not measurably reduced by filtering. These heuristics do not constitute a formal privacy guarantee (e.g., no $(\varepsilon,\delta)$-DP bound).

\subsection{Entropy-Weighted HEOM and $k$-NN Radii (AIA-only)}
\label{sec:ablation-weighted-heom-knn}

We investigate whether Attribute Inference Attack (AIA) outcomes are mechanically driven by the geometry used in the rejection-with-replacement sampler rather than by privacy tightness. To isolate geometry, we ablate two choices: (i) replacing vanilla HEOM with an \emph{entropy-weighted} encoder, and (ii) enlarging the baseline 2-NN radii to general $k$-NN radii. Throughout, the privacy threshold is fixed at $\tau_{\mathrm{ANY}}{=}0.01$, so any change in performance is attributable to weights or radii, not to privacy strength. When feasible, the returned synthetic sets satisfy $\widehat{\varepsilon}_{\mathrm{ANY}}(S) < \tau_{\mathrm{ANY}}$.

First, we implement our \textbf{entropy-weighted HEOM}.
Let $w_c$ be a per-column weight (a single $w_c$ for all one-hot dummies of a categorical feature). We set $w_c \propto 1/(H_c + \varepsilon)$, where $H_c$ is the empirical Shannon entropy~\cite{shannon} (for numerics, $H_c$ comes from histograms with Freedman--Diaconis bins~\cite{freedman1981histogram}). Expanding to the one-hot space yields $w_{\mathrm{dim}} \in \mathbb{R}^d$; features are scaled by $w_{\mathrm{scale}} = \sqrt{w_{\mathrm{dim}}}$, and distances are $\lVert u - v \rVert_w = \lVert (u - v) \odot w_{\mathrm{scale}} \rVert_2$. This upweights low-entropy attributes that tend to concentrate risk.

To enlarge the baseline to \textbf{$k$-NN radii}, we define, for each real record $x_i$, its privacy radius $r_i$ as the distance to its $k$-th nearest \emph{real} neighbor in the weighted space. We use $k_{\mathrm{eff}} = \min(\max(k, 2), n_r)$, where $n_r$ is the number of real records, so larger $k$ produces larger, more conservative balls and generalizes the 2-NN baseline.

We combine these changes into three variants. Variant \textbf{V0} uses unweighted HEOM with 2-NN radii; variant \textbf{V1} uses entropy-weighted HEOM with $k{=}2$; and variant \textbf{V2} uses entropy-weighted HEOM with $k{=}5$.

\begin{figure*}[t]
\centering
\captionsetup[sub]{justification=centering}
\begin{subfigure}{0.44\textwidth}
\centering
\includegraphics[width=\linewidth]{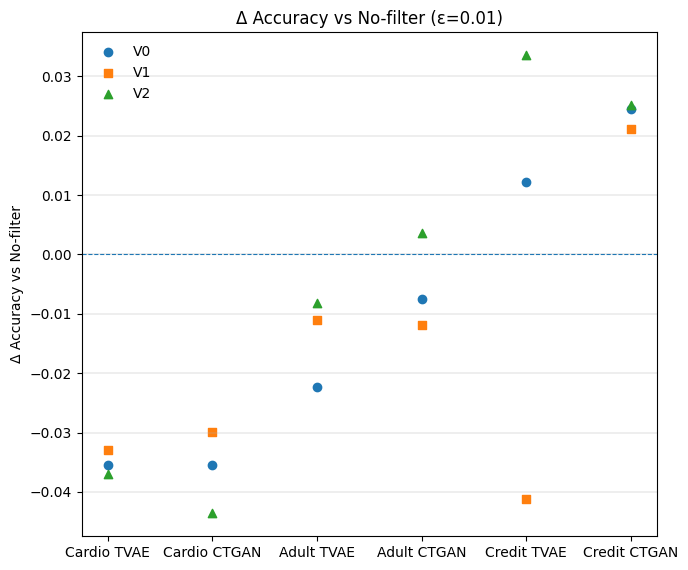}
\caption{$\Delta$Accuracy}
\end{subfigure}\hfill
\begin{subfigure}{0.44\textwidth}
\centering
\includegraphics[width=\linewidth]{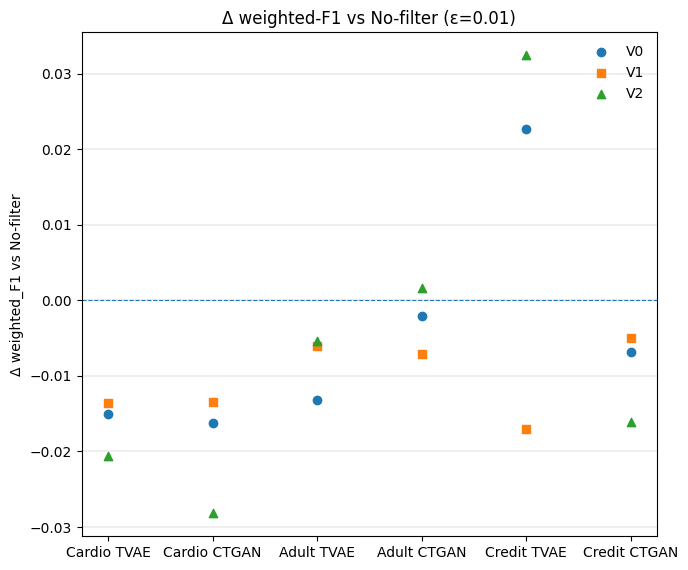}
\caption{$\Delta F_1$}
\end{subfigure}

\vspace{0.6em}

\begin{subfigure}{0.44\textwidth}
\centering
\includegraphics[width=\linewidth]{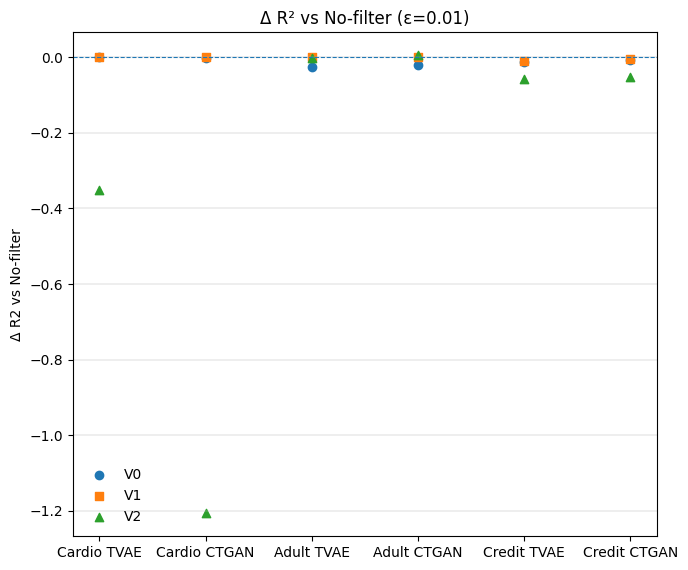}
\caption{$\Delta R^2$}
\end{subfigure}\hfill
\begin{subfigure}{0.44\textwidth}
\centering
\includegraphics[width=\linewidth]{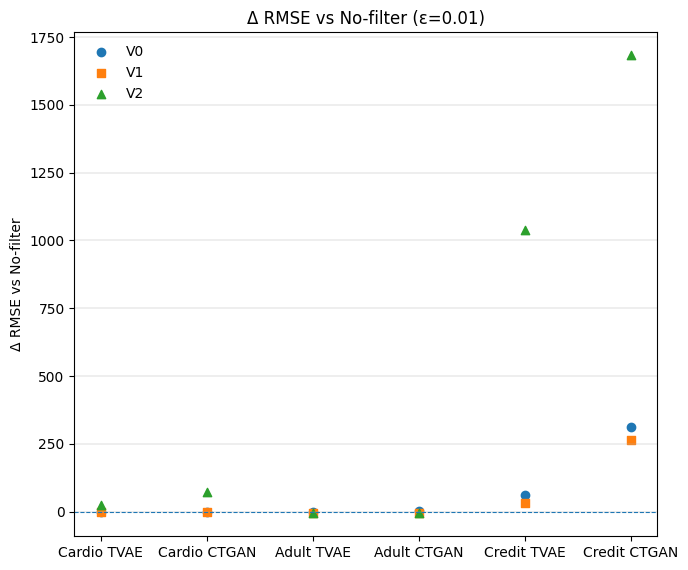}
\caption{$\Delta$RMSE}
\end{subfigure}

\caption{\textbf{Geometry ablations at fixed privacy threshold.} Panels report changes in attribute inference attack (AIA) performance---$\Delta$ Accuracy and $\Delta$ weighted-$F_1$ (classification), and $\Delta R^2$ and $\Delta$ RMSE (regression)---for each dataset--generator pair (Cardio/Adult/Credit $\times$ TVAE/CTGAN). Each point compares a geometry variant to the corresponding \emph{no-filter} baseline for the same dataset--generator pair, with the $\widehat{\varepsilon}_{\mathrm{ANY}}$-based rejection-with-replacement threshold fixed to $\tau_{\mathrm{ANY}}=0.01$ (so differences reflect geometry/weights/radii rather than a tighter threshold). For each metric, $\Delta$ is computed within a dataset--generator pair by first evaluating the metric separately for each attacked attribute and then taking the unweighted mean across attacked attributes; $\Delta$ is the variant mean minus the \emph{no-filter} mean, and the dashed line indicates parity ($\Delta=0$). Variants differ only in filter geometry: \emph{no-filter} disables $\tau_{\mathrm{ANY}}$ filtering; V0 uses unweighted HEOM with radii from real-data $2$-NN (baseline); V1 uses entropy-weighted HEOM with real-data $k$-NN radii ($k=2$); V2 uses entropy-weighted HEOM with $k=5$. Interpretation: $\Delta>0$ implies improved AIA (worse privacy) for Accuracy, weighted-$F_1$, and $R^2$, whereas $\Delta<0$ implies improved AIA for RMSE; RMSE is in task-native units and should not be compared across datasets.}

\label{fig:aia-classification-multiples}
\end{figure*}

We compare V0--V2 to the no-filter sampler (same generator, no $\tau_{\mathrm{ANY}}$-based rejection). For classification tasks, we report $\Delta$Accuracy and $\Delta F_1$; for regression tasks, we report $\Delta R^2$ and $\Delta$RMSE. To summarize across tasks within each dataset--generator pair, we use the median $\Delta$ (robust to outliers).

Our results are presented in  Fig.~\ref{fig:aia-classification-multiples}. They show that classification effects are small across datasets/models (median $\Delta$Accuracy and $\Delta F_1$ near zero). For regression on \emph{Credit} (PAY\_0–PAY\_6), we observe modest median gains (e.g., V2 $\Delta R^2\!\approx\!+0.22$, $\Delta$RMSE$\!\approx\!-0.11$). We do not claim statistical significance.

Overall, our experiments in this section demonstrate that geometry choices at fixed privacy have limited classification impact and modest, generator‑specific regression benefits.

\section{Incidence and repair of categorical mode collapse}
\label{sec:mode-collapse-incidence}

We quantify categorical mode collapse using the detector in Alg.~\ref{alg:detection}, which cross‑tabulates real versus synthetic values and flags any real category with zero count in the synthetic data. Repair is then performed by the layer‑frozen fine‑tuning routine in Alg.~\ref{alg:patch}, which freezes the early layers and adapts only the top layers on slices of the real data corresponding to the missing category (Sec.~\ref{sec:mode-gap-filling}). This design restores support without overwriting previously learned structure. \textit{All resemblance figures for categoricals are reported as the Jensen–Shannon \emph{distance}} (the square root of the JS divergence; metric in $[0,1]$), consistent with our fidelity protocol.

To quantify this, let $\mathcal{C}_j$ be the set of levels of categorical column $j$, and let $\widehat p_R$ and $\widehat p_S$ denote empirical real and synthetic level frequencies.

\begin{itemize}
  \item \textbf{Zero‑coverage rate (ZCR)} for column $j$ is
  $\mathrm{ZCR}_j=\frac{1}{|\mathcal{C}_j|}\sum_{c\in\mathcal{C}_j}\mathbf{1}\{\widehat p_S(X_j=c)=0\}$.
  ``Columns with ZCR$>$0 (pre)’’ counts how many categorical columns have at least one missing level before patching.

  \item \textbf{Total missing‑mass rate (MAR)} is the real probability mass carried by levels that are missing in the synthetic data, aggregated over columns:
  $\mathrm{MAR}=\sum_j\sum_{c\in\mathcal{C}_j:\,\widehat p_S=0}\widehat p_R(X_j=c)$, reported as a percentage. Intuitively, MAR answers: ``What fraction of real records fall into categories that the baseline generator never produces?’’  (Small MAR implies that collapse affects rare levels.)

  \item \textbf{\# Patched levels} is the number of distinct missing levels that are explicitly repaired by Alg.~\ref{alg:patch}.

  \item \textbf{Support‑recovery rate (SRR)} is the fraction of initially missing levels that receive non‑zero synthetic mass after patching: 
  $\mathrm{SRR}=1-\frac{\#\{\text{missing levels post}\}}{\#\{\text{missing levels pre}\}}$.
  $\mathrm{SRR}=1$ means full recovery of categorical support.

  \item $\Delta$\textbf{ mean JSD} is the absolute change in the \emph{mean} per‑column Jensen–Shannon distance (post minus pre); values closer to $0$ indicate negligible shift in average categorical marginal discrepancy. We report the absolute change on the $[0,1]$ JSDistance scale.
\end{itemize}

\begin{table}[!t]
\centering
\scriptsize
\caption{Incidence and repair of categorical mode collapse. 
ZCR detects dropped levels; MAR quantifies the real mass parked in those levels; SRR measures support recovery; 
$\Delta$JSD is the absolute change in average categorical Jensen--Shannon \emph{distance} after patching (lower magnitude is better).}
\label{tab:modepatch-summary}

\setlength{\tabcolsep}{3pt} 
\renewcommand{\arraystretch}{1.1} 
\begin{tabular}{@{}lccccc@{}}
\toprule
Dataset \& Model & ZCR$>$0 (pre) & MAR (\%) & Patched & SRR (post) & $\Delta$JSD \\
\midrule
\textsc{Credit} TVAE & 4 & 0.0583 & 4 & 1.0 & 0.0245 \\
\textsc{Adult}  TVAE & 1 & 0.0031 & 1 & 1.0 & 0.0093 \\
\bottomrule
\end{tabular}
\end{table}

Table~\ref{tab:modepatch-summary} summarizes the incidence of collapse and the effect of repair on the two settings where it was observed (TVAE trained on \textsc{Credit} and \textsc{Adult}). 
Our results show that only the TVAE baselines exhibited categorical mode dropping in our runs: four columns on \textsc{Credit} and one column on \textsc{Adult}. The associated \emph{total} missing mass is minute—$\mathrm{MAR}=0.058\%$ on \textsc{Credit} and $0.003\%$ on \textsc{Adult}—indicating that the missing levels are rare in the real data. This aligns with the detector’s design (Alg.~\ref{alg:detection}) and with prior observations that rare categories are the most fragile under tabular generation. 
Applying the layer‑frozen patching routine (Alg.~\ref{alg:patch}) restored \emph{every} dropped level (\mbox{SRR = 1.0}) with a one‑to‑one mapping between the number of missing and patched levels. Freezing the lower (feature‑extracting) layers while fine‑tuning the head confines updates to a low‑rank subspace and mitigates catastrophic forgetting—precisely the behavior we sought when transferring on a single rare category. 
Because the MAR is so small, global resemblance metrics change little after patching: the average per‑column JSDistance shifts by only $0.0245$ on \textsc{Credit}–TVAE and $0.0093$ on \textsc{Adult}–TVAE. This is expected: we reallocate probability mass from dominant levels into previously unseen, very rare levels, which improves support coverage but barely moves dataset‑level averages. More detailed fidelity behavior (use of JSDistance for categoricals and effect‑size summaries for numerics) is reported in our extended results \ref{app:baseline_resemblance}.

Even when aggregate divergences are nearly unchanged, support recovery has practical value:
(i) it eliminates zero‑probability artifacts that can confound downstream analyses or fairness audits,
(ii) it guarantees the ability to draw any real categorical combination at least once from the synthetic data, and
(iii) it prevents brittle behavior in pipelines that rely on full categorical coverage (e.g., one‑hot encoders fitted on real).
Our results show that the proposed patching achieves these benefits \emph{without} retraining from scratch and with negligible impact on overall resemblance, exactly as intended by the design of the detector and layer‑frozen transfer.

Mode collapse in our experiments was infrequent and low‑mass, but when present it was repaired completely (SRR$=1$) by a short, layer‑frozen fine‑tune targeted to the missing levels. As anticipated for rare categories, the mean categorical JSDistance moved by only a few hundredths, suggesting that mode patching is best viewed as a \emph{support‑completeness} correction that safeguards combinatorial coverage with near‑zero cost to global resemblance.

\section{Cross-Metric Coherence}

The interplay between resemblance, utility, and privacy metrics reveals areas of agreement (where metrics move in tandem) and tension (where improving one metric might come at the cost of another or give a misleading picture). Throughout, ``stricter'' means a \emph{smaller} threshold $\tau_{\mathrm{ANY}}$; when feasible under filtering, the achieved risk satisfies $\widehat{\varepsilon}_{\mathrm{ANY}}(S)<\tau_{\mathrm{ANY}}$.

\subsection{Where Metrics Agree}

\textbf{Moderate $\tau_{\mathrm{ANY}}$ sweet spot}: The clearest consensus is that a moderate privacy threshold ($\tau_{\mathrm{ANY}} \sim 0.1$--$0.3$) often yields a win--win: categorical resemblance improves (lower JS, better $V$) while utility remains unchanged. For instance, at $\tau_{\mathrm{ANY}} = 0.2$, \textit{Credit/CTGAN} saw JS divergence drop $\sim$20\% (Fig~\ref{fig:js-distance}) and $V$ error drop 14\% (Table~\ref{tab:best-by-family}) with essentially no change in accuracy or AUC (Table~\ref{tab:heom-any-xgboost}). All models showed some privacy gain (higher $RPR$) in this range with negligible utility cost. This alignment suggests an ``easy privacy win'' regime where privacy proxies and fidelity strengthen concurrently. The likely reason is that removing duplicates curbs overfitting (boosting resemblance) without sacrificing real signal (so utility remains constant).
    
\textbf{Loose filtering vs.\ TVAE performance:} Another point of agreement is that TVAE’s best fidelity is at loose/no filtering, which is consistent with its utility outcomes. Under stricter thresholds (smaller $\tau_{\mathrm{ANY}}$), TVAE’s resemblance metrics slightly worsen relative to CTGAN, and those stricter settings did not yield utility gains for TVAE (except on \textit{Adult}). In other words, if using TVAE, metrics suggest leaning toward minimal filtering unless privacy requirements demand otherwise. This is because TVAE’s univariate fidelity and correlation structure were strongest at Baseline (especially for \textit{Credit}), and its utility was already stable without filtering. In contrast, CTGAN benefited from moderate filtering—fidelity (and privacy) improved with no utility loss. Thus, metrics indicate that CTGAN can be paired with filtering more profitably than TVAE on certain datasets.
    
\textbf{Balanced accuracy vs.\ Accuracy (in Cardio):} In \textit{Cardio} results, we observe Balanced Accuracy $\approx$ Accuracy across all $\tau_{\mathrm{ANY}}$ settings (Table~\ref{tab:heom-any-xgboost}). This indicates strong coherence in class-wise performance—an agreement between these metrics. The likely reason is class balance in the Cardio dataset (or effective oversampling by the models): if the dataset has roughly equal positive and negative instances, then overall accuracy already reflects balanced performance, so Balanced Accuracy adds little new information. The implication is that no significant class imbalance issues are present; the synthetic data did not introduce or mask any skew. Thus, both metrics tell the same story (indeed, for \textit{Cardio} CTGAN they are numerically identical at 68.9\% (Table~\ref{tab:heom-any-xgboost}). This agreement suggests that our evaluation of classifier utility is not confounded by class distribution shifts—a reassuring sign of fidelity on that aspect.

\subsection{Where Metrics Conflict or Decouple}

\textbf{Univariate vs.\ Multivariate:} We find cases where improvements in marginal distributions do not carry over to multivariate structure. For example, in \textit{Cardio}/CTGAN at moderate $\tau_{\mathrm{ANY}} = 0.20$, the filter dramatically improved the univariate metrics (JS divergence $\downarrow57\%$, Cohen's~$d$ $\downarrow44\%$). However, the numeric correlation error rose by 3--6\%, and the rank-order correlation did not improve (Table~\ref{tab:best-by-family}). The filter made each synthetic marginal distribution closer to the real one (e.g., blood pressure and cholesterol values individually looked more realistic) but disrupted how those features co-vary. For instance, the correlation between systolic and diastolic blood pressure may have been distorted. This cautions that evaluating only one-dimensional distributions (JS or Cohen's~$d$) can overstate fidelity if joint structure is overlooked. In other words, synthetic data could ``get the marginals right'' while breaking the correlations—a conflict here where resemblance metrics disagree.
    
\textbf{Magnitude vs.\ Rank in correlations:} We observed cases where Spearman rank alignment improved even as the Frobenius-norm error worsened. For example, in \textit{Adult}/TVAE (numeric features), at $\tau_{\mathrm{ANY}} = 0.10$ the Spearman correlation jumped by +0.20 (better rank-order agreement) (Table~\ref{tab:best-by-family}), yet the actual correlation magnitudes deviated more at $\tau_{\mathrm{ANY}} = 0.10$ than at $\tau_{\mathrm{ANY}} = 0.40$ (the Frobenius error was larger at 0.10). This decoupling suggests that the filter removed certain outlier relationships (improving the ordering of correlation strengths) but also shrank or reshuffled some correlation values. Which metric is more relevant depends on the goal: if preserving the \textit{pattern} of which features are strongly related is key (structure), the stricter threshold is preferable; if preserving exact correlation \textit{values} is crucial (for covariance-sensitive modeling), the looser threshold is better. In other words, rank-based metrics can signal improved qualitative structure even when the quantitative fit degrades. Thus, both types of correlation metrics should be reported.
    
\textbf{Privacy metrics vs.\ each other:} \textbf{L-diversity decouples from other privacy metrics.} While $RPR$ and CAP steadily improved with stricter thresholds (smaller $\tau_{\mathrm{ANY}}$), L-diversity sometimes moved in the opposite direction. For instance, at a moderate $\tau_{\mathrm{ANY}}$, \textit{Credit}/TVAE showed a huge L-diversity spike (implying better privacy via diversity) even though $RPR$ only improved moderately. Then at the strictest threshold, $RPR$ was highest (best privacy by distance) but L-diversity plummeted. This conflict means one could cherry-pick a metric to reach conflicting conclusions: e.g., ``moderate $\tau_{\mathrm{ANY}}$ tripled L-diversity, so it’s very safe'' versus ``moderate $\tau_{\mathrm{ANY}}$ only raised $RPR$ by 2 points, so privacy gain is minor.'' In truth, the filter reliably improves certain privacy aspects (distance-based leakage and attribute inference) but does not guarantee a diverse representation of sensitive values—it may overshoot or undershoot on that front. We identify a tension here: \textbf{privacy-by-novelty vs.\ privacy-by-disguise}. The filter ensures novelty (no exact copies), but it might over-diversify or under-diversify sensitive information. This reflects a known limitation of generative models: they tend to either memorize data (bad) or over-generalize (introduce unrealistic variety). In our results, $RPR$ and CAP confirm that memorization risk is reduced (good), but the swings in L-diversity show that the nature of variability in sensitive attributes is altered. This shift could pose a new privacy risk, or at least a utility concern, for those attributes.
    
\textbf{Privacy vs.\ Utility decoupling:} Within our tested range, we did not observe a strong privacy--utility trade-off—privacy gains generally came with no loss in utility. However, at the most extreme privacy setting ($\tau_{\mathrm{ANY}} = 0.005$) subtle tensions emerged: some data artifacts appeared (e.g., in \textit{Adult}/TVAE the marginals degraded, and in \textit{Credit} the diversity metrics dropped), hinting that pushing privacy further could eventually hurt utility. Thus, while utility stayed flat across our $\tau_{\mathrm{ANY}}$ settings, one can extrapolate that beyond a certain point a trade-off would surface. We also noted a minor decoupling in 
\textit{Cardio}/TVAE: the privacy metrics improved monotonically, yet utility (accuracy) dipped by about \(0.5\text{–}0.9\%\) at the strictest threshold (Table~\ref{tab:heom-any-xgboost}). This suggests that in this case the strict filter removed some useful information (perhaps a few informative but unique patient profiles), nudging performance down slightly. Although the effect is small, it highlights the expected tension: push privacy far enough, and utility will eventually give.


\subsection{Analysis}

Most metrics align on the big picture. Moderate filtering tends to help or at least not hurt, whereas extremely strict filtering can start to trade fidelity for privacy. However, the details reveal nuances. No single metric is sufficient on its own. Improvements in one dimension (marginal, correlation, privacy) do not automatically guarantee improvements elsewhere. Indeed, we observed marginals--vs.--correlations and different privacy notions diverging in certain cases. 

The takeaway is that \textbf{evaluation must be multi-faceted}. For example, report both JS divergence and correlation differences (to catch cases where good marginals mask poor structure), and report multiple privacy metrics (so that one metric’s ``safety'' isn’t misleading if another reveals a gap). Our findings underscore that the thresholded ANY filter’s effect is mostly orthogonal to utility (a positive result), moderately aligned with many resemblance metrics (especially for CTGAN), but partially orthogonal among privacy metrics. Hence, a careful threshold choice is needed depending on which privacy aspect one prioritizes.

\section{Limitations}
\label{sec:limitations}

Throughout this section, $\tau_{\mathrm{ANY}}$ denotes the filtering \emph{threshold} and $\widehat{\varepsilon}_{\mathrm{ANY}}(S)$ the \emph{achieved} ANY-risk on the returned synthetic set (when feasible, $\widehat{\varepsilon}_{\mathrm{ANY}}(S) < \tau_{\mathrm{ANY}}$).

Despite our encouraging results, several limitations must be acknowledged:

\begin{itemize}

    \item \textbf{Small Effect Sizes \& Practical Significance:} Many improvements in privacy metrics are quantitatively small (e.g., \texttt{RPR} +3–4\%, CAP +2\%). While statistically monotonic, such changes may not translate to a significantly reduced attack in practice—an adversary might not be much deterred by a $\sim 3\%$ drop in accuracy. Similarly, utility differences $<1\%$ mean our claim of ``no utility loss'' holds, but also imply limited sensitivity: a slightly larger disturbance might show an effect, and we might miss subtle degradation in more complex tasks.

    \item \textbf{L-Diversity Instability:} The wildly fluctuating L-diversity results highlight a potential threat to validity in conclusions about sensitive attribute protection. Depending on which \emph{threshold} $\tau_{\mathrm{ANY}}$ one looks at, one could draw opposite conclusions (some moderate $\tau_{\mathrm{ANY}}$ had higher synthetic diversity than real, others much lower). This instability suggests the need for caution: the filter might protect against identity disclosure in one sense (no exact matches) but fail in another (it could inadvertently make certain sensitive values more predictable by eliminating alternatives). Our evaluation of privacy might be incomplete if L-diversity can swing so much – it invites further scrutiny or complementary metrics like $k$-anonymity or $(\varepsilon,\delta)$-DP bounds.

    \item \textbf{Multiple Comparisons \& Optimal Choices:} We evaluated many metrics across many $\tau_{\mathrm{ANY}}$ values and reported the ``best'' in each case. This extensive search raises the risk of capitalizing on chance – some improvements might occur by luck (especially those 1–2\% differences). For example, the \textit{Adult}~TVAE rank-corr improvement of +0.200 is large and likely real, but a 1\% Frobenius drop could be noise. We did not adjust for multiple comparisons; thus some ``optimal~$\tau_{\mathrm{ANY}}$'' might not be significantly better than neighboring values. In practice, the true optimal might be a range. We mitigated this by focusing on consistent trends (monotonic changes, U-shapes), but the specific percentages should be taken as indicative, not exact.

    \item \textbf{Fixed Generator, No Retraining:} Importantly, our $\tau_{\mathrm{ANY}}$ sweeps were done on a single fixed generator per dataset/model (tuned for baseline utility). We did not retrain the generative model for each $\tau_{\mathrm{ANY}}$. This isolates the effect of filtering but is also a limitation: perhaps a model retrained with knowledge of an impending filter could learn differently (maybe generate slightly more diverse samples anticipating some will be filtered out). Also, because we fixed the random seed for generation (aside from filtering replacements), we might not have fully explored variance in synthetic sets. Our results assume the generator quality as given; a weaker generator might show more utility loss when filtered, or a stronger one might tolerate even stricter filtering. Thus, these results are conditional on using the \textit{post-hoc} filter approach on an already well-trained model.

    \item \textbf{Generality to Other Data/Attacks:} We evaluated on three datasets and specific attribute inference attacks. Different data (e.g., with image-like high dimensional features or highly imbalanced classes) might behave differently under filtering. Our privacy attacks are basic; adaptive adversaries might exploit synthetic data in ways we didn’t test (e.g., membership inference directly, which our distance metrics proxy but we didn’t explicitly compute membership error). We assume HEOM distance is a good proxy for record closeness – if an attacker had a different notion of closeness, our filter might not protect against that. These factors limit direct generalization.

    \item \textbf{Evaluation Budget \& Convergence:} We used up to 50 trials for hyperparameter tuning and fixed training epochs (1000–1500). With more training or alternative models, results might differ. There’s a chance that some observed small differences could become larger with a more powerful generator or more complex filter. Conversely, our CTGAN and TVAE were well-tuned; out-of-the-box models might overfit more and thus show greater utility drop when filtered (meaning our near-zero utility loss might be optimistic for less optimized models).

    \item \textbf{Composite Metric Considerations:} We did not formulate a single composite metric for the privacy–utility trade-off (though we recommended one in reporting). Thus, our analysis required subjective judgment to weigh, say, a 2\% JS improvement versus a 1\% accuracy dip versus a 3\% attacker performance drop. Different users might value these differently. Without a unified metric or cost function, this remains a limitation: we cannot definitively say ``$\tau_{\mathrm{ANY}} = 0.1$ is optimal'' for all cases—it is optimal in our view given how we prioritized metrics. Future work could formalize this trade-off (e.g., assign weights to privacy versus utility and solve for the best~$\tau_{\mathrm{ANY}}$).

\end{itemize}

While our findings are robust in showing that moderate filtering is beneficial and not harmful, the magnitude of benefits is modest and context-dependent. Stakeholders should interpret improvements with an understanding of these limitations. We advise additional safeguards (e.g., confidence intervals, multiple seeds) to ensure that the reported trends hold universally and that the \emph{achieved} risk $\widehat{\varepsilon}_{\mathrm{ANY}}(S)$ truly meets the chosen threshold $\tau_{\mathrm{ANY}}$ in practice.


\section{Conclusions}
\label{sec:conclusions}

In this paper, we proposed a simple post-processing pipeline that can be applied after any synthetic tabular data generator. The pipeline targets two recurring problems: synthetic tables that fail to reproduce some real categories, and synthetic records that lie suspiciously close to real individuals. To address these, we add (i) a targeted fine-tuning step that “patches” missing categories without retraining the model from scratch, and (ii) a distance-based sampling filter that rejects synthetic rows that are too close to the real data. We applied this pipeline to two widely used tabular generators (CTGAN and TVAE) on three public datasets (Credit, Adult, and Cardio), and evaluated the resulting synthetic data along three dimensions: resemblance to the original data, usefulness for training predictive models, and several indicators of privacy risk.

Overall, our experiments show that moderate filtering settings strike a favourable balance between these three goals. When the distance threshold is neither too loose nor too strict, the synthetic data better match the distribution of the original variables, while the performance of downstream classifiers remains within about \(1\%\) of the unfiltered generators. At the same time, distance-based privacy indicators improve, suggesting that synthetic records move farther away from individual real records without sacrificing predictive performance. Beyond these numerical gains, a key contribution of our work is to demonstrate that such improvements can be obtained as a post-hoc add-on, without changing the underlying generator architecture.

Our results lead to several practical guidelines for users of synthetic tabular data. First, whenever some real categories are absent from the synthetic tables, a focused “mode-patching” step can restore them and reduce hidden biases in downstream analyses. Second, the privacy filter should be tuned to avoid extremes: overly strict thresholds can harm both realism and utility, while moderate thresholds provide a good compromise, with CTGAN tolerating somewhat stronger filtering than TVAE in our experiments. Third, evaluations should be reported using a small set of complementary metrics rather than a single score, since improvements in marginal distributions, predictive performance, and privacy attacks do not always move in the same direction.

This study also has limitations. Our privacy filter is heuristic and relies on a specific notion of distance between records; it does not provide formal differential privacy guarantees, and the observed improvements, while consistent, are modest in absolute terms. The evidence is based on three medium-sized public datasets and on a limited family of privacy attacks, so results may differ on other data types or under more powerful adversaries. Future work includes combining our post-hoc filter with generators trained under formal privacy mechanisms, tuning the filtering threshold using explicit multi-objective criteria that reflect the needs of a given application, exploring alternative or learned distance measures for mixed-type data, and extending the evaluation beyond classification tasks. In addition, our analysis of local-density variance in the numeric feature space indicates that baseline generators may over-smooth certain datasets (e.g., Cardio), improving average utility while compressing variability in low-density regions. A promising direction is therefore to design geometry-aware generators or post-hoc corrections that explicitly regulate such numerical variance—e.g., by matching the distribution of local density statistics between real and synthetic data—so as to better preserve rare but application-relevant patterns without reintroducing proximity-based privacy risks. Taken together, these directions would strengthen the external validity of our conclusions and help move synthetic tabular data toward safer and more accountable use in practice.

\section*{Acknowledgment}
This research was made possible thanks to the support of Canadian insurance company Beneva,
NSERC research grant RDCPJ 537198-18 and an FRQNT doctoral research grant. We wish
to thank the reviewers for their comments regarding our work.

\ifCLASSOPTIONcaptionsoff
  \newpage
\fi



\bibliographystyle{IEEEtran}

%

\bibliography{bibtex/bib/IEEEexample}

%








\clearpage
\appendices
\setcounter{section}{0}
\renewcommand\thesection{A\arabic{section}}

\section{Baseline Comparisons}

\subsection{Resemblance Evaluation}
\label{app:baseline_resemblance}

We first compare the unfiltered CTGAN and TVAE baselines against the real data along the univariate and bivariate resemblance metrics introduced in Section~ \ref{Pipeline}.

\subsubsection{Univariate Marginals}

\begin{figure*}[t]
\centering

\begin{subfigure}{0.55\textwidth}
    \centering
    \includegraphics[width=\linewidth]{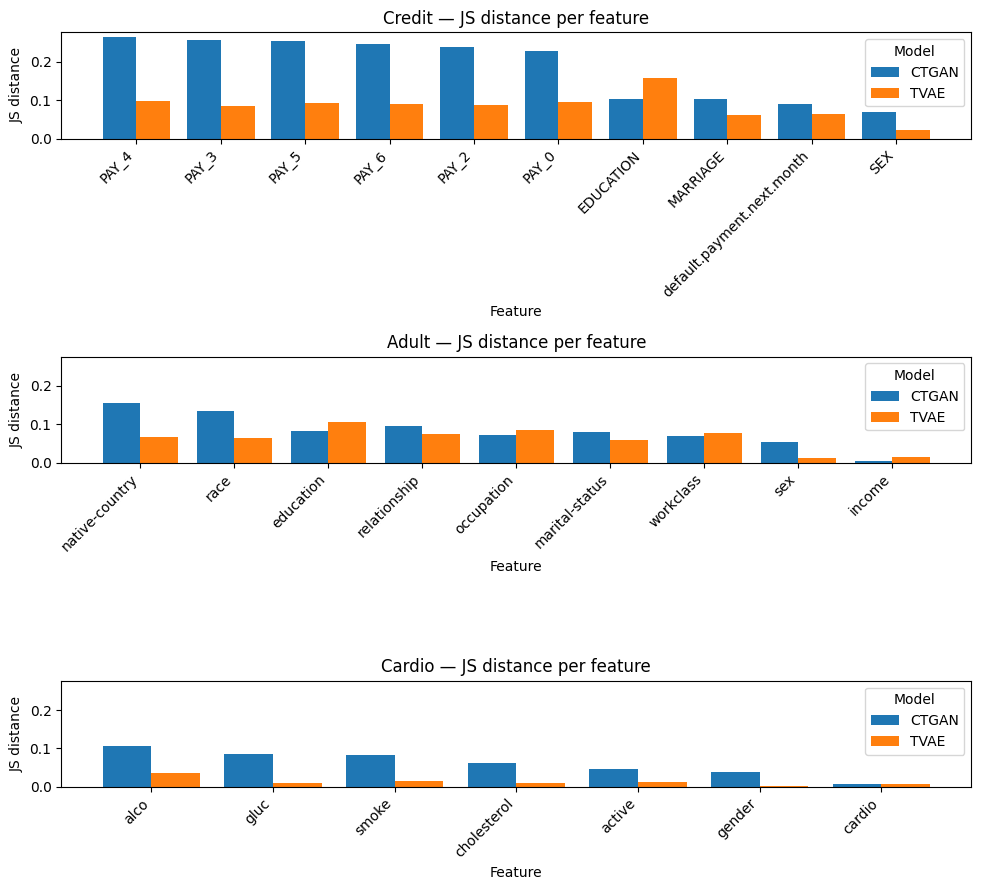}
    \caption{Per-feature Jensen--Shannon (JS) distance between real and synthetic categorical marginals for the Credit (top), Adult (middle), and Cardio (bottom) datasets. For each feature, the two bars compare the discrepancies obtained with CTGAN and TVAE.}
    \label{fig:baseline_js}
\end{subfigure}
\hfill
\begin{subfigure}{0.55\textwidth}
    \centering
    \includegraphics[width=\linewidth]{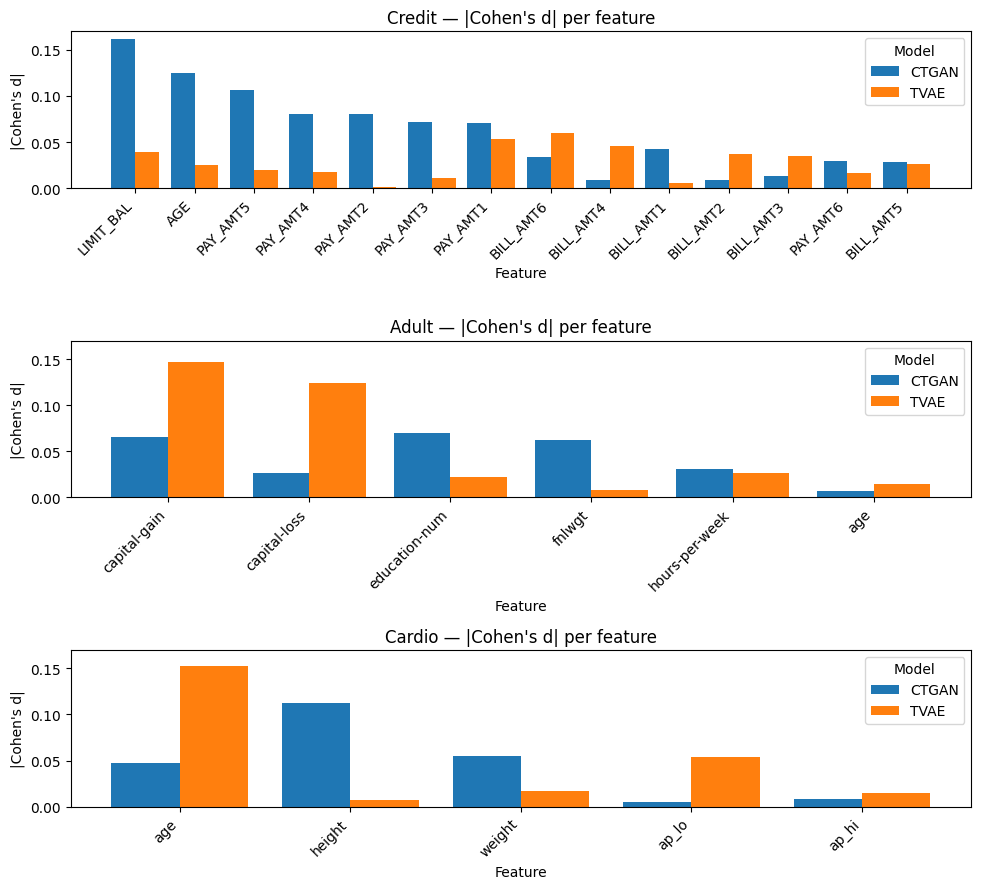}
    \caption{Per-feature absolute Cohen's $d$ between real and synthetic numeric marginals for the same three datasets, with paired bars per feature showing CTGAN versus TVAE.}
    \label{fig:baseline_cohen}
\end{subfigure}

\caption{Univariate resemblance between real data and the unfiltered CTGAN and TVAE baselines. Each panel shows grouped bar charts reporting, for every displayed feature, the magnitude of the discrepancy between real and synthetic marginal distributions: (a) JS distance for categorical variables and (b) absolute Cohen's $d$ for numeric variables. Higher bars indicate poorer marginal matching.}
\label{fig:baseline_js_cohen}
\end{figure*}

\textbf{Categorical features (JS distance).}
Across all datasets, both generators achieve reasonably small Jensen--Shannon (JS) distances, but TVAE tends to better match categorical marginals.

On \textbf{Credit}, CTGAN exhibits pronounced drift on the repayment-status variables PAY\_0--PAY\_6, with JS in the 0.23--0.26 range, yielding an average categorical JS of $\approx 0.19$. In contrast, TVAE keeps all repayment-status JS values below 0.10 and reduces the average JS to $\approx 0.09$. The main residual discrepancy for TVAE is EDUCATION (JS $\approx 0.16$), while the default label and demographic attributes show modest divergence in both models.

On \textbf{Adult}, categorical divergence is again higher for CTGAN than for TVAE (average JS $\approx 0.083$ vs.\ $\approx 0.063$). CTGAN's drift is concentrated in \emph{race} and \emph{native-country}, which together account for most of the categorical mismatch, whereas TVAE substantially reduces JS on these features and yields near-zero JS for the income label.

On \textbf{Cardio}, TVAE almost perfectly matches the categorical marginals, with all JS values below 0.04 and an average of $\approx 0.013$. CTGAN's categorical JS is already small but noticeably larger (average $\approx 0.062$), with the biggest discrepancies on lifestyle indicators (\emph{alco}, \emph{smoke}, \emph{gluc}) and cholesterol. The heatmaps in Fig.~\ref{fig:baseline_js} (right column) visualize these patterns: darker bands appear for CTGAN on the Credit repayment statuses and Adult demographic features, while TVAE is mostly near-uniform except for a few cells (e.g., EDUCATION in Credit).

\textbf{Numeric features (Cohen's $d$).}
Effect sizes for numeric variables are small across all dataset--model pairs: every feature has $|d| \leq 0.17$, indicating that the synthetic and real means differ by at most about 0.17 pooled standard deviations.

For \textbf{Credit}, CTGAN shows its largest deviations on LIMIT\_BAL and AGE ($d \approx 0.16$ and $0.13$), with bill amounts and payment amounts generally below 0.11. TVAE further reduces these shifts, keeping all numeric $|d|$ values below $\approx 0.06$.

For \textbf{Adult}, both generators closely match the numeric marginals. CTGAN's effect sizes remain below $\approx 0.07$ for all features. TVAE behaves similarly, except for \emph{capital-gain} and \emph{capital-loss}, where $|d|$ rises to $\approx 0.15$ and $\approx 0.12$ respectively, reflecting the difficulty of capturing the highly skewed tails of these variables.

For \textbf{Cardio}, numeric resemblance is again strong. CTGAN yields modest discrepancies on \emph{height} ($d \approx 0.11$) and smaller ones elsewhere, whereas TVAE more accurately reproduces \emph{height} and \emph{weight} but exhibits a somewhat larger shift on \emph{age} ($d \approx 0.15$). Overall, the heatmaps in Fig.~\ref{fig:baseline_cohen} (left column) show that both generators keep numeric effect sizes in a ``small'' regime, with no feature exhibiting a large univariate distortion.

Taken together, these results indicate that TVAE is generally the stronger baseline for \textbf{categorical} marginals---especially on Cardio and on the repayment-status variables in Credit---while CTGAN offers comparable or slightly better alignment for some \textbf{numeric} variables (e.g., Adult's capital-related features, Cardio age).

\subsubsection{Bivariate Dependence Structure}

\begin{figure*}[t]
\centering
\includegraphics[width=0.7\textwidth]{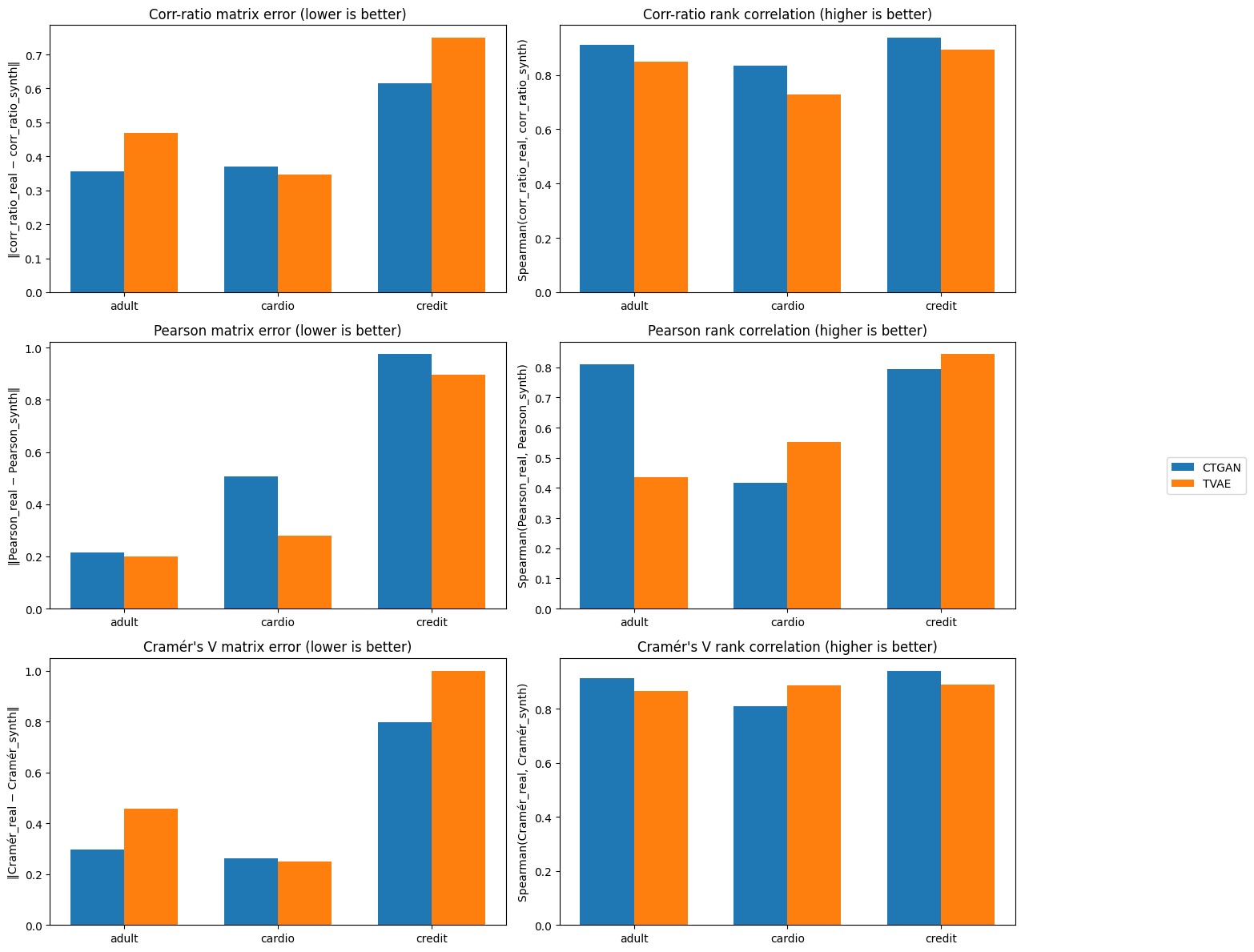}
\caption{Bivariate resemblance between real data and the unfiltered CTGAN and TVAE baselines on the Adult, Cardio, and Credit datasets. Each row corresponds to a different association measure between variables: the correlation ratio $\eta^2$ for numeric--categorical pairs (top), Pearson's $\rho$ for numeric--numeric pairs (middle), and Cramer's $V$ for categorical--categorical pairs (bottom). For every dataset and association type, the left column reports the Frobenius norm of the difference between the real and synthetic dependence matrices (\texttt{diff\_norm}$_\cdot$, lower is better), while the right column shows the Spearman rank correlation between their off-diagonal entries (\text{spearman}, higher is better). Bars compare CTGAN (blue) and TVAE (orange).}
\label{fig:baseline_correlation}
\end{figure*}

To assess multivariate resemblance, we compare real and synthetic dependence matrices for Pearson's $\rho$ (numeric--numeric), Cramer's $V$ (categorical--categorical), and the correlation ratio $\eta^2$ (numeric--categorical), summarizing discrepancies by the Frobenius norm of the difference matrices (\texttt{diff\_norm}$_\cdot$, lower is better) and by the Spearman rank correlation between their off-diagonal entries ($\cdot_{\text{spearman}}$, higher is better). The summary is shown in Fig.~A.2.

On \textbf{Adult}, CTGAN better preserves most aspects of dependence structure. For categorical--categorical and mixed pairs, its Frobenius errors are substantially smaller than TVAE's ($V$: 0.30 vs.\ 0.46; $\eta^2$: 0.36 vs.\ 0.47) and its rank correlations are higher ($r_{\mathrm{S}} \approx 0.91$ for both $V$ and $\eta^2$, vs.\ $\approx 0.87$ and 0.85 for TVAE). For numeric--numeric associations, TVAE slightly reduces the matrix error (0.20 vs.\ 0.21) but at the cost of much poorer rank alignment (Spearman 0.44 vs.\ 0.81), suggesting that CTGAN better preserves the ordering of strong vs.\ weak correlations even if absolute magnitudes are similar.

On \textbf{Cardio}, TVAE is clearly superior for numeric and categorical dependencies. Its Pearson error is roughly half that of CTGAN (0.28 vs.\ 0.51) with higher rank correlation (0.55 vs.\ 0.42), and its Cramer's $V$ error is slightly smaller (0.25 vs.\ 0.26) with noticeably higher rank agreement (0.89 vs.\ 0.81). For mixed-type correlations ($\eta^2$), the two models are comparable in magnitude (0.35 vs.\ 0.37), with CTGAN achieving somewhat better rank correlation (0.83 vs.\ 0.73).

On \textbf{Credit}, both models struggle more with dependencies than on the other datasets, but their strengths differ by association type. CTGAN better matches categorical and mixed relationships ($V$ error 0.80 vs.\ 1.00; $\eta^2$ error 0.62 vs.\ 0.75, with rank correlations 0.94 vs.\ 0.89 and 0.94 vs.\ 0.89 respectively), whereas TVAE more faithfully reproduces numeric correlations (Pearson error 0.90 vs.\ 0.98, Spearman 0.84 vs.\ 0.79). This mirrors the univariate results: CTGAN is comparatively stronger on the complex categorical and ordinal structure of Credit, while TVAE excels on its numeric variables.

Overall, the baseline comparison shows that:
\begin{itemize}
    \item Both generators yield \textbf{good univariate resemblance}, with small effect sizes and moderate JS distances;
    \item \textbf{TVAE} systematically offers better \textbf{categorical} marginals and stronger multivariate fidelity on \textbf{Cardio}, and for numeric relations on \textbf{Credit};
    \item \textbf{CTGAN} better captures \textbf{mixed and categorical dependencies} on \textbf{Adult} and \textbf{Credit}, and often maintains a more faithful ordering of pairwise association strengths even when matrix errors are similar.
\end{itemize}

These baseline characteristics provide a reference point for interpreting the impact of the post-hoc privacy mechanisms and downstream utility analyses reported in the main text and subsequent sections of this appendix.


\subsection{Baseline Utility (ROC--AUC)}
\label{app:baseline_utility}

Baseline downstream classification utility for the unfiltered CTGAN and TVAE samples is reported and discussed in the main text (see Figure~\ref{fig:baseline_utility} and Section~\ref{sec:baseline_utility}). In brief, ROC--AUC under TSTR closely tracks the corresponding TRTR performance across all datasets and classifiers, with only modest degradation on Credit and Adult and occasional improvements on Cardio. We therefore use those main-text results as the reference utility level when assessing the impact of the post-hoc privacy mechanisms, and do not repeat the plots or detailed commentary here.

\subsection{Privacy Evaluation}
\label{app:privacy}

\begin{figure*}[t]
\centering

\begin{subfigure}{0.48\textwidth}
    \centering
    \includegraphics[width=\linewidth]{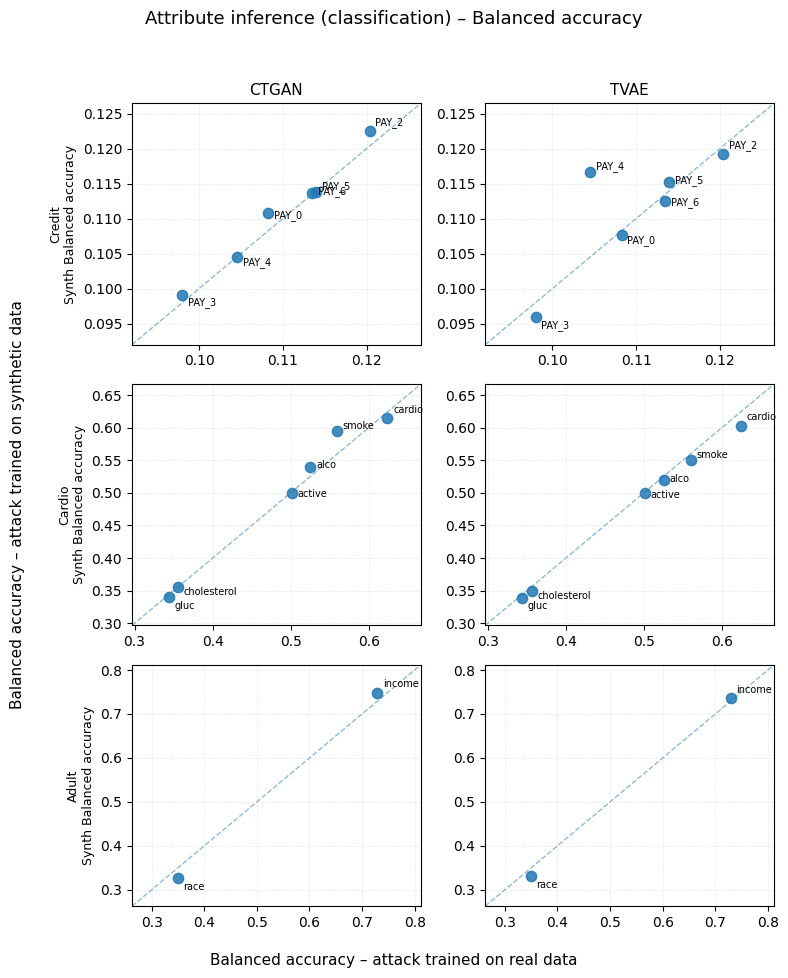}
    \caption{Attribute-inference attacks (classification) on the baseline CTGAN and TVAE samples. Each scatter plot corresponds to a dataset--generator pair (rows: Credit, Cardio, Adult; columns: CTGAN, TVAE). Points denote sensitive categorical attributes, with the $x$-axis showing the balanced accuracy of an attack classifier trained on \emph{real} data and the $y$-axis showing the same classifier trained on \emph{synthetic} data. The dashed diagonal marks parity between the two training sources; points above (below) the line indicate stronger (weaker) attacks when trained on synthetic data.}
    \label{fig:baseline_aia_classification}
\end{subfigure}
\hfill
\begin{subfigure}{0.48\textwidth}
    \centering
    \includegraphics[width=\linewidth]{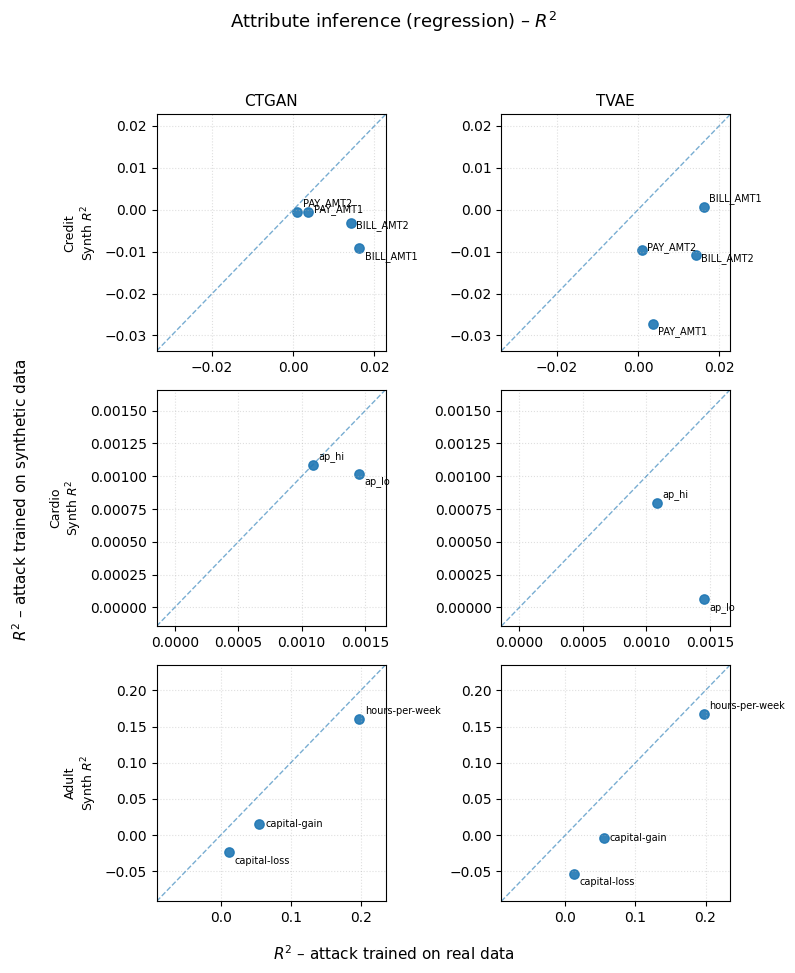}
    \caption{Attribute-inference attacks (regression) for continuous sensitive attributes. Layout matches panel~\subref{fig:baseline_aia_classification}. Each point is a target attribute, with the $x$-axis giving the coefficient of determination $R^2$ for a linear attack model trained on real data and the $y$-axis giving $R^2$ when the same model is trained on synthetic data. The dashed diagonal again indicates equal attack strength under real- vs.\ synthetic-trained attacks.}
    \label{fig:baseline_aia_regression}
\end{subfigure}

\caption{Baseline attribute-inference privacy evaluation for CTGAN and TVAE. The left column reports classification attacks in terms of balanced accuracy, while the right column reports regression attacks in terms of $R^2$. In all panels, proximity to the diagonal indicates that training attacks on synthetic data does not substantially change their effectiveness relative to training on the original real records.}
\label{fig:baseline_aia}
\end{figure*}

\begin{figure*}[t]
\centering
\includegraphics[width=0.8\textwidth]{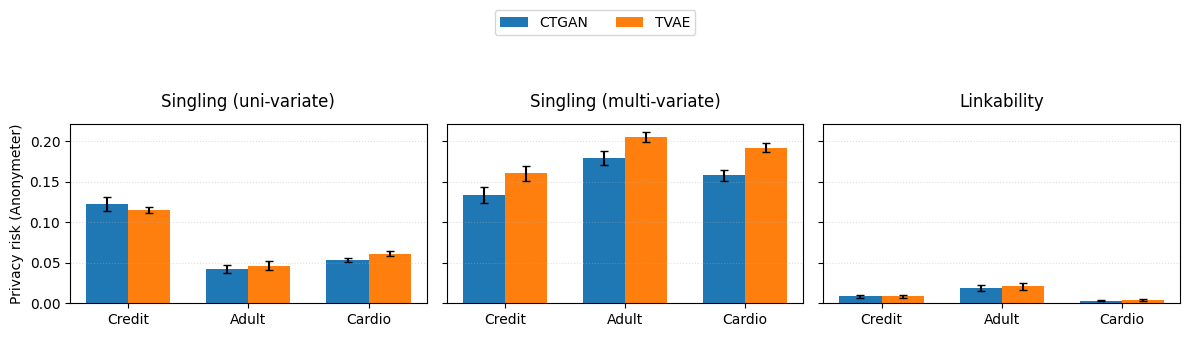}
\caption{Record-level privacy risks estimated with Anonymeter for the baseline CTGAN and TVAE models. Each panel shows the mean risk (vertical axis, lower is better) with one standard deviation as error bars, computed over ten runs for the Credit, Adult, and Cardio datasets. Left: singling-out risk when the attacker uses a single attribute (univariate). Middle: singling-out risk when the attacker can combine multiple attributes (multivariate). Right: linkability risk, i.e., the probability of correctly linking several synthetic records back to the same individual. In all panels, blue bars correspond to CTGAN and orange bars to TVAE.}
\label{fig:baseline_meanrisk_anonymeter}
\end{figure*}

We assess the privacy properties of CTGAN and TVAE using two complementary families of attacks: \emph{attribute-inference attacks (AIA)} and \emph{record-level attacks with Anonymeter}. The former measure how well an adversary can reconstruct sensitive attributes from quasi-identifiers, while the latter quantify singling-out and linkability risks at the individual record level.

\subsubsection*{Attribute-Inference -- Classification}

Figure~\ref{fig:baseline_aia_classification} (top) shows the balanced accuracy of attribute-inference classifiers trained either on \emph{real} data (x-axis) or on \emph{synthetic} data (y-axis). Points close to the diagonal correspond to attacks that are about as effective on both training sources.

\paragraph*{Credit.} 
All \texttt{PAY\_k} status variables yield very low balanced accuracy, around $0.10$--$0.12$ for both CTGAN and TVAE, irrespective of whether the attack is trained on real or synthetic data. This is essentially close to random guessing for these multi-class targets, indicating that neither baseline model substantially helps an attacker infer repayment status from the remaining attributes.

\paragraph*{Adult.}
For the \texttt{income} attribute, the attack reaches balanced accuracies around $0.73$--$0.75$ when trained on either real or synthetic data, for both CTGAN and TVAE. For \texttt{race}, the balanced accuracy is much lower ($\approx 0.33$--$0.35$). Differences between real-trained and synthetic-trained attacks are within roughly $\pm 0.02$, with no systematic advantage to training on synthetic samples. Hence, synthetic data preserves the inherent predictability of income and race in this dataset but does not noticeably amplify it.

\paragraph*{Cardio.}
For binary lifestyle attributes such as \texttt{smoke} and \texttt{alco}, balanced accuracies lie between $\approx 0.52$ and $0.60$, while \texttt{cholesterol}, \texttt{gluc}, \texttt{active}, and the disease label (\texttt{cardio}) range between $\approx 0.34$ and $0.62$. Again, attacks trained on synthetic data perform very similarly to those trained on real records; deviations are typically below $0.03$ balanced-accuracy points and are not consistently in favor of either CTGAN or TVAE.

Overall, the classification AIA results indicate that \textbf{synthetic data does not make attribute inference easier than real data}. For credit and cardio, absolute attack performance is low to moderate, while for adult the risk is dominated by strong correlations already present in the original dataset.

\subsubsection*{Attribute-Inference -- Regression}

We next train linear regression attacks to predict continuous attributes, reporting the coefficient of determination $R^2$ (Figure~\ref{fig:baseline_aia_regression} bottom).

\paragraph*{Credit.}
For CTGAN, attacks trained on real data achieve very small $R^2$ values ($0.01$--$0.02$ for \texttt{BILL\_AMT1/2} and below $0.004$ for \texttt{PAY\_AMT1/2}). When trained on synthetic samples, $R^2$ drops to values close to zero or slightly negative for both CTGAN and TVAE, meaning the best linear attack barely improves over predicting the mean.

\paragraph*{Adult.}
The continuous targets are somewhat more predictable. On real data, $R^2$ is $\approx 0.20$ for \texttt{hours-per-week} and $\approx 0.06$ / $0.01$ for \texttt{capital-gain} / \texttt{capital-loss}. Attacks trained on CTGAN or TVAE samples reproduce these dependencies only partially: for hours-per-week $R^2$ decreases to $\approx 0.16$--$0.17$, and for capital gain/loss it drops toward zero or slightly negative values. Thus, synthetic data somewhat \emph{attenuates} the regression-based inference risk for this dataset.

\paragraph*{Cardio.}
For both \texttt{ap\_hi} and \texttt{ap\_lo} the $R^2$ values are $\approx 0.001$ whether the attack is trained on real or synthetic data, indicating essentially no linear predictive signal.

Taken together, the regression experiments show \textbf{very limited ability to reconstruct continuous sensitive attributes from the synthetic tables}, and in the adult dataset they even reduce the accuracy of such attacks relative to using the original data.

\subsubsection*{Record-Level Privacy -- Anonymeter}

Finally, we quantify \emph{singling-out} and \emph{linkability} risks using Anonymeter, running each configuration ten times. Figure~\ref{fig:baseline_meanrisk_anonymeter} summarizes the mean privacy risk (vertical axis) and one standard deviation (error bars) across datasets and generators.

\paragraph*{Linkability.}
Linkability risks are consistently the smallest across all settings. The mean risk stays below $0.03$ in every case: around $0.018$--$0.021$ for \texttt{Adult}, $\approx 0.009$ for \texttt{Credit}, and $\approx 0.003$--$0.004$ for \texttt{Cardio}. The difference between CTGAN and TVAE is negligible at this scale, suggesting that \textbf{linking multiple synthetic records back to the same original individual is difficult}.

\paragraph*{Singling-Out (Univariate).}
Using single attributes, the average singling-out risk ranges between about $0.04$ and $0.06$ for \texttt{Adult} and \texttt{Cardio}, and between $0.11$ and $0.12$ for \texttt{Credit}. TVAE tends to be slightly more exposed than CTGAN on Adult and Cardio (e.g., mean risk $0.046$ vs.\ $0.042$ for Adult, and $0.061$ vs.\ $0.053$ for Cardio), while CTGAN has slightly higher univariate risk on the Credit dataset ($0.123$ vs.\ $0.115$).

\paragraph*{Singling-Out (Multivariate).}
When an attacker can combine several attributes, the risks increase, as expected, but remain moderate. For \texttt{Adult}, the mean multivariate singling-out risk is $\approx 0.18$ for CTGAN and $\approx 0.21$ for TVAE. For \texttt{Cardio} it is $\approx 0.16$ (CTGAN) and $\approx 0.19$ (TVAE), and for \texttt{Credit} $\approx 0.13$ (CTGAN) and $\approx 0.16$ (TVAE). The corresponding mean \emph{specific} risk (the part attributable to training on the actual records rather than on a control dataset) is on the order of $0.04$ for Adult CTGAN and up to $\approx 0.12$ for Credit TVAE, indicating that \textbf{only a fraction of the total risk stems from membership in the training data}.

Across all three datasets, Anonymeter therefore paints a consistent picture:

\begin{itemize}
    \item \textbf{Absolute risks are non-zero but relatively low}, especially for linkability.
    \item \textbf{TVAE tends to incur slightly higher singling-out risks than CTGAN}, particularly in the multivariate setting for Adult and Cardio, while the converse holds for the univariate Credit scenario.
\end{itemize}

These baseline results provide a reference privacy profile against which we compare the proposed method in the main text.

\section{Univariate Evaluation Extended Results}

\subsection{Metrics and Reading Guide}

\subsubsection{Categorical marginals}

\medskip
We compare the empirical level frequencies of a categorical feature $X$ in the real data ($p_{\mathrm{real}}$) and a synthetic variant ($p_{\mathrm{synth}}$).

\paragraph{log2 fold-change (log2FC).}
\[
\mathrm{log2FC}(c)=\log_2\!\Big(\frac{p_{\mathrm{synth}}(X{=}c)}{p_{\mathrm{real}}(X{=}c)}\Big).
\]

\emph{Interpretation.} Measures the directional, multiplicative change of each level $c$ between synthetic and real.  

\emph{Reading.} $\mathrm{log2FC}(c)=0$ indicates no change; $+1$ means $p_{\mathrm{synth}}$ doubles relative to $p_{\mathrm{real}}$; $-1$ halves it. Values are in log base~2.  

\emph{Role.} Highlights over- and under-representation of individual levels, especially for commonly occurring categories.

\medskip
\paragraph{Per-level JS contribution (JS\_contrib).}

\medskip
Using base-2 logs, the per-level contribution of category $c$ is
\[
\begin{aligned}
\mathrm{JS\_contrib}(c)
&= \tfrac{1}{2}\,p_{\mathrm{real}}(c)\,\log_{2}\!\frac{p_{\mathrm{real}}(c)}{M(c)}
\\[-2pt]
&\quad+ \tfrac{1}{2}\,p_{\mathrm{synth}}(c)\,\log_{2}\!\frac{p_{\mathrm{synth}}(c)}{M(c)},
\\[-2pt]
M(c) &= \tfrac{1}{2}\big(p_{\mathrm{real}}(c)+p_{\mathrm{synth}}(c)\big).
\end{aligned}
\]

Each term is non-negative, and  
\[
\mathrm{JS}(X)=\sum_{c} \mathrm{JS\_contrib}(c)\in[0,1]\text{ bit}.
\]

\emph{Interpretation.} Quantifies how much each level $c$ contributes to the total Jensen–Shannon divergence, weighted by its probability mass.  

\emph{Reading.} Summing $\mathrm{JS\_contrib}(c)$ over levels of a feature yields the feature-level JS; summing feature-level JS over features yields the total categorical divergence.  

\emph{Role.} Complements log2FC by down-weighting very rare levels: a level can have a large log2FC but small $\mathrm{JS\_contrib}$ if its mass is negligible.

\medskip
\paragraph{Heatmap structure.}
In log2FC heatmaps, columns correspond to acceptance targets $\tau$, and rows to category levels.  
Warm colors indicate over-representation ($\mathrm{log2FC}>0$); cool colors indicate under-representation ($\mathrm{log2FC}<0$).  
As $\tau$ tightens, fading of large-magnitude colors indicates improved alignment of synthetic and real level frequencies.

\paragraph{Common pitfalls.}
(i) If $p_{\mathrm{real}}(c)=0$, log2FC is undefined; such cells are omitted or numerically clipped.  
(ii) Rare levels often show extreme log2FC but contribute little to total JS because their $\mathrm{JS\_contrib}$ is tiny.

\medskip
\medskip

\subsubsection{Numeric marginals (quantile heuristic)}

For a numeric feature $Y$, we compute baseline quantiles at the 5th, 50th, and 95th percentiles and compare them with the variant:
\[
Q^{B}_L,Q^{B}_{50},Q^{B}_H; \qquad Q^{V}_L,Q^{V}_{50},Q^{V}_H.
\]

\paragraph{Relative quantile shifts.}
\[
\Delta_{\mathrm{med}}=\frac{Q^{V}_{50}}{Q^{B}_{50}}-1,\quad
\Delta_{\mathrm{low}}=\frac{Q^{V}_{L}}{Q^{B}_{L}}-1,\quad
\Delta_{\mathrm{high}}=\frac{Q^{V}_{H}}{Q^{B}_{H}}-1.
\]

\emph{Interpretation.}  
$\Delta_{\mathrm{med}}$ captures the relative shift of the center;  
$\Delta_{\mathrm{low}}$ and $\Delta_{\mathrm{high}}$ capture relative distortions of the lower and upper tails.  
We report these as percentages.

\emph{Reading.} For example, $\Delta_{\mathrm{high}}=+0.10$ means the 95th-percentile in the variant is 10\% larger than in the baseline. Comparing $\Delta_{\mathrm{med}}$ vs.\ $(\Delta_{\mathrm{low}},\Delta_{\mathrm{high}})$ indicates whether shifts are mostly central or driven by tails.

\paragraph{Tail-mass deltas.}
Using the \emph{baseline} thresholds $Q^{B}_L$ and $Q^{B}_H$:
\begin{align*}
\Delta p_{\mathrm{low}}  &= \Pr_V[Y\le Q^{B}_L]-\Pr_B[Y\le Q^{B}_L],\\
\Delta p_{\mathrm{high}} &= \Pr_V[Y\ge Q^{B}_H]-\Pr_B[Y\ge Q^{B}_H].
\end{align*}

\emph{Interpretation.} Measures how many more or fewer samples fall into the lower and upper tails (relative to the baseline), in \emph{percentage points}. This is complementary to quantile shifts: quantiles track where the cutpoints move, while tail-mass deltas track how much mass lies beyond fixed cutpoints.

\paragraph{Reading numeric tables.}
We summarize (i) median absolute relative quantile shifts and (ii) mean absolute tail-mass deltas across features.  
These tables highlight whether distortions are primarily in the center ($|\Delta_{\mathrm{med}}|$), lower tail ($|\Delta_{\mathrm{low}}|$, $|\Delta p_{\mathrm{low}}|$), or upper tail ($|\Delta_{\mathrm{high}}|$, $|\Delta p_{\mathrm{high}}|$).

\paragraph{Example.}
Baseline $(Q_L,Q_{50},Q_H)=(40,100,200)$; variant $(44,110,180)$.  
Relative shifts are $+10\%$, $+10\%$, $-10\%$ for $(\Delta_{\mathrm{low}},\Delta_{\mathrm{med}},\Delta_{\mathrm{high}})$.  
If $5\%$ of baseline samples and $8\%$ of variant samples exceed $200$, then $\Delta p_{\mathrm{high}}=+3$ percentage points.

\paragraph{Common pitfalls.}
Undefined ratios arise if $Q^{B}_L=0$ or $Q^{B}_H=0$; we then report NaN (or add a small $\epsilon$ if needed for numerical stability).  
For signed features, relative ratios around zero can be hard to interpret; we therefore emphasize effect sizes and tail-mass deltas in those cases.

\medskip
\medskip
\subsubsection{Cohen’s \texorpdfstring{$d$}{d} (numeric effect size)}
\[
d=\frac{\bar{Y}_V-\bar{Y}_B}{s_p},\qquad
s_p=\sqrt{\frac{(n_V-1)s_V^2+(n_B-1)s_B^2}{n_V+n_B-2}}.
\]

\emph{Interpretation.} Standardized mean shift between variant and baseline, measured in pooled standard deviations.  

\emph{Reading.} As a heuristic, $|d|{<}0.2$ is “small”, $|d|\approx 0.5$ “medium”, and $|d|\ge 0.8$ “large”. These thresholds are only rough guidelines.  

\emph{Role.} Captures central shifts but is insensitive to pure tail distortions when means remain stable.

\medskip
\medskip
\subsubsection{Putting the metrics together}
\begin{itemize}\setlength{\itemsep}{2pt}
  \item Large log2FC but small JS\_contrib: a large relative change on a rare level with little impact on overall divergence.
  \item Feature-level JS decreases as $\tau$ tightens: categorical level frequencies move closer to the real distribution.
  \item Small $|d|$ but large $\Delta p_{\mathrm{high}}$: mean is well matched, but the synthetic distribution has a heavier upper tail.
  \item Large $|d|$ but small tail deltas: discrepancies are dominated by central shifts rather than tails.
  \item As $\tau$ becomes stricter, we typically see:
    \begin{itemize}\setlength{\itemsep}{1pt}
      \item improved center alignment ($d$, $\Delta_{\mathrm{med}}$),
      \item possible re-weighting into tails or rare categorical levels.
    \end{itemize}
\end{itemize}

\medskip
\medskip
\subsubsection{Units and scales}
\begin{itemize}\setlength{\itemsep}{2pt}
  \item log2FC: dimensionless ($+1$ corresponds to $\times 2$; $-1$ to $\times \tfrac{1}{2}$).
  \item JS\_contrib/JS: non-negative; $0$ means identical distributions; measured in bits (log base~2).
  \item Relative quantile shifts: percentages.
  \item Tail-mass deltas: percentage points (pp).
  \item Cohen’s $d$: pooled standard deviations (standardized effect size).
\end{itemize}

\subsection{Feature-level fidelity (credit CTGAN)}

\subsubsection{Categorical marginals (log2FC \& JS)}
\begin{figure*}[t]
\centering
\includegraphics[width=\textwidth]{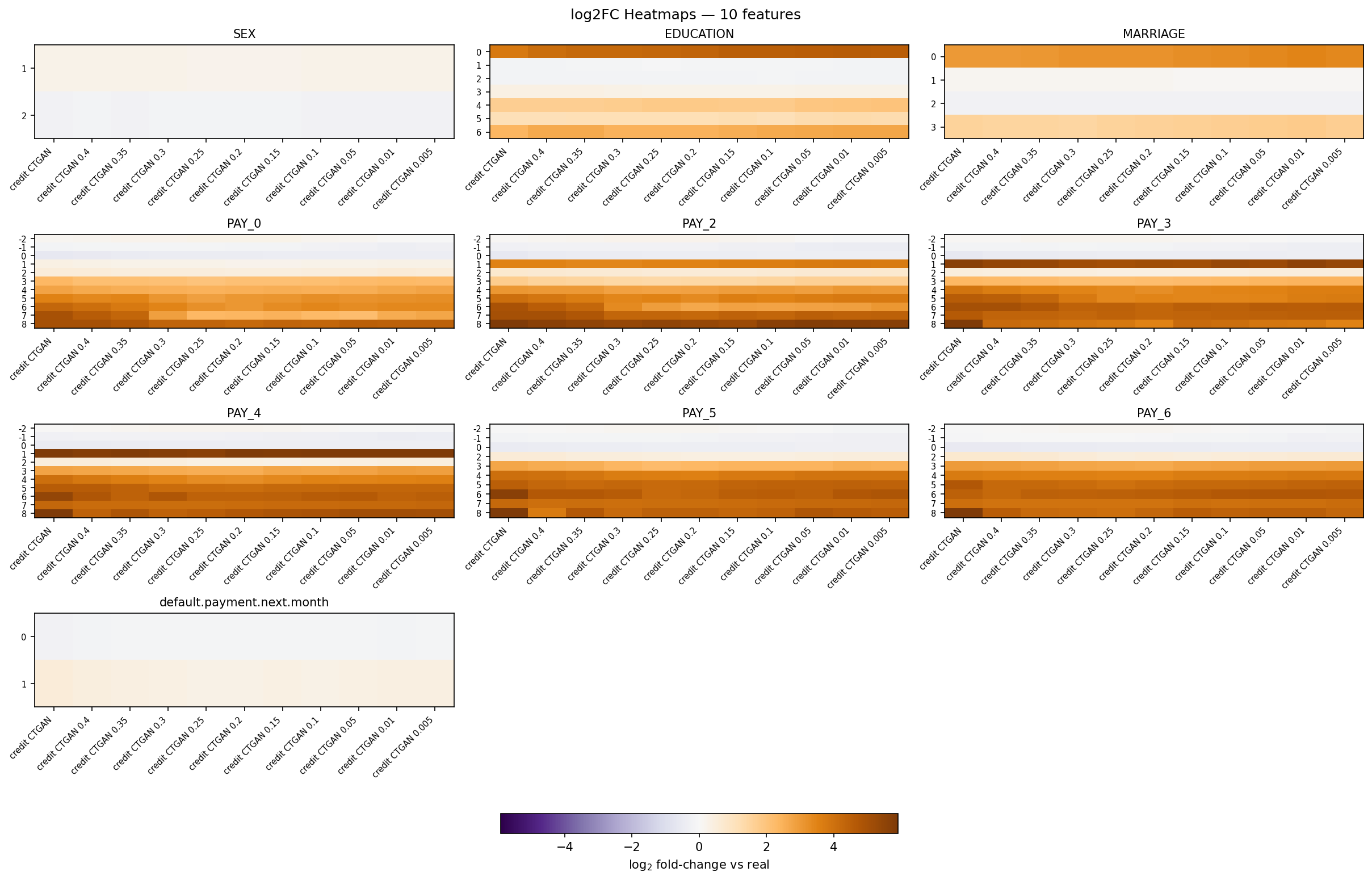}
\caption{Feature‐level categorical fidelity across acceptance targets.
Each panel shows the log\(_2\) fold‐change \(\mathrm{log2FC}=\log_2(p_{\mathrm{synth}}/p_{\mathrm{real}})\) for every level of a categorical feature, comparing
the unconstrained CTGAN sample (“credit CTGAN”) and its HEOM–kNN post‐filtered variants at targets
\(\tauany \in \{0.40,0.35,\ldots,0.005\}\) (columns, left–to–right). Rows are category levels in code order; the colorbar (bottom) encodes direction and magnitude
(white \(\approx 0\); warm \(>\!0\), over‐represented in the synthetic data; cool \(<\!0\), under‐represented; scale clipped for readability).
Divergence concentrates in repayment‐status features \(\mathrm{PAY}_0\text{--}\mathrm{PAY}_6\), which exhibit a shift away from level “0” (no delay) and toward higher delay codes.
Moderate acceptance (\(\tauany \approx 0.20\text{--}0.25\)) visibly reduces large deviations without re‐inflating rare levels, consistent with the
aggregate JS results in Table~\ref{tab:cat-total-js}.}
\label{fig:log2fc-heatmaps}
\end{figure*}

Figure~\ref{fig:log2fc-heatmaps} visualizes log2FC by level and acceptance target. Divergence concentrates in the repayment-status features
\(\mathrm{PAY}_0\text{--}\mathrm{PAY}_6\), with a systematic shift \emph{away} from level “0” (no delay) and \emph{toward} higher delay codes. Summed over levels,
the total categorical JS (in bits, log base~2) is lowest at \(\tau\!\approx\!0.20\text{--}0.25\) and rises when the acceptance target is too loose or too tight (Table~\ref{tab:cat-total-js}).

\begin{table}[!t]
\centering
\small
\setlength{\tabcolsep}{3.5pt}
\begin{threeparttable}
\caption{Total categorical divergence (sum of per‐feature JS, \emph{log base 2; bits}) vs.\ acceptance target. Lower is better. $\Delta$ columns show change vs.\ unconstrained CTGAN; negative values are improvements. Minimum at $\tauany \approx 0.20$.}
\label{tab:cat-total-js}
\begin{tabularx}{\columnwidth}{
  >{\raggedright\arraybackslash}X
  S[table-format=1.4]
  S[table-format=+1.4]
  S[table-format=+2.1]}
\toprule
Variant & {Total JS} & {$\Delta$ vs CTGAN} & {\% $\Delta$} \\
\midrule
credit CTGAN         & 0.4049 & \multicolumn{1}{c}{—}     & \multicolumn{1}{c}{—} \\
credit CTGAN 0.40    & 0.3446 & -0.0604 & -14.9 \\
credit CTGAN 0.35    & 0.3177 & -0.0873 & -21.6 \\
credit CTGAN 0.30    & 0.2802 & -0.1248 & -30.8 \\
credit CTGAN 0.25    & 0.2635 & -0.1414 & -34.9 \\
\textbf{credit CTGAN 0.20} & \bfseries 0.2599 & \bfseries -0.1450 &  \bfseries -35.8 \\
credit CTGAN 0.15    & 0.2789 & -0.1261 & -31.1 \\
credit CTGAN 0.10    & 0.2863 & -0.1186 & -29.3 \\
credit CTGAN 0.05    & 0.3040 & -0.1010 & -24.9 \\
credit CTGAN 0.01    & 0.3230 & -0.0819 & -20.2 \\
credit CTGAN 0.005   & 0.3228 & -0.0822 & -20.3 \\
\bottomrule
\end{tabularx}
\begin{tablenotes}[flushleft]\footnotesize
\item \textit{Note.} “Total JS’’ sums per‐level contributions within each categorical feature and then across features. All JS values are reported in \emph{log base 2} (bits).
\end{tablenotes}
\end{threeparttable}
\end{table}

\subsubsection{Numeric marginals (quantile heuristics and \(d\))}
Aggregating the quantile heuristics across numeric features reveals a center–tail trade-off.
The median absolute center shift \(\mathrm{median}(|\Delta_{\mathrm{med}}|)\) is smallest under tight acceptance
(0.01: 8.4\%), while the upper tail aligns best near 0.25--0.30
(0.25: 4.5\%; 0.30: 4.8\%), with the smallest mean absolute upper-tail mass difference at 0.30
(0.82 pp). Lower-tail mass differences increase as the target tightens.
These dataset-level summaries are provided in Table~\ref{tab:numeric-heuristic-summary}.

\begin{table}[t]
\centering
\small
\setlength{\tabcolsep}{2.5pt} 
\caption{Numeric marginal fidelity at \(q_L{=}0.05, q_H{=}0.95\). Reported are median absolute relative shifts (\%) and mean absolute tail‐mass deltas (\%). Lower is better; bold indicates the best per column.}
\label{tab:numeric-heuristic-summary}
\resizebox{\columnwidth}{!}{%
\begin{tabular}{lccccc}
\toprule
Dataset & Med.~$|\Delta_{\mathrm{med}}|$ & Med.~$|\Delta_{\mathrm{low}}|$ & Med.~$|\Delta_{\mathrm{high}}|$ & Mean~$|\Delta p_{\mathrm{low}}|$ & Mean~$|\Delta p_{\mathrm{high}}|$\\
\midrule
Baseline & 18.7\% & 14.3\% & 8.4\% & \textbf{3.2\%} & 1.1\% \\
$\tau_{\mathrm{ANY}}{=}0.40$ & 18.9\% & 13.1\% & 8.6\% & 3.4\% & 1.0\% \\
$\tau_{\mathrm{ANY}}{=}0.35$ & 19.6\% & 13.0\% & 7.5\% & 3.5\% & 1.0\% \\
$\tau_{\mathrm{ANY}}{=}0.30$ & 20.9\% & \textbf{12.5\%} & 4.8\% & 3.5\% & \textbf{0.8\%} \\
$\tau_{\mathrm{ANY}}{=}0.25$ & 20.5\% & 12.8\% & \textbf{4.5\%} & 3.7\% & 0.9\% \\
$\tau_{\mathrm{ANY}}{=}0.20$ & 20.3\% & 13.3\% & 4.9\% & 3.7\% & 0.9\% \\
$\tau_{\mathrm{ANY}}{=}0.15$ & 17.0\% & 13.7\% & 8.1\% & 3.8\% & 1.1\% \\
$\tau_{\mathrm{ANY}}{=}0.10$ & 15.8\% & 13.4\% & 10.9\% & 4.0\% & 1.3\% \\
$\tau_{\mathrm{ANY}}{=}0.05$ & 10.9\% & 13.5\% & 10.6\% & 4.2\% & 1.4\% \\
$\tau_{\mathrm{ANY}}{=}0.01$ & \textbf{8.4\%} & 13.9\% & 10.7\% & 4.6\% & 1.5\% \\
$\tau_{\mathrm{ANY}}{=}0.005$ & 8.6\% & 14.1\% & 11.1\% & 4.5\% & 1.5\% \\
\bottomrule
\end{tabular}}
\begin{tablenotes}[flushleft]\footnotesize
\item \textit{Note.} Medians/means are across numeric features. All values are reported in percent (\%).
\end{tablenotes}
\end{table}

Per-feature results confirm this pattern: many variables prefer
\(0.25\text{--}0.30\) for tail alignment, while very tight targets minimize center shift; full details appear in Table~\ref{tab:numeric-heuristic-best-per-feature}.

\begin{table}[t]
\centering
\footnotesize
\setlength{\tabcolsep}{3pt}
\resizebox{\columnwidth}{!}{%
\begin{threeparttable}
\caption{Per‐feature best datasets by absolute center shift \((|\Delta_{\mathrm{med}}|)\) and upper‑tail stretch \((|\Delta_{\mathrm{high}}|)\). Reported are the winning dataset (Baseline or \(\tau_{\mathrm{ANY}}=\cdot\)) and its value.}
\label{tab:numeric-heuristic-best-per-feature}

\begin{tabular}{l l S[table-format=2.1] l S[table-format=2.1]}
\toprule
& \multicolumn{2}{c}{Best center (median)} & \multicolumn{2}{c}{Best upper tail (95th)} \\
\cmidrule(lr){2-3}\cmidrule(lr){4-5}
Feature & Winning dataset & {$\Delta_{\mathrm{med}}$ (\%)} & Winning dataset & {$\Delta_{\mathrm{high}}$ (\%)} \\
\midrule
AGE         & \(\tau_{\mathrm{ANY}}{=}0.005\) & 0.0  & Baseline                & 3.8  \\
BILL\_AMT1  & \(\tau_{\mathrm{ANY}}{=}0.01\)  & -0.8 & \(\tau_{\mathrm{ANY}}{=}0.30\) & 20.6 \\
BILL\_AMT2  & \(\tau_{\mathrm{ANY}}{=}0.01\)  & -4.5 & \(\tau_{\mathrm{ANY}}{=}0.30\) & 8.8  \\
BILL\_AMT3  & \(\tau_{\mathrm{ANY}}{=}0.005\) & 2.3  & \(\tau_{\mathrm{ANY}}{=}0.30\) & 1.3  \\
BILL\_AMT4  & \(\tau_{\mathrm{ANY}}{=}0.01\)  & -8.7 & \(\tau_{\mathrm{ANY}}{=}0.30\) & 9.1  \\
BILL\_AMT5  & \(\tau_{\mathrm{ANY}}{=}0.005\) & -8.4 & \(\tau_{\mathrm{ANY}}{=}0.40\) & 1.3  \\
BILL\_AMT6  & \(\tau_{\mathrm{ANY}}{=}0.01\)  & -5.4 & \(\tau_{\mathrm{ANY}}{=}0.30\) & 15.5 \\
LIMIT\_BAL  & \(\tau_{\mathrm{ANY}}{=}0.25\)  & -20.2& \(\tau_{\mathrm{ANY}}{=}0.01\)  & -0.9 \\
PAY\_AMT1   & \(\tau_{\mathrm{ANY}}{=}0.01\)  & -2.5 & \(\tau_{\mathrm{ANY}}{=}0.20\) & -0.4 \\
PAY\_AMT2   & \(\tau_{\mathrm{ANY}}{=}0.01\)  & -8.4 & \(\tau_{\mathrm{ANY}}{=}0.01\)  & -5.4 \\
PAY\_AMT3   & \(\tau_{\mathrm{ANY}}{=}0.01\)  & -30.3& \(\tau_{\mathrm{ANY}}{=}0.10\) & -1.0 \\
PAY\_AMT4   & \(\tau_{\mathrm{ANY}}{=}0.01\)  & -11.7& \(\tau_{\mathrm{ANY}}{=}0.30\) & -1.0 \\
PAY\_AMT5   & \(\tau_{\mathrm{ANY}}{=}0.01\)  & -25.3& \(\tau_{\mathrm{ANY}}{=}0.01\)  & -9.8 \\
PAY\_AMT6   & \(\tau_{\mathrm{ANY}}{=}0.005\) & -43.5& \(\tau_{\mathrm{ANY}}{=}0.40\) & -0.0 \\
\bottomrule
\end{tabular}
\begin{tablenotes}[flushleft]\footnotesize
\item \textit{Note.} Negative values indicate synthetic quantiles slightly below the baseline reference; “best’’ is by absolute magnitude.
\end{tablenotes}
\end{threeparttable}}
\end{table}

Cohen’s \(d\) (reported in the main text) is consistently small (most \(|d|<0.2\)) yet informative across thresholds:
LIMIT\_BAL improves from \(d=0.162\) (CTGAN) to \(0.128\text{--}0.131\) at \(0.20\text{--}0.25\);
AGE is best at \(0.15\) (\(d=0.113\));
BILL\_AMT1 is best at \(0.30\) (\(d=0.033\)) and degrades when overly tight (\(0.05\text{--}0.01\));
PAY\_AMT variables improve monotonically with tightening (e.g., PAY\_AMT6 drops from \(d=0.030\) to \(0.00066\) at \(0.05\)).
Together with the quantile heuristics, this indicates that strict acceptance reduces central discrepancies but can distort tails, while \(\tau\!\approx\!0.25\) (similar at 0.20--0.30) offers the best overall balance: it minimizes categorical JS, keeps numeric tails close to real, and retains small effect sizes.

\subsubsection{Interpretation.}
The HEOM–kNN rejection-with-replacement rule surrounds dense regions (dominant categorical levels and numeric centers) with many small exclusion balls. Tightening the target therefore rejects samples near those regions more often, which reduces center shifts but can move probability mass into sparser tails. Moderate targets curb near-duplicates without re-inflating tails, explaining the empirical sweet spot at \(\tau\!\approx\!0.25\).

\subsection{Feature-level fidelity (credit TVAE)}

\subsubsection{Categorical marginals (log2FC \& JS)}
\begin{figure*}[t]
\centering
\includegraphics[width=\textwidth]{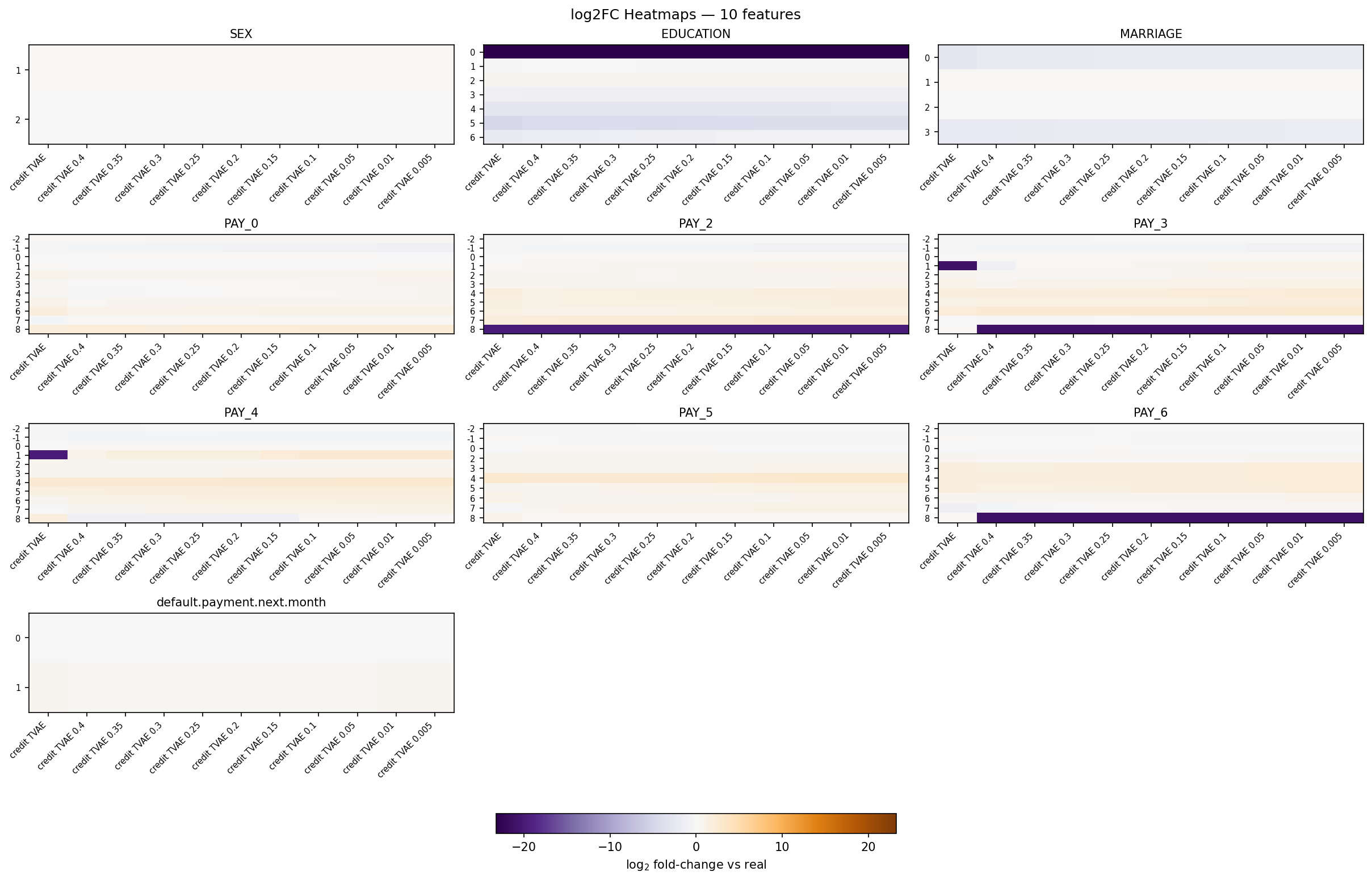}
\caption{Credit TVAE categorical fidelity across acceptance targets.
Each panel shows the log\(_2\) fold‐change \(\mathrm{log2FC}=\log_2(p_{\mathrm{synth}}/p_{\mathrm{real}})\) for every level of a categorical feature, comparing
the baseline TVAE sample and HEOM–kNN post‐filtered variants at \(\tauany \in \{0.40,0.35,\ldots,0.005\}\) (columns). Rows enumerate category levels in code order.
The colorbar encodes direction and magnitude (white \(\approx 0\); warm \(>\!0\) indicates over‐representation in the synthetic data; cool \(<\!0\) indicates under‐representation; extreme values are clipped for readability).
For TVAE, overall categorical divergence is already low and is lowest under a loose acceptance rule (\(\tauany\!=\!0.40\)); see Table~\ref{tab:cat-total-js-tvae}.
The main residual discrepancy is in EDUCATION, which shows re-weighting between levels 2–3 and deflation of rare codes. In contrast, repayment‐status features \(\mathrm{PAY}_0\text{–}\mathrm{PAY}_6\) remain close to real across targets, with only mild inflation of extreme tails as \(\tauany\) tightens.}
\label{fig:log2fc-heatmaps-tvae}
\end{figure*}

For credit TVAE, overall categorical divergence is low and is lowest under a loose acceptance target. Summing JS\_contrib across features (in bits, log base~2) yields a total JS of 0.0736 at TVAE 0.4 (an 11.2\% reduction vs.\ baseline \(0.0828\)); divergence worsens as the target tightens (e.g., \(0.1171\) at \(0.05\), \(0.1244\) at \(0.01\)); see Table~\ref{tab:cat-total-js-tvae}. Figure~\ref{fig:log2fc-heatmaps-tvae} visualizes log2FC by level and target.

\begin{table}[t]
\centering
\small
\setlength{\tabcolsep}{3.5pt}
\begin{threeparttable}
\caption{Credit TVAE: total categorical divergence (sum of per‐feature JS, \emph{log base 2; bits}) vs.\ acceptance target. Lower is better. $\Delta$ columns show the change relative to the baseline TVAE sample; negative values indicate improvement. The minimum occurs for a loose target (\(\tauany \!=\! 0.40\)).}
\label{tab:cat-total-js-tvae}
\begin{tabularx}{\columnwidth}{
  >{\raggedright\arraybackslash}X
  S[table-format=1.6]
  S[table-format=+1.4]
  S[table-format=+2.1]}
\toprule
Variant & {Total JS} & {$\Delta$ vs baseline} & {\% $\Delta$} \\
\midrule
credit TVAE         & 0.082808 & \multicolumn{1}{c}{—}     & \multicolumn{1}{c}{—} \\
credit TVAE 0.4     & \bfseries 0.073570 & \bfseries -0.0092 & \bfseries -11.2 \\
credit TVAE 0.35    & 0.079135 & -0.0037 & -4.4 \\
credit TVAE 0.3     & 0.082526 & -0.0003 & -0.3 \\
credit TVAE 0.25    & 0.087351 & +0.0045 & +5.5 \\
credit TVAE 0.2     & 0.093209 & +0.0104 & +12.6 \\
credit TVAE 0.15    & 0.098808 & +0.0160 & +19.3 \\
credit TVAE 0.1     & 0.107438 & +0.0246 & +29.7 \\
credit TVAE 0.05    & 0.117056 & +0.0342 & +41.4 \\
credit TVAE 0.01    & 0.124449 & +0.0416 & +50.3 \\
credit TVAE 0.005   & 0.125481 & +0.0427 & +51.5 \\
\bottomrule
\end{tabularx}
\begin{tablenotes}[flushleft]\footnotesize
\item \textit{Note.} “Total JS’’ sums per‐level Jensen–Shannon contributions within each categorical feature and then across features. All values are in \emph{bits} (log base~2).
\end{tablenotes}
\end{threeparttable}
\end{table}

\subsubsection{Numeric marginals (quantile heuristics and $d$)}
Dataset-level summaries (Table~\ref{tab:numeric-heuristic-summary-tvae}) show a center–tail trade-off distinct from CTGAN. The median absolute center shift $\mathrm{median}(|\Delta_{\mathrm{med}}|)$ is smallest at baseline (11.8\%) and grows as the target tightens, indicating that the HEOM–kNN acceptance step tends to worsen center alignment for TVAE. Upper-tail alignment $\mathrm{median}(|\Delta_{\mathrm{high}}|)$ improves with tightening (best at 0.1: 8.3\%), while the mean absolute upper-tail mass difference is smallest near 0.15 (1.43~pp). Low-tail mass differences change little across thresholds.

\begin{table}[t]
\centering
\small
\setlength{\tabcolsep}{2.5pt}
\caption{Credit TVAE: numeric marginal fidelity at \(q_L{=}0.05, q_H{=}0.95\). Reported are median absolute relative shifts (\%) and mean absolute tail‐mass deltas (\%). Lower is better; bold indicates the best per column.}
\label{tab:numeric-heuristic-summary-tvae}
\resizebox{\columnwidth}{!}{%
\begin{tabular}{lccccc}
\toprule
Dataset & Med.~$|\Delta_{\mathrm{med}}|$ & Med.~$|\Delta_{\mathrm{low}}|$ & Med.~$|\Delta_{\mathrm{high}}|$ & Mean~$|\Delta p_{\mathrm{low}}|$ & Mean~$|\Delta p_{\mathrm{high}}|$\\
\midrule
Baseline & \textbf{11.8\%} & 10.4\% & 15.1\% & 9.9\% & 1.6\% \\
$\tau_{\mathrm{ANY}}{=}0.40$ & 19.3\% & \textbf{9.9\%} & 13.3\% & 10.1\% & 1.5\% \\
$\tau_{\mathrm{ANY}}{=}0.35$ & 21.9\% & \textbf{9.9\%} & 12.7\% & 10.1\% & 1.4\% \\
$\tau_{\mathrm{ANY}}{=}0.30$ & 22.6\% & 10.4\% & 12.0\% & 10.1\% & 1.4\% \\
$\tau_{\mathrm{ANY}}{=}0.25$ & 22.6\% & 10.8\% & 11.3\% & 10.1\% & 1.4\% \\
$\tau_{\mathrm{ANY}}{=}0.20$ & 22.8\% & 11.0\% & 11.4\% & 10.1\% & 1.4\% \\
$\tau_{\mathrm{ANY}}{=}0.15$ & 22.4\% & 11.2\% & 9.9\% & 10.0\% & \textbf{1.4\%} \\
$\tau_{\mathrm{ANY}}{=}0.10$ & 23.4\% & 11.6\% & \textbf{8.3\%} & 10.0\% & 1.5\% \\
$\tau_{\mathrm{ANY}}{=}0.05$ & 24.2\% & 12.1\% & 8.8\% & 9.9\% & 1.5\% \\
$\tau_{\mathrm{ANY}}{=}0.01$ & 24.6\% & 12.2\% & 8.5\% & 9.9\% & 1.6\% \\
$\tau_{\mathrm{ANY}}{=}0.005$ & 24.6\% & 12.2\% & 8.4\% & \textbf{9.9\%} & 1.6\% \\
\bottomrule
\end{tabular}}
\begin{tablenotes}[flushleft]\footnotesize
\item \textit{Note.} Medians/means are across numeric features. All values are reported in percent (\%).
\end{tablenotes}
\end{table}

Per-feature “winners’’ (Table~\ref{tab:numeric-heuristic-best-per-feature-tvae}) and Cohen’s $d$ corroborate this: many numeric centers are best at baseline or looser targets. Examples (best $d$ per feature): LIMIT\_BAL (0.4, \(d = 0.010\)), AGE (0.2, \(0.0042\)), BILL\_AMT1 (baseline, \(0.0056\)), BILL\_AMT2 (0.4, \(0.0044\)), BILL\_AMT3 (0.35, \(0.0012\)), BILL\_AMT4 (0.3, \(0.0016\)), BILL\_AMT6 (0.1, \(0.0028\)); PAY\_AMT1 and PAY\_AMT2 are best at baseline (\(0.0535\) and \(0.0015\)), PAY\_AMT3 at 0.3 (\(0.00024\)), PAY\_AMT4 and PAY\_AMT6 at 0.4 (\(0.0128\), \(0.0118\)), and PAY\_AMT5 at baseline (\(0.0196\)).
Counting per-feature minima across the $d$ table gives: baseline (4 wins), 0.4 (5), 0.35 (1), 0.3 (2), 0.2 (1), 0.1 (1).

\begin{table}[t]
\centering
\scriptsize
\setlength{\tabcolsep}{3pt}
\caption{Credit TVAE: per-feature best dataset by absolute center shift \((|\Delta_{\mathrm{med}}|)\) and upper-tail stretch \((|\Delta_{\mathrm{high}}|)\). Reported are the winning dataset (Baseline or \(\tau_{\mathrm{ANY}}=\cdot\)) and its value.}
\label{tab:numeric-heuristic-best-per-feature-tvae}
\resizebox{\columnwidth}{!}{%
\begin{tabular}{l l S[table-format=2.1] l S[table-format=2.1]}
\toprule
& \multicolumn{2}{c}{Best center (median)} & \multicolumn{2}{c}{Best upper tail (95th)} \\
\cmidrule(lr){2-3}\cmidrule(lr){4-5}
Feature & Winning dataset & {$\Delta_{\mathrm{med}}$ (\%)} & Winning dataset & {$\Delta_{\mathrm{high}}$ (\%)} \\
\midrule
AGE         & \(\tau_{\mathrm{ANY}}{=}0.005\) & 0.0  & \(\tau_{\mathrm{ANY}}{=}0.40\)  & -7.5 \\
BILL\_AMT1  & Baseline                      & 11.4 & \(\tau_{\mathrm{ANY}}{=}0.20\) & -0.4 \\
BILL\_AMT2  & Baseline                      & 17.4 & \(\tau_{\mathrm{ANY}}{=}0.005\) & -4.3 \\
BILL\_AMT3  & Baseline                      & 12.3 & \(\tau_{\mathrm{ANY}}{=}0.005\) & -5.2 \\
BILL\_AMT4  & Baseline                      & 7.2  & \(\tau_{\mathrm{ANY}}{=}0.005\) & -5.8 \\
BILL\_AMT5  & Baseline                      & 8.2  & \(\tau_{\mathrm{ANY}}{=}0.005\) & -5.5 \\
BILL\_AMT6  & Baseline                      & 6.0  & \(\tau_{\mathrm{ANY}}{=}0.005\) & -9.3 \\
LIMIT\_BAL  & \(\tau_{\mathrm{ANY}}{=}0.40\) & -10.2 & \(\tau_{\mathrm{ANY}}{=}0.005\) & -1.7 \\
PAY\_AMT1   & Baseline                      & 28.9 & Baseline                      & 38.5 \\
PAY\_AMT2   & Baseline                      & 22.1 & \(\tau_{\mathrm{ANY}}{=}0.35\) & 23.5 \\
PAY\_AMT3   & \(\tau_{\mathrm{ANY}}{=}0.35\) & -1.2 & \(\tau_{\mathrm{ANY}}{=}0.35\) & 8.1 \\
PAY\_AMT4   & Baseline                      & 18.4 & \(\tau_{\mathrm{ANY}}{=}0.40\) & 25.0 \\
PAY\_AMT5   & Baseline                      & 26.2 & Baseline                      & 23.7 \\
PAY\_AMT6   & Baseline                      & 10.1 & \(\tau_{\mathrm{ANY}}{=}0.15\) & -0.5 \\
\bottomrule
\end{tabular}}
\begin{tablenotes}[flushleft]\footnotesize
\item \textit{Note.} Negative values indicate the synthetic quantile lies below the baseline reference; “best’’ is by absolute magnitude for each criterion.
\end{tablenotes}
\end{table}

\subsubsection{Interpretation.}
TVAE’s categorical marginals are already well-calibrated; the HEOM–kNN rejection-with-replacement rule—by rejecting candidates close to dense regions—can push mass away from those centers, slightly increasing categorical divergence (most visibly within EDUCATION) and worsening numeric centers as the target tightens. Moderate/loose targets preserve TVAE’s strengths while avoiding unnecessary drift.

\paragraph{Practical choice of target.}
For TVAE, a moderate/loose target (\mbox{$\tau \!\approx\! 0.30$–$0.40$}) offers the best overall balance: it minimizes categorical JS (0.051 at 0.4 vs.\ 0.057 baseline), keeps numeric centers close (many features’ $d$ are minimal at 0.3–0.4 or baseline), and yields competitive upper-tail alignment. If upper-tail fidelity on numerics is prioritized above all else, a somewhat tighter choice (\mbox{$\tau \!\approx\! 0.10$–$0.15$}) improves $\mathrm{median}(|\Delta_{\mathrm{high}}|)$ and $|\Delta p_{\mathrm{high}}|$, but at the cost of larger center shifts and slightly higher categorical drift.

\subsection{Feature-level fidelity (Adult CTGAN)}

\subsubsection{Categorical marginals (log2FC \& JS)}
\begin{figure*}[t]
\centering
\includegraphics[width=\textwidth]{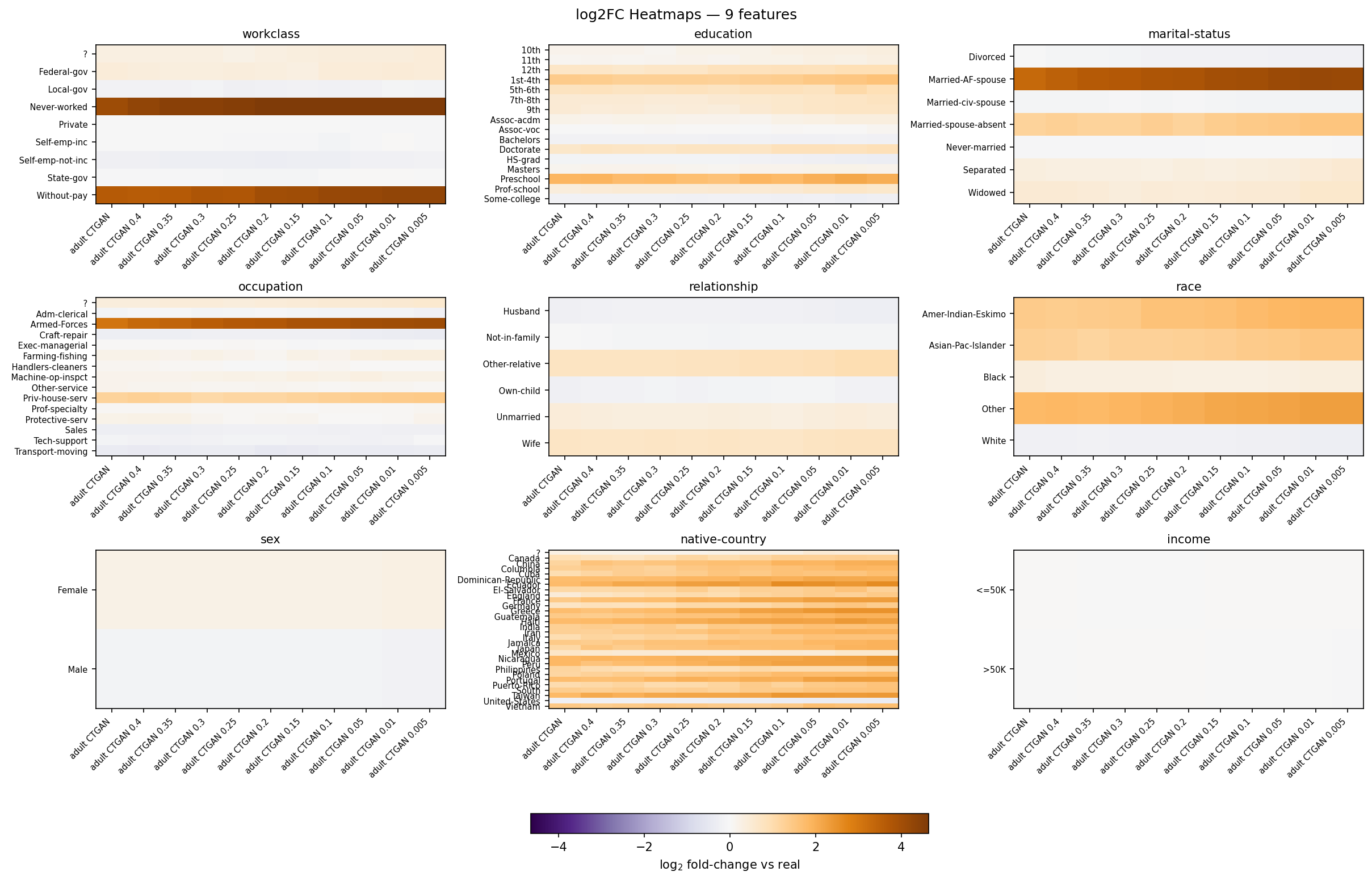}
\caption{Adult CTGAN categorical fidelity across acceptance targets.
Each panel shows the log\(_2\) fold‐change \(\mathrm{log2FC}=\log_2(p_{\mathrm{synth}}/p_{\mathrm{real}})\) for all levels of a categorical feature, comparing the baseline CTGAN sample and HEOM–kNN post‐filtered variants at \(\tau_{\mathrm{ANY}}\in\{0.40,0.35,\ldots,0.005\}\) (columns). Rows list category levels in code order. The color scale encodes direction and magnitude (white \(\approx 0\); warm \(>\!0\) over‐representation; cool \(<\!0\) under‐representation; extremes clipped for readability).
Adult CTGAN exhibits low—but non-trivial—drift that is minimized at moderate/loose targets (Table~\ref{tab:adult-cat-total-js}). The most persistent deviations arise in native-country (broad inflation of non-US categories and deflation of “United-States”) and race (inflation of minority groups with deflation of “White”); rare levels in marital-status and workclass (e.g., “Never-worked”) inflate with tighter acceptance. In contrast, income remains effectively unchanged.}
\label{fig:adult-log2fc-heatmaps}
\end{figure*}

Adult CTGAN exhibits low but non-trivial categorical drift that is minimized at moderate/loose acceptance.
Summing JS\_contrib across features (in bits, log base~2) yields total JS of
0.0753 at CTGAN 0.35 and 0.0765 at 0.40 (baseline: \(0.0770\)); divergence increases as the target tightens (e.g., \(0.1242\) at \(0.05\), \(0.1383\) at \(0.01\)).
The dominant baseline contributors are native-country (JS \(0.0238\)) and race (JS \(0.0182\)), then relationship (JS \(0.0090\)), education (JS \(0.0069\)), and marital-status (JS \(0.0063\)).
Representative level shifts (all from the JS\_contrib/log2FC table): 
native-country—United-States deflates (\(0.896 \rightarrow 0.774\), log2FC \( -0.211\)); several minorities inflate: Philippines (\(+1.175\)), Mexico (\(+0.678\)), Dominican-Republic (\(+1.710\)), Taiwan (\(+1.803\)), Jamaica (\(+1.435\)).
race—Asian-Pac-Islander inflates (\(0.0319 \rightarrow 0.0774\), log2FC \(+1.278\)), “Other” \(+1.799\), Amer-Indian-Eskimo \(+1.383\), while “White” deflates (log2FC \( -0.201\)).
marital-status—rare levels drift most (Married-AF-spouse \(+3.33\); Married-spouse-absent \(+1.21\)).
workclass—“Never-worked” grows from \(0.0002\) to \(0.0037\) (log2FC \(+4.09\)).
sex—Female is over-represented (\(0.331 \rightarrow 0.392\), log2FC \(+0.246\)).
Income is nearly unchanged (JS \(\approx 1.9\times10^{-5}\)).
These trends persist across thresholds (see Fig.~\ref{fig:adult-log2fc-heatmaps}); tightening generally amplifies minority/rare categories.

\begin{table}[t]
\centering
\small
\setlength{\tabcolsep}{3.5pt}
\begin{threeparttable}
\caption{Adult CTGAN: total categorical divergence (sum of per‐feature JS, \emph{log base 2; bits}) vs.\ acceptance target. Lower is better. The $\Delta$ columns show the change relative to the baseline CTGAN sample; negative values indicate improvement. The minimum occurs for a moderate/loose target (\(\tauany{=}0.35\)).}
\label{tab:adult-cat-total-js}
\begin{tabularx}{\columnwidth}{
  >{\raggedright\arraybackslash}X
  S[table-format=1.6]
  S[table-format=+1.5]
  S[table-format=+2.1]}
\toprule
Variant & {Total JS} & {$\Delta$ vs baseline} & {\% $\Delta$} \\
\midrule
adult CTGAN        & 0.076961 & \multicolumn{1}{c}{—}      & \multicolumn{1}{c}{—} \\
adult CTGAN 0.40   & 0.076485 & -0.00048 & -0.6 \\
adult CTGAN 0.35   & \bfseries 0.075325 & \bfseries -0.00164 & \bfseries-2.1 \\
adult CTGAN 0.30   & 0.079542 & +0.00258 & +3.4 \\
adult CTGAN 0.25   & 0.086695 & +0.00973 & +12.6 \\
adult CTGAN 0.20   & 0.091105 & +0.01414 & +18.4 \\
adult CTGAN 0.15   & 0.101621 & +0.02466 & +32.0 \\
adult CTGAN 0.10   & 0.111415 & +0.03445 & +44.8 \\
adult CTGAN 0.05   & 0.124179 & +0.04722 & +61.4 \\
adult CTGAN 0.01   & 0.138319 & +0.06136 & +79.7 \\
adult CTGAN 0.005  & 0.140706 & +0.06375 & +82.8 \\
\bottomrule
\end{tabularx}
\begin{tablenotes}[flushleft]\footnotesize
\item \textit{Note.} “Total JS’’ sums the per‐level Jensen–Shannon contributions within each categorical feature and then across features. All values are reported in \emph{bits} (log base~2).
\end{tablenotes}
\end{threeparttable}
\end{table}

\subsubsection{Numeric marginals (quantile heuristics and $d$)}
Dataset-level summaries (Table~\ref{tab:adult-numeric-heuristic-summary}) show that numeric centers and upper tails are remarkably stable for Adult CTGAN, while lower tails are the main source of mismatch.
The median absolute center shift $\mathrm{median}(|\Delta_{\mathrm{med}}|)$ is essentially zero at 0.10–0.05 and \(\approx 1.1\)–\(1.4\%\) elsewhere.
Upper-tail quantiles change minimally (best \(\mathrm{median}(|\Delta_{\mathrm{high}}|) = 0\%\) at baseline through \(0.20\)); mean absolute upper-tail mass difference is smallest around 0.25 (1.01~pp).
In contrast, the mean absolute lower-tail mass difference is sizeable (\(\approx 18\)~pp on average) and is slightly smallest at 0.35.

\begin{table}[t]
\centering
\small
\setlength{\tabcolsep}{2.5pt}
\caption{Adult CTGAN: numeric marginal fidelity at \(q_L{=}0.05, q_H{=}0.95\). Reported are median absolute relative shifts (\%) and mean absolute tail‐mass deltas (\%). Lower is better; bold indicates the best per column.}
\label{tab:adult-numeric-heuristic-summary}
\resizebox{\columnwidth}{!}{%
\begin{tabular}{lccccc}
\toprule
Dataset & Med.~$|\Delta_{\mathrm{med}}|$ & Med.~$|\Delta_{\mathrm{low}}|$ & Med.~$|\Delta_{\mathrm{high}}|$ & Mean~$|\Delta p_{\mathrm{low}}|$ & Mean~$|\Delta p_{\mathrm{high}}|$\\
\midrule
Baseline & 1.4\% & \textbf{9.2\%} & \textbf{0.0\%} & 18.1\% & 1.0\% \\
$\tau_{\mathrm{ANY}}{=}0.40$ & 1.3\% & 10.3\% & \textbf{0.0\%} & 18.2\% & 1.0\% \\
$\tau_{\mathrm{ANY}}{=}0.35$ & 1.1\% & 10.6\% & \textbf{0.0\%} & \textbf{18.1\%} & 1.1\% \\
$\tau_{\mathrm{ANY}}{=}0.30$ & 1.3\% & 10.5\% & \textbf{0.0\%} & 18.1\% & 1.0\% \\
$\tau_{\mathrm{ANY}}{=}0.25$ & 1.4\% & 10.9\% & \textbf{0.0\%} & 18.2\% & \textbf{1.0\%} \\
$\tau_{\mathrm{ANY}}{=}0.20$ & 1.3\% & 12.4\% & \textbf{0.0\%} & 18.2\% & 1.1\% \\
$\tau_{\mathrm{ANY}}{=}0.15$ & 1.4\% & 12.7\% & \textbf{0.0\%} & 18.3\% & 1.1\% \\
$\tau_{\mathrm{ANY}}{=}0.10$ & \textbf{0.0\%} & 12.2\% & 1.7\% & 18.4\% & 1.1\% \\
$\tau_{\mathrm{ANY}}{=}0.05$ & \textbf{0.0\%} & 12.6\% & 1.7\% & 18.7\% & 1.3\% \\
$\tau_{\mathrm{ANY}}{=}0.01$ & 1.4\% & 14.8\% & 1.7\% & 18.7\% & 1.3\% \\
$\tau_{\mathrm{ANY}}{=}0.005$ & 1.4\% & 13.0\% & 1.7\% & 18.7\% & 1.3\% \\
\bottomrule
\end{tabular}}
\begin{tablenotes}[flushleft]\footnotesize
\item \textit{Note.} Medians/means are across numeric features. All values are reported in percent (\%). Adult CTGAN shows very small center and upper-tail shifts; the dominant mismatch is in the lower-tail probabilities.
\end{tablenotes}
\end{table}

Per-feature “winners’’ (Table~\ref{tab:adult-numeric-heuristic-best-per-feature}) and the Cohen’s $d$ table point to moderate/loose targets as well:
age and fnlwgt are best near 0.35 ($d = 0.00225$ and $0.04826$);
education-num is best at 0.30 ($d = 0.0567$);
capital-gain improves monotonically with tightening, best at 0.01 ($d = 0.0491$);
capital-loss is best at 0.25 ($d = 0.0157$);
hours-per-week is best at baseline ($d = 0.0310$).
Overall, numeric centers are well-matched without aggressive rejection; tightening primarily affects tails.

\begin{table}[t]
\centering
\scriptsize
\setlength{\tabcolsep}{3pt}
\resizebox{\columnwidth}{!}{%
\begin{threeparttable}
\caption{Adult CTGAN: per‐feature best dataset by absolute center shift \((|\Delta_{\mathrm{med}}|)\) and upper‐tail stretch \((|\Delta_{\mathrm{high}}|)\). Reported are the winning dataset (Baseline or \(\tau_{\mathrm{ANY}}=\cdot\)) and its value.}
\label{tab:adult-numeric-heuristic-best-per-feature}
\begin{tabular}{l l S[table-format=2.1] l S[table-format=2.1]}
\toprule
& \multicolumn{2}{c}{Best center (median)} & \multicolumn{2}{c}{Best upper tail (95th)} \\
\cmidrule(lr){2-3}\cmidrule(lr){4-5}
Feature & Winning dataset & {$\Delta_{\mathrm{med}}$ (\%)} & Winning dataset & {$\Delta_{\mathrm{high}}$ (\%)} \\
\midrule
age             & \(\tau_{\mathrm{ANY}}{=}0.05\) & 0.0  & Baseline & 0.0 \\
capital-gain    & Baseline & \multicolumn{1}{c}{n/a\tnote{*}} & \(\tau_{\mathrm{ANY}}{=}0.15\) & 0.7 \\
capital-loss    & Baseline & \multicolumn{1}{c}{n/a\tnote{*}} & Baseline & \multicolumn{1}{c}{n/a\tnote{*}} \\
education-num   & \(\tau_{\mathrm{ANY}}{=}0.30\) & 0.0  & \(\tau_{\mathrm{ANY}}{=}0.35\) & 0.0 \\
fnlwgt          & \(\tau_{\mathrm{ANY}}{=}0.35\) & 2.1  & Baseline & 4.2 \\
hours-per-week  & \(\tau_{\mathrm{ANY}}{=}0.005\) & 0.0  & Baseline & 0.0 \\
\bottomrule
\end{tabular}
\begin{tablenotes}[flushleft]\footnotesize
\item[*] \textit{Undefined} relative shift because the baseline reference quantile at that position equals zero (e.g., \(Q_{50}^B{=}0\) or \(Q_{95}^B{=}0\)); ratios are indeterminate but effect sizes (Cohen’s \(d\)) remain interpretable.
\end{tablenotes}
\end{threeparttable}}
\end{table}

\subsubsection{Interpretation.}
The HEOM–kNN rejection rule removes candidates near dense regions (dominant categories and numeric centers). On Adult, tightening $\tau$ therefore re-weights mass toward minority/rare categories (native-country, race, some marital-status levels) and has little to gain for already-stable numeric centers and upper tails, while slightly worsening lower-tail coverage.

\paragraph{Practical choice of target.}
A moderate acceptance (\mbox{$\tau \!\approx\! 0.35$}, similar at \(0.40\)) offers the best overall balance: it minimizes total categorical JS (0.0522–0.0530 vs.\ 0.0533 baseline), preserves numeric centers and upper tails, and avoids over-amplifying rare categories. If one specifically wants to improve the upper tail of capital-gain, a tighter setting (\(\tau \approx 0.01\)) helps there—but at the cost of higher categorical drift and slightly worse lower-tail alignment.

\subsection{Feature-level fidelity (Adult TVAE)}

\subsubsection{Categorical marginals (log2FC \& JS)}
\begin{figure*}[t]
\centering
\includegraphics[width=\textwidth]{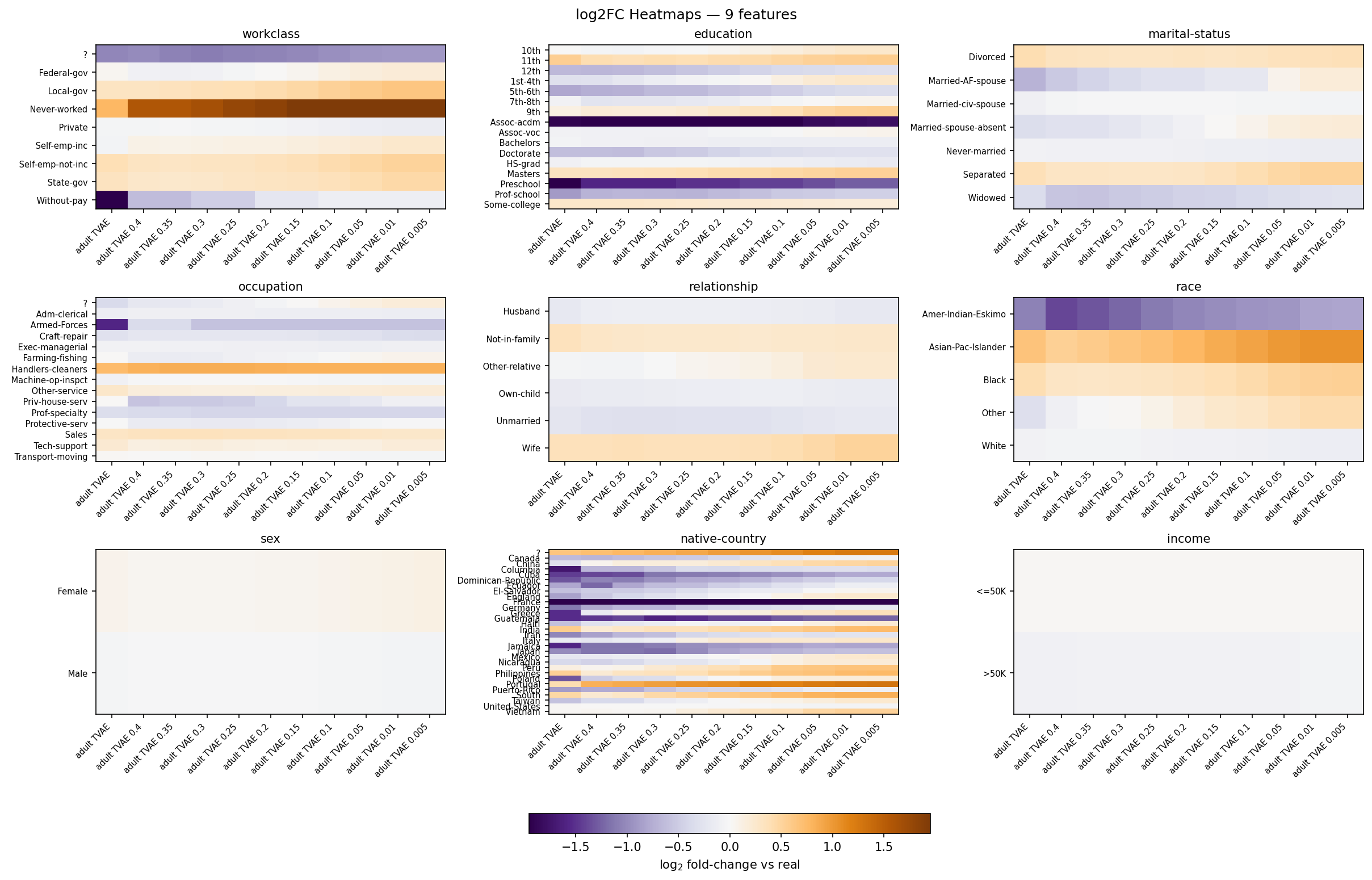}
\caption{Adult TVAE categorical fidelity across acceptance targets.
Each panel shows the log\(_2\) fold‐change \(\mathrm{log2FC}=\log_2(p_{\mathrm{synth}}/p_{\mathrm{real}})\) for every level of a categorical feature, comparing the baseline TVAE sample and HEOM–kNN post‐filtered variants at \(\tau_{\mathrm{ANY}}\in\{0.40,0.35,\ldots,0.005\}\) (columns). Rows list category levels in code order. The colorbar encodes direction and magnitude (white \(\approx 0\); warm \(>\!0\) over‐representation; cool \(<\!0\) under‐representation; extremes clipped for readability).
Adult TVAE exhibits low categorical drift that is lowest under a loose acceptance (\(\tau_{\mathrm{ANY}}{=}0.40\), Table~\ref{tab:adult-tvae-cat-total-js}). Residual discrepancies concentrate in education, occupation, workclass, and relationship; e.g., deflation of \emph{Assoc‐acdm} within \emph{education}, and reduced “?” in \emph{workclass}. The direction of these shifts is stable across targets: magnitudes shrink for looser settings (0.30–0.40) and grow as \(\tau\) tightens.}
\label{fig:adult-tvae-log2fc-heatmaps}
\end{figure*}

Adult TVAE has low categorical drift that is lowest under a loose acceptance. Summing JS\_contrib across features gives total JS 0.0371 at TVAE 0.4, a 13.7\% reduction vs.\ baseline (0.0430), and progressively worse values as the target tightens (e.g., 0.0480 at 0.05; 0.0533 at 0.01); see Table~\ref{tab:adult-tvae-cat-total-js}.
Across thresholds (Fig.~\ref{fig:adult-tvae-log2fc-heatmaps}), residual discrepancies concentrate in education, occupation, workclass, and relationship; e.g., deflation of \emph{Assoc‐acdm} within \emph{education}, and reduced “?” in \emph{workclass}. The direction of these shifts is stable, with magnitudes shrinking at looser targets (0.30–0.40) and growing as the target tightens.

\begin{table}[t]
\centering
\small
\setlength{\tabcolsep}{3.5pt}
\begin{threeparttable}
\caption{Adult TVAE: total categorical divergence (sum of per‐feature JS, \emph{log base 2; bits}) vs.\ acceptance target. Lower is better. The $\Delta$ columns show change relative to the baseline TVAE sample; negative values indicate improvement. The minimum occurs for a loose target (\(\tauany{=}0.40\)).}
\label{tab:adult-tvae-cat-total-js}
\begin{tabularx}{\columnwidth}{
  >{\raggedright\arraybackslash}X
  S[table-format=1.6]
  S[table-format=+1.4]
  S[table-format=+2.1]}
\toprule
Variant & {Total JS} & {$\Delta$ vs baseline} & {\% $\Delta$} \\
\midrule
adult TVAE        & 0.042966 & \multicolumn{1}{c}{—}     & \multicolumn{1}{c}{—} \\
adult TVAE 0.40   & \bfseries 0.037100 & \bfseries -0.0059 & \bfseries -13.7 \\
adult TVAE 0.35   & 0.037371 & -0.0056 & -13.0 \\
adult TVAE 0.30   & 0.037275 & -0.0057 & -13.2 \\
adult TVAE 0.25   & 0.037230 & -0.0057 & -13.4 \\
adult TVAE 0.20   & 0.038318 & -0.0046 & -10.8 \\
adult TVAE 0.15   & 0.040310 & -0.0027 & -6.2 \\
adult TVAE 0.10   & 0.043489 & +0.0005 & +1.2 \\
adult TVAE 0.05   & 0.048031 & +0.0051 & +11.8 \\
adult TVAE 0.01   & 0.053265 & +0.0103 & +24.0 \\
adult TVAE 0.005  & 0.053870 & +0.0109 & +25.4 \\
\bottomrule
\end{tabularx}
\begin{tablenotes}[flushleft]\footnotesize
\item \textit{Note.} “Total JS’’ sums per‐level Jensen–Shannon contributions within each categorical feature and then across features. All values are in \emph{bits} (log base~2).
\end{tablenotes}
\end{threeparttable}
\end{table}

\subsubsection*{Numeric marginals (quantile heuristics and $d$)}
Dataset-level summaries (Table~\ref{tab:adult-tvae-numeric-heuristic-summary}) show that Adult TVAE’s centers are essentially matched (\(\mathrm{median}(|\Delta_{\mathrm{med}}|) = 0\%\) for all but the tightest targets), upper tails change little (median \(|\Delta_{\mathrm{high}}| \approx 0.95\)–\(1.67\%\)), and the lower tail is the main source of mismatch (mean \(|\Delta p_{\mathrm{low}}| \approx 10\)–\(10.7\)~pp). The smallest lower-tail difference occurs at 0.4 (10.16~pp), while the smallest upper-tail mass difference is around 0.25–0.30 (about 1.00–1.01~pp). Very tight targets (0.01–0.005) increase both tail deltas.

\begin{table}[t]
\centering
\small
\setlength{\tabcolsep}{2.5pt}
\caption{Adult TVAE: numeric marginal fidelity at \(q_L{=}0.05, q_H{=}0.95\). Reported are median absolute relative shifts (\%) and mean absolute tail‐mass deltas (\%). Lower is better; bold indicates the best per column.}
\label{tab:adult-tvae-numeric-heuristic-summary}
\resizebox{\columnwidth}{!}{%
\begin{tabular}{lccccc}
\toprule
Dataset & Med.~$|\Delta_{\mathrm{med}}|$ & Med.~$|\Delta_{\mathrm{low}}|$ & Med.~$|\Delta_{\mathrm{high}}|$ & Mean~$|\Delta p_{\mathrm{low}}|$ & Mean~$|\Delta p_{\mathrm{high}}|$\\
\midrule
Baseline & \textbf{0.0\%} & 15.3\% & 1.7\% & \textbf{10.2\%} & 1.1\% \\
$\tau_{\mathrm{ANY}}{=}0.40$ & \textbf{0.0\%} & 13.8\% & 1.7\% & \textbf{10.2\%} & 1.1\% \\
$\tau_{\mathrm{ANY}}{=}0.35$ & \textbf{0.0\%} & 13.4\% & 1.7\% & \textbf{10.2\%} & \textbf{1.0\%} \\
$\tau_{\mathrm{ANY}}{=}0.30$ & \textbf{0.0\%} & 16.1\% & 1.7\% & \textbf{10.2\%} & \textbf{1.0\%} \\
$\tau_{\mathrm{ANY}}{=}0.25$ & \textbf{0.0\%} & 15.8\% & 1.6\% & 10.3\% & \textbf{1.0\%} \\
$\tau_{\mathrm{ANY}}{=}0.20$ & \textbf{0.0\%} & 15.3\% & 1.3\% & 10.3\% & \textbf{1.0\%} \\
$\tau_{\mathrm{ANY}}{=}0.15$ & \textbf{0.0\%} & 11.6\% & \textbf{1.0\%} & 10.3\% & 1.1\% \\
$\tau_{\mathrm{ANY}}{=}0.10$ & \textbf{0.0\%} & 11.2\% & 1.6\% & 10.4\% & 1.1\% \\
$\tau_{\mathrm{ANY}}{=}0.05$ & \textbf{0.0\%} & 11.1\% & 1.6\% & 10.6\% & 1.2\% \\
$\tau_{\mathrm{ANY}}{=}0.01$ & 0.3\% & 10.4\% & 1.6\% & 10.7\% & 1.4\% \\
$\tau_{\mathrm{ANY}}{=}0.005$ & 0.3\% & \textbf{10.3\%} & 1.6\% & 10.7\% & 1.4\% \\
\bottomrule
\end{tabular}}
\begin{tablenotes}[flushleft]\footnotesize
\item \textit{Note.} Medians/means are across numeric features. All values are reported in percent (\%). For Adult TVAE, centers are essentially matched, upper tails change little, and lower-tail mass differences dominate.
\end{tablenotes}
\end{table}

Cohen’s $d$ aligns with this picture (table provided in the main text): 
\emph{age} improves with moderate loosening (best around 0.30–0.35, \(d \approx 0.005\)–\(0.006\)) and worsens as the target tightens; 
\emph{fnlwgt} is smallest at baseline/0.4 (0.0085/0.0077); 
\emph{education-num} increases monotonically with tightening; 
\emph{capital-gain} steadily improves as the target tightens (baseline 0.146 \(\rightarrow\) 0.128 at 0.005); 
\emph{capital-loss} moves the other way (baseline 0.124 \(\rightarrow\) 0.190 at 0.005); 
\emph{hours-per-week} gradually improves as the target tightens to 0.05–0.01 (\(\sim\)0.018–0.019).

\begin{table}[t]
\centering
\scriptsize
\setlength{\tabcolsep}{3pt}
\resizebox{\columnwidth}{!}{%
\begin{threeparttable}
\caption{Adult TVAE: per‐feature best dataset by absolute center shift \((|\Delta_{\mathrm{med}}|)\) and upper‐tail stretch \((|\Delta_{\mathrm{high}}|)\). Reported are the winning dataset (Baseline or \(\tau_{\mathrm{ANY}}=\cdot\)) and its value.}
\label{tab:adult-tvae-numeric-heuristic-best-per-feature}
\begin{tabular}{l l S[table-format=+2.1] l S[table-format=+2.1]}
\toprule
& \multicolumn{2}{c}{Best center (median)} & \multicolumn{2}{c}{Best upper tail (95th)} \\
\cmidrule(lr){2-3}\cmidrule(lr){4-5}
Feature & Winning dataset & {$\Delta_{\mathrm{med}}$ (\%)} & Winning dataset & {$\Delta_{\mathrm{high}}$ (\%)} \\
\midrule
age             & Baseline & 0.0  & \(\tau_{\mathrm{ANY}}{=}0.15\) & 0.0 \\
capital-gain    & Baseline & \multicolumn{1}{c}{n/a\tnote{*}} & \(\tau_{\mathrm{ANY}}{=}0.005\) & -38.0 \\
capital-loss    & Baseline & \multicolumn{1}{c}{n/a\tnote{*}} & Baseline & \multicolumn{1}{c}{n/a\tnote{*}} \\
education-num   & \(\tau_{\mathrm{ANY}}{=}0.35\) & 0.0  & \(\tau_{\mathrm{ANY}}{=}0.20\) & 0.0 \\
fnlwgt          & \(\tau_{\mathrm{ANY}}{=}0.15\) & 0.2  & \(\tau_{\mathrm{ANY}}{=}0.05\) & -0.2 \\
hours-per-week  & \(\tau_{\mathrm{ANY}}{=}0.40\) & 0.0  & \(\tau_{\mathrm{ANY}}{=}0.30\) & 1.7 \\
\bottomrule
\end{tabular}
\begin{tablenotes}[flushleft]\footnotesize
\item[*] \textit{Undefined} relative shift because the baseline reference quantile at that position equals zero (e.g., \(Q_{50}^B{=}0\) or \(Q_{95}^B{=}0\)); ratios are indeterminate but standardized effects (Cohen’s \(d\)) remain interpretable.
\end{tablenotes}
\end{threeparttable}}
\end{table}

\paragraph{Interpretation.}
TVAE’s numeric centers are already aligned; the HEOM–kNN rejection step mostly re-weights tails and moderately shifts certain categorical levels (especially within \emph{education}, \emph{workclass}, \emph{relationship}). Looser acceptance (\(\tau \approx 0.30\!-\!0.40\)) preserves the model’s native calibration—categorical JS is lowest, center shifts are near zero, and tail deltas are smaller—whereas very tight targets trade lower capital-gain $d$ for worse categorical drift and heavier tail deltas (notably for capital-loss).

\paragraph{Practical choice of target.}
For Adult TVAE, a moderate/loose target (\mbox{$\tau \!\approx\! 0.30$–$0.40$}) offers the best overall balance: it minimizes categorical JS (e.g., 0.0257 at 0.4, \(-13.7\%\) vs.\ baseline), keeps numeric centers essentially perfect, and yields smaller tail-mass differences (upper tail best near 0.25–0.30; lower tail best at 0.4). If upper-tail matching for \emph{capital-gain} dominates the objective, one may tighten further (\(\tau \le 0.05\)), at the cost of higher categorical drift and larger discrepancies elsewhere.

\paragraph{Bridge to the numeric heuristics.}
We interpret \emph{Cohen’s $d$} as an overall standardized location shift, while the numeric heuristic’s \emph{relative quantile shifts} \((\Delta_{\mathrm{med}}, \Delta_{\mathrm{low}}, \Delta_{\mathrm{high}})\) and \emph{tail-mass deltas} \((\Delta p_{\mathrm{low}}, \Delta p_{\mathrm{high}})\)—evaluated at the \emph{baseline} 5th/95th-percentile cutpoints—localize where the distribution differs (center vs.\ tails) and by how much in multiplicative terms; thus small $|d|$ with $\Delta_{\mathrm{high}} > 0$ or $\Delta p_{\mathrm{high}} > 0$ flags heavier upper tails despite a stable mean, whereas large $|d|$ with near-zero tail deltas indicates a primarily central shift. On Adult TVAE, this lens explains why 0.30–0.40 balances categorical and numeric fidelity.

\subsection{Feature-level fidelity (Cardio CTGAN)}

\subsubsection{Categorical marginals (log2FC \& JS)}

\begin{figure*}[t]
\centering
\includegraphics[width=\textwidth]{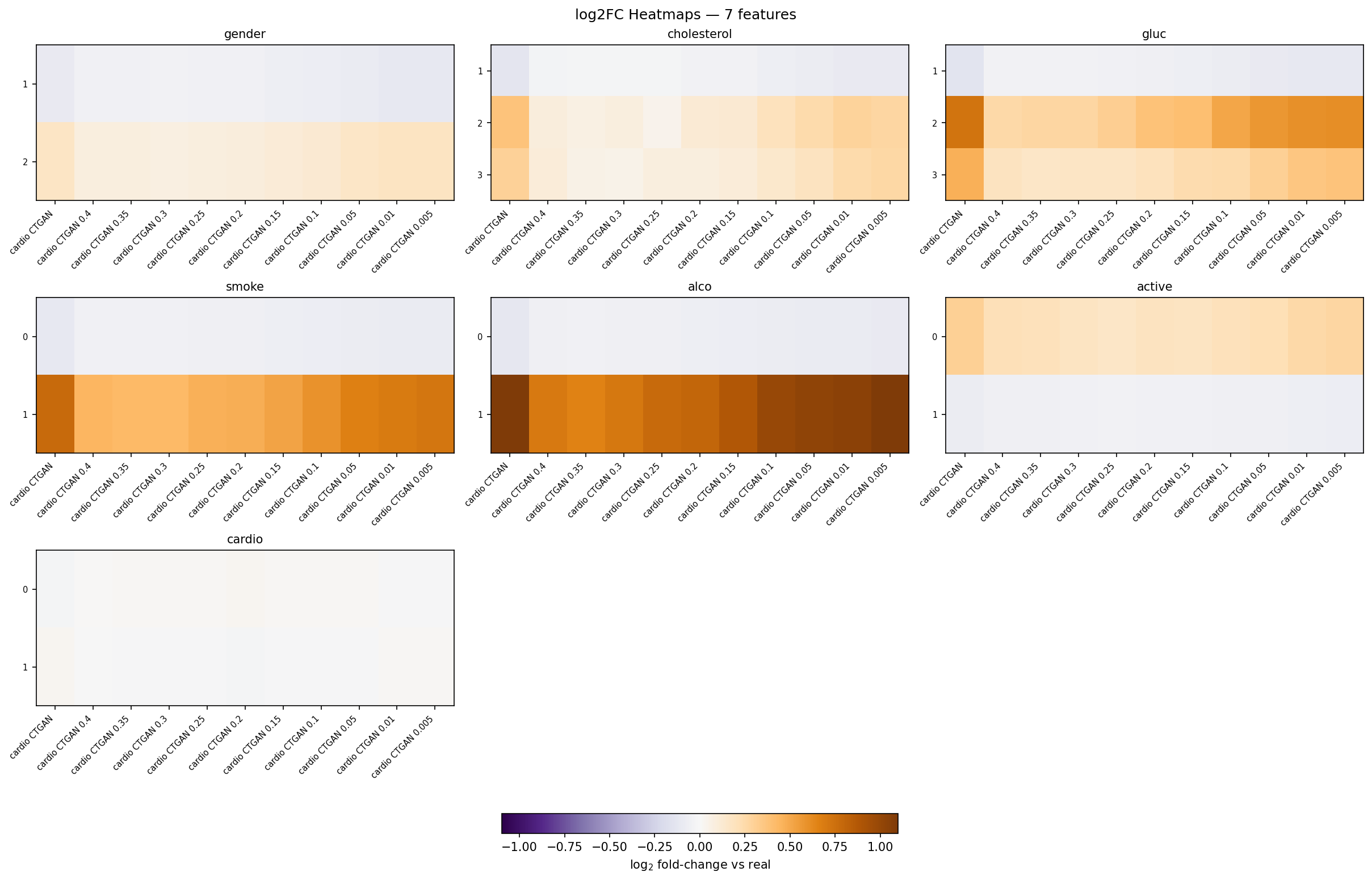}
\caption{Cardio CTGAN categorical fidelity across acceptance targets.
Each panel shows the log\(_2\) fold‐change \(\mathrm{log2FC}=\log_2(p_{\mathrm{synth}}/p_{\mathrm{real}})\) for every level of a categorical feature, comparing the baseline CTGAN sample and HEOM–kNN post‐filtered variants at \(\tau_{\mathrm{ANY}}\in\{0.40,0.35,\ldots,0.005\}\) (columns). Rows list category levels in code order. The colorbar encodes direction and magnitude (white \(\approx 0\); warm \(>\!0\) over‐representation; cool \(<\!0\) under‐representation; extremes clipped for readability).
Drift concentrates in alco, smoke, and gluc (and, to a lesser extent, cholesterol and active). Moderate acceptance (\(\tau_{\mathrm{ANY}}\!\approx\!0.30\text{–}0.35\)) visibly damps the inflation of \emph{alco=1}, \emph{smoke=1}, and elevated \emph{gluc}/\emph{cholesterol} categories, consistent with the large drop in total JS in Table~\ref{tab:cardio-ctgan-cat-total-js}. The label \emph{cardio} remains essentially unchanged across targets.}
\label{fig:cardio-ctgan-log2fc-heatmaps}
\end{figure*}

Categorical drift concentrates in alco, gluc, smoke, and (to a lesser extent) cholesterol and active. Summing JS\_contrib across features yields total JS 0.00670 at CTGAN 0.35, a 79.9\% reduction vs.\ baseline (\(0.03335\)); divergence rises again as the target tightens (e.g., \(0.02030\) at \(0.05\); \(0.02323\) at \(0.01\)). See Table~\ref{tab:cardio-ctgan-cat-total-js} and Fig.~\ref{fig:cardio-ctgan-log2fc-heatmaps}.

\begin{table}[t]
\centering
\small
\setlength{\tabcolsep}{3.5pt}
\begin{threeparttable}
\caption{Cardio CTGAN: total categorical divergence (sum of per‐feature JS, \emph{log base 2; bits}) vs.\ acceptance target. Lower is better. The $\Delta$ columns show change relative to the baseline CTGAN sample; negative values indicate improvement. The minimum occurs for a moderate target (\(\tauany{=}0.35\)).}
\label{tab:cardio-ctgan-cat-total-js}
\begin{tabularx}{\columnwidth}{
  >{\raggedright\arraybackslash}X
  S[table-format=1.6]
  S[table-format=+1.6]
  S[table-format=+2.1]}
\toprule
Variant & {Total JS} & {$\Delta$ vs baseline} & {\% $\Delta$} \\
\midrule
cardio CTGAN        & 0.033345 & \multicolumn{1}{c}{—}       & \multicolumn{1}{c}{—} \\
cardio CTGAN 0.40   & 0.007542 & -0.025803 & -77.4 \\
cardio CTGAN 0.35   & \bfseries 0.006697 & \bfseries -0.026648 & \bfseries -79.9 \\
cardio CTGAN 0.30   & 0.006984 & -0.026361 & -79.1 \\
cardio CTGAN 0.25   & 0.008330 & -0.025014 & -75.0 \\
cardio CTGAN 0.20   & 0.009708 & -0.023636 & -70.9 \\
cardio CTGAN 0.15   & 0.011641 & -0.021704 & -65.1 \\
cardio CTGAN 0.10   & 0.016311 & -0.017033 & -51.1 \\
cardio CTGAN 0.05   & 0.020295 & -0.013049 & -39.1 \\
cardio CTGAN 0.01   & 0.023232 & -0.010112 & -30.3 \\
cardio CTGAN 0.005  & 0.025147 & -0.008198 & -24.6 \\
\bottomrule
\end{tabularx}
\begin{tablenotes}[flushleft]\footnotesize
\item \textit{Note.} “Total JS’’ sums per‐level Jensen–Shannon contributions within each categorical feature and then across features. All values are in \emph{bits} (log base~2).
\end{tablenotes}
\end{threeparttable}
\end{table}

\subsubsection{Numeric marginals (quantile heuristics and $d$)}
Aggregate heuristics (Table~\ref{tab:cardio-ctgan-numeric-heuristic-summary}) show a clear \emph{sweet spot} around 0.30–0.35: 
the upper-tail relative shift is smallest at 0.30 (median \(|\Delta_{\mathrm{high}}| = 0.56\%\)), and both tail-mass deltas are near their minima there (mean \(|\Delta p_{\mathrm{low}}|\) 3.27~pp, \(|\Delta p_{\mathrm{high}}|\) \(1.19\)~pp).
The center is already close to real and improves modestly with tightening, reaching its minimum at 0.20 (median \(|\Delta_{\mathrm{med}}| = 0.53\%\)).
Lower-tail quantiles are best at baseline (median \(|\Delta_{\mathrm{low}}| = 1.33\%\)) and drift upward as the target tightens.

\begin{table}[t]
\centering
\small
\setlength{\tabcolsep}{2.5pt}
\caption{Cardio CTGAN: numeric marginal fidelity at \(q_L{=}0.05, q_H{=}0.95\). Reported are median absolute relative shifts (\%) and mean absolute tail‐mass deltas (\%). Lower is better; bold indicates the best per column.}
\label{tab:cardio-ctgan-numeric-heuristic-summary}
\resizebox{\columnwidth}{!}{%
\begin{tabular}{lccccc}
\toprule
Dataset & Med.~$|\Delta_{\mathrm{med}}|$ & Med.~$|\Delta_{\mathrm{low}}|$ & Med.~$|\Delta_{\mathrm{high}}|$ & Mean~$|\Delta p_{\mathrm{low}}|$ & Mean~$|\Delta p_{\mathrm{high}}|$\\
\midrule
Baseline & 0.9\% & \textbf{1.3\%} & 0.8\% & 3.5\% & \textbf{1.0\%} \\
$\tau_{\mathrm{ANY}}{=}0.40$ & 0.7\% & 1.7\% & 1.1\% & 3.4\% & 1.3\% \\
$\tau_{\mathrm{ANY}}{=}0.35$ & 0.6\% & 1.7\% & 1.1\% & 3.3\% & 1.3\% \\
$\tau_{\mathrm{ANY}}{=}0.30$ & 0.6\% & 1.7\% & \textbf{0.6\%} & \textbf{3.3\%} & 1.2\% \\
$\tau_{\mathrm{ANY}}{=}0.25$ & 0.7\% & 1.8\% & \textbf{0.6\%} & 3.3\% & 1.2\% \\
$\tau_{\mathrm{ANY}}{=}0.20$ & \textbf{0.5\%} & 1.9\% & \textbf{0.6\%} & 3.3\% & 1.2\% \\
$\tau_{\mathrm{ANY}}{=}0.15$ & 0.6\% & 2.3\% & 1.0\% & 3.3\% & 1.2\% \\
$\tau_{\mathrm{ANY}}{=}0.10$ & 0.7\% & 2.3\% & 1.2\% & 3.4\% & 1.3\% \\
$\tau_{\mathrm{ANY}}{=}0.05$ & 0.7\% & 3.2\% & 1.2\% & 3.4\% & 1.3\% \\
$\tau_{\mathrm{ANY}}{=}0.01$ & 0.8\% & 3.2\% & 1.2\% & 3.4\% & 1.3\% \\
\bottomrule
\end{tabular}}
\begin{tablenotes}[flushleft]\footnotesize
\item \textit{Note.} Medians/means are across numeric features. All values are reported in percent (\%). Cardio CTGAN shows a sweet spot near \(\tau_{\mathrm{ANY}} \approx 0.30\!-\!0.35\): smallest upper-tail shift and low tail-mass deltas, with center shifts already small (best at \(\tau_{\mathrm{ANY}}{=}0.20\)).
\end{tablenotes}
\end{table}

Per-feature winners (Table~\ref{tab:cardio-ctgan-numeric-heuristic-best-per-feature}) echo this: some features favor 0.20 for center alignment while 0.30 minimizes upper-tail distortion.

\begin{table}[t]
\centering
\scriptsize
\setlength{\tabcolsep}{3pt}
\resizebox{\columnwidth}{!}{%
\begin{threeparttable}
\caption{Cardio CTGAN: per‐feature best dataset by absolute center shift \((|\Delta_{\mathrm{med}}|)\) and upper‐tail stretch \((|\Delta_{\mathrm{high}}|)\). Reported are the winning dataset (Baseline or \(\tau_{\mathrm{ANY}}=\cdot\)) and its value.}
\label{tab:cardio-ctgan-numeric-heuristic-best-per-feature}
\begin{tabular}{l l S[table-format=+2.1] l S[table-format=+2.1]}
\toprule
& \multicolumn{2}{c}{Best center (median)} & \multicolumn{2}{c}{Best upper tail (95th)} \\
\cmidrule(lr){2-3}\cmidrule(lr){4-5}
Feature & Winning dataset & {$\Delta_{\mathrm{med}}$ (\%)} & Winning dataset & {$\Delta_{\mathrm{high}}$ (\%)} \\
\midrule
age     & \(\tau_{\mathrm{ANY}}{=}0.20\) & 0.5  & Baseline & 0.0 \\
ap\_hi  & \(\tau_{\mathrm{ANY}}{=}0.005\) & 0.8  & \(\tau_{\mathrm{ANY}}{=}0.30\) & -1.2 \\
ap\_lo  & \(\tau_{\mathrm{ANY}}{=}0.01\)  & 2.5  & Baseline & 2.0 \\
height  & \(\tau_{\mathrm{ANY}}{=}0.005\) & 0.0  & Baseline & 0.0 \\
weight  & \(\tau_{\mathrm{ANY}}{=}0.20\) & 0.0  & \(\tau_{\mathrm{ANY}}{=}0.25\) & -0.1 \\
\bottomrule
\end{tabular}
\begin{tablenotes}[flushleft]\footnotesize
\item \textit{Note.} Signs reflect the direction of the relative quantile shift: negative values indicate the synthetic quantile lies below the baseline reference. “Best’’ is by absolute magnitude for each criterion.
\end{tablenotes}
\end{threeparttable}}
\end{table}

\paragraph{Cohen’s $d$ (numeric effect sizes).}
Effect sizes are uniformly small.  
\emph{age} improves from \(d = 0.0478\) (baseline) to 0.00457 at 0.05 (similar at 0.005);  
\emph{height} is best at 0.35 (\(d = 0.0874\));  
\emph{weight} is best at 0.20 (\(d = 0.00019\));  
\emph{ap\_hi} is essentially stable (0.008–0.012 across targets, best near baseline/0.01);  
\emph{ap\_lo} is best near 0.30 (\(d = 0.00178\)) but worsens for very tight targets (0.038 at 0.01).

\paragraph{Interpretation.}
The HEOM–kNN acceptance rule preferentially rejects candidates near dense regions (e.g., non-alcoholic, non-smoker, normal glucose/cholesterol), which initially pushed mass toward \emph{alco=1}, \emph{smoke=1}, and elevated \emph{gluc}/\emph{cholesterol} categories. Moderate acceptance (\(\tau \approx 0.30\!-\!0.35\)) curbs these re-weightings (large drop in total JS) while keeping numeric centers and tails well aligned. Over-tightening shifts probability into tails and re-inflates categorical imbalances.

\paragraph{Practical choice of target.}
A moderate target (\mbox{$\tau \!\approx\! 0.30$–$0.35$}) offers the best overall balance on Cardio CTGAN: it reduces categorical JS by \(\sim\)80\% vs.\ baseline and simultaneously gives the best tail alignment (both tail masses and upper-tail quantiles), while keeping centers very close. If a specific application values center matching above all else for \emph{weight} and \emph{age}, 0.20 further minimizes median shifts and $d$ for those features, at the cost of slightly larger tail deltas.

\subsection{Feature-level fidelity (Cardio TVAE)}

\subsubsection{Categorical marginals (log2FC \& JS)}
Cardio TVAE starts with very low categorical divergence and tightening/loosening the acceptance rule generally \emph{increases} drift.
Summing JS\_contrib across features (in bits, log base~2) yields total JS 0.001889 at baseline, rising to 0.013676 at 0.40 (+623.8\%) and to 0.002261 at 0.01 (+19.7\%); see Table~\ref{tab:cardio-tvae-cat-total-js}.
Drift concentrates in alco (dominant), then smoke, active, cholesterol, and gluc; gender and the cardio label contribute negligibly.

\begin{figure*}[t]
\centering
\includegraphics[width=\textwidth]{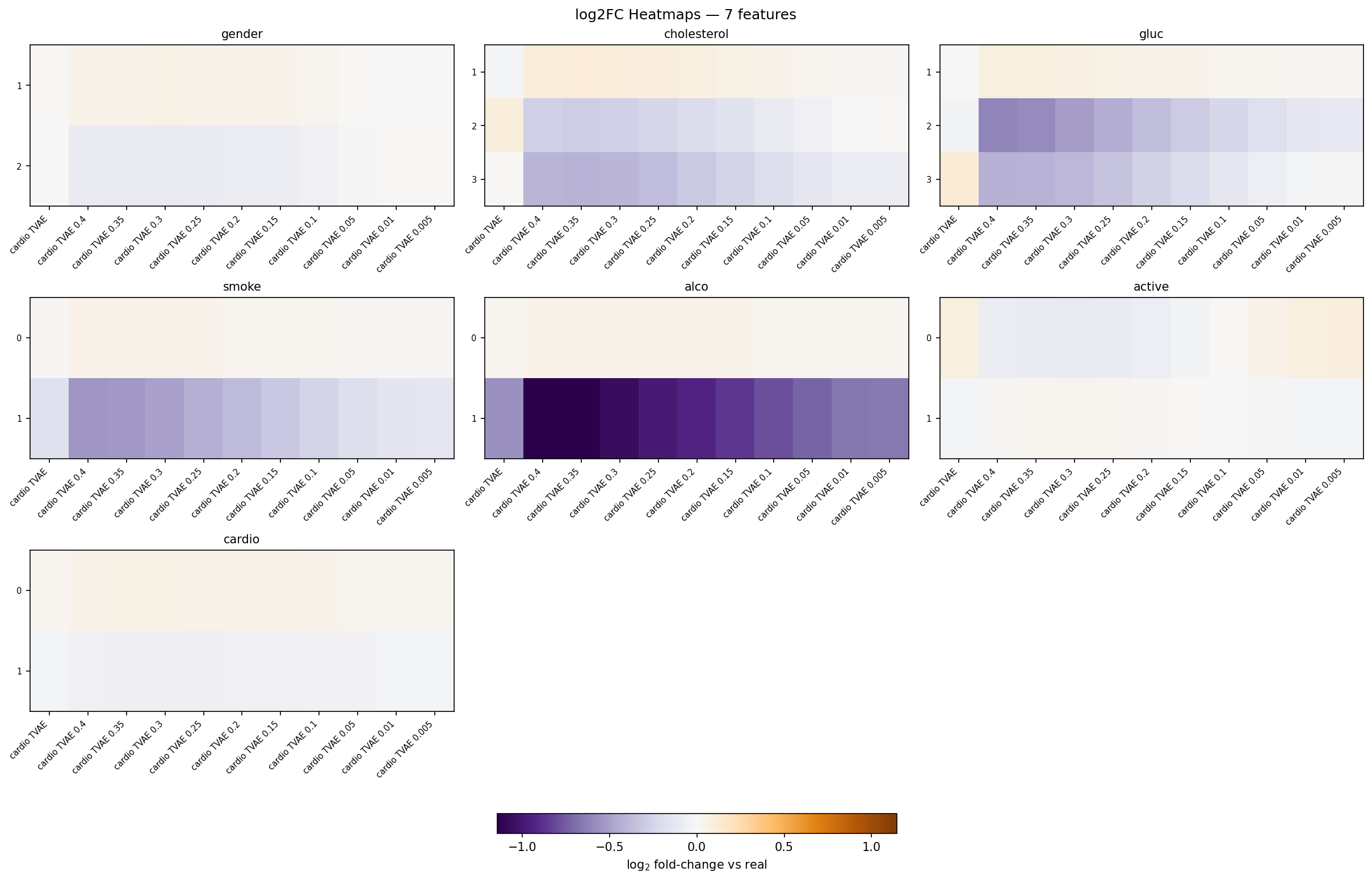}
\caption{Cardio TVAE categorical fidelity across acceptance targets.
Each panel shows the log\(_2\) fold‐change \(\mathrm{log2FC}=\log_2(p_{\mathrm{synth}}/p_{\mathrm{real}})\) for all levels of a categorical feature, comparing the baseline TVAE sample and HEOM–kNN post‐filtered variants at \(\tau_{\mathrm{ANY}}\in\{0.40,0.35,\ldots,0.005\}\) (columns). Rows list category levels in code order. The colorbar encodes direction and magnitude (white \(\approx0\); purple \(<0\) under-representation; orange \(>0\) over-representation; extremes clipped for readability).
Cardio TVAE begins with very low divergence (Table~\ref{tab:cardio-tvae-cat-total-js}); both loosening and tightening generally increase drift. Residual discrepancies are concentrated in alco, with smaller effects in smoke, active, cholesterol, and gluc; gender and the cardio label are essentially stable across targets.}
\label{fig:cardio-tvae-log2fc-heatmaps}
\end{figure*}

\begin{table}[t]
\centering
\small
\setlength{\tabcolsep}{3.5pt}
\begin{threeparttable}
\caption{Cardio TVAE: total categorical divergence (sum of per‐feature JS, \emph{log base 2; bits}) vs.\ acceptance target. Lower is better. The $\Delta$ columns show change relative to the baseline TVAE sample; negative values indicate improvement.}
\label{tab:cardio-tvae-cat-total-js}
\begin{tabularx}{\columnwidth}{
  >{\raggedright\arraybackslash}X
  S[table-format=1.6]
  S[table-format=+1.6]
  S[table-format=+3.1]}
\toprule
Variant & {Total JS} & {$\Delta$ vs baseline} & {\% $\Delta$} \\
\midrule
cardio TVAE        & \bfseries 0.001889 & \multicolumn{1}{c}{—}       & \multicolumn{1}{c}{—} \\
cardio TVAE 0.40   & 0.013676 & +0.011787 & +623.8 \\
cardio TVAE 0.35   & 0.013413 & +0.011524 & +609.9 \\
cardio TVAE 0.30   & 0.011970 & +0.010081 & +533.5 \\
cardio TVAE 0.25   & 0.009884 & +0.007995 & +423.2 \\
cardio TVAE 0.20   & 0.007848 & +0.005958 & +315.4 \\
cardio TVAE 0.15   & 0.005963 & +0.004074 & +215.6 \\
cardio TVAE 0.10   & 0.004173 & +0.002283 & +120.8 \\
cardio TVAE 0.05   & 0.002947 & +0.001058 & +56.0 \\
cardio TVAE 0.01   & 0.002261 & +0.000372 & +19.7 \\
cardio TVAE 0.005  & 0.002199 & +0.000310 & +16.4 \\
\bottomrule
\end{tabularx}
\begin{tablenotes}[flushleft]\footnotesize
\item \textit{Note.} “Total JS’’ sums per‐level Jensen–Shannon contributions within each categorical feature and then across features. All values are reported in \emph{bits} (log base~2).
\end{tablenotes}
\end{threeparttable}
\end{table}

\subsubsection{Numeric marginals (quantile heuristics and $d$)}
Aggregate heuristics (Table~\ref{tab:cardio-tvae-numeric-heuristic-summary}) reveal a center–tail tradeoff that differs from CTGAN:
\begin{itemize}
  \item The center alignment is best when tight: $\mathrm{median}(|\Delta_{\mathrm{med}}|)$ is 0.09\% at 0.01 (baseline: 0.22\%).
  \item The lower-tail relative shift is smallest at 0.2 (median $|\Delta_{\mathrm{low}}| = 1.69\%$), and grows as we tighten further.
  \item The upper-tail relative shift is smallest at 0.25 (median $|\Delta_{\mathrm{high}}| = 0.39\%$); elsewhere it is around 0.56–0.63\%.
  \item Tail masses are most stable near 0.10–0.15: mean $|\Delta p_{\mathrm{low}}|$ 1.78~pp at 0.15 (1.78~pp at 0.10) vs.\ 2.04~pp baseline, and mean $|\Delta p_{\mathrm{high}}|$ 1.70~pp at 0.15 (1.70–1.71~pp at 0.10–0.25) vs.\ 1.80~pp baseline.
\end{itemize}

\begin{table}[t]
\centering
\small
\setlength{\tabcolsep}{2.5pt}
\caption{Cardio TVAE: numeric marginal fidelity at \(q_L{=}0.05, q_H{=}0.95\). Reported are median absolute relative shifts (\%) and mean absolute tail‐mass deltas (\%). Lower is better; bold indicates the best per column.}
\label{tab:cardio-tvae-numeric-heuristic-summary}
\resizebox{\columnwidth}{!}{%
\begin{tabular}{lccccc}
\toprule
Dataset & Med.~$|\Delta_{\mathrm{med}}|$ & Med.~$|\Delta_{\mathrm{low}}|$ & Med.~$|\Delta_{\mathrm{high}}|$ & Mean~$|\Delta p_{\mathrm{low}}|$ & Mean~$|\Delta p_{\mathrm{high}}|$\\
\midrule
Baseline & 0.2\% & 2.1\% & 0.6\% & 2.0\% & 1.8\% \\
$\tau_{\mathrm{ANY}}{=}0.40$ & 0.4\% & 1.8\% & 0.6\% & 1.9\% & 2.0\% \\
$\tau_{\mathrm{ANY}}{=}0.35$ & 0.4\% & 1.8\% & 0.6\% & 1.8\% & 2.0\% \\
$\tau_{\mathrm{ANY}}{=}0.30$ & 0.5\% & 1.8\% & 0.6\% & 1.8\% & 2.0\% \\
$\tau_{\mathrm{ANY}}{=}0.25$ & 0.5\% & 1.8\% & \textbf{0.4\%} & 1.8\% & 1.9\% \\
$\tau_{\mathrm{ANY}}{=}0.20$ & 0.5\% & \textbf{1.7\%} & 0.6\% & 1.8\% & \textbf{1.7\%} \\
$\tau_{\mathrm{ANY}}{=}0.15$ & 0.5\% & 1.8\% & 0.6\% & \textbf{1.8\%} & \textbf{1.7\%} \\
$\tau_{\mathrm{ANY}}{=}0.10$ & 0.5\% & 2.1\% & 0.6\% & \textbf{1.8\%} & \textbf{1.7\%} \\
$\tau_{\mathrm{ANY}}{=}0.05$ & 0.2\% & 2.3\% & 0.6\% & \textbf{1.8\%} & 1.8\% \\
$\tau_{\mathrm{ANY}}{=}0.01$ & \textbf{0.1\%} & 2.5\% & 0.6\% & \textbf{1.8\%} & 1.8\% \\
\bottomrule
\end{tabular}}
\begin{tablenotes}[flushleft]\footnotesize
\item \textit{Note.} Medians/means are taken across numeric features. All values are reported in percent (\%). Cardio TVAE shows best medians at very tight targets (0.01), lowest lower‐tail shift at 0.20, lowest upper‐tail shift at 0.25, and the smallest tail‐mass deltas around 0.10–0.15.
\end{tablenotes}
\end{table}

Per-feature “winners’’ (Table~\ref{tab:cardio-tvae-numeric-heuristic-best-per-feature}) reflect the same balance: several features prefer 0.10–0.25 for tail stability, while the very tight setting (0.01) best matches medians.

\begin{table}[t]
\centering
\scriptsize
\setlength{\tabcolsep}{3pt}
\resizebox{\columnwidth}{!}{%
\begin{threeparttable}
\caption{Cardio TVAE: per‐feature best dataset by absolute center shift \((|\Delta_{\mathrm{med}}|)\) and upper‐tail stretch \((|\Delta_{\mathrm{high}}|)\). Reported are the winning dataset (Baseline or \(\tau_{\mathrm{ANY}}=\cdot\)) and its value.}
\label{tab:cardio-tvae-numeric-heuristic-best-per-feature}
\begin{tabular}{l l S[table-format=+2.1] l S[table-format=+2.1]}
\toprule
& \multicolumn{2}{c}{Best center (median)} & \multicolumn{2}{c}{Best upper tail (95th)} \\
\cmidrule(lr){2-3}\cmidrule(lr){4-5}
Feature & Winning dataset & {$\Delta_{\mathrm{med}}$ (\%)} & Winning dataset & {$\Delta_{\mathrm{high}}$ (\%)} \\
\midrule
age     & \(\tau_{\mathrm{ANY}}{=}0.35\) & 0.4  & \(\tau_{\mathrm{ANY}}{=}0.005\) & -0.1 \\
ap\_hi  & Baseline & 0.0  & \(\tau_{\mathrm{ANY}}{=}0.35\) & -0.6 \\
ap\_lo  & \(\tau_{\mathrm{ANY}}{=}0.005\) & 0.0  & \(\tau_{\mathrm{ANY}}{=}0.05\) & -3.0 \\
height  & \(\tau_{\mathrm{ANY}}{=}0.005\) & 0.0  & \(\tau_{\mathrm{ANY}}{=}0.10\) & 0.0 \\
weight  & \(\tau_{\mathrm{ANY}}{=}0.01\) & 0.1  & \(\tau_{\mathrm{ANY}}{=}0.35\) & -0.1 \\
\bottomrule
\end{tabular}
\begin{tablenotes}[flushleft]\footnotesize
\item \textit{Note.} Values are absolute relative shifts at the median (center) and 95th percentile (upper tail); negative signs indicate synthetic quantiles below the baseline reference.
\end{tablenotes}
\end{threeparttable}}
\end{table}

\paragraph{Cohen’s $d$ (numeric effect sizes).}
Effect sizes are small overall but improve with moderate tightening:
\emph{age} and \emph{height} are best at 0.15 ($d = 0.121$ and $0.0025$);
\emph{weight} is best at 0.10 ($d = 0.00298$);
\emph{ap\_hi} and \emph{ap\_lo} are best at 0.005 ($d = 0.0144$ and $0.0315$).
Counting minima across features yields: 0.15 (2 wins), 0.10 (1), 0.005 (2).

\paragraph{Interpretation.}
For Cardio TVAE, the generator already places mass close to dense regions of the real data. The HEOM–kNN rejection rule—by excluding points proximate to those dense regions—tends to \emph{exaggerate} categorical deficits (notably \textit{alco}, \textit{smoke}, \textit{gluc}/\textit{cholesterol}) and only modestly benefits numeric tails. Tighter thresholds improve numeric centers (and some tails) but raise categorical JS; loose thresholds raise JS substantially without numeric benefits.

\paragraph{Practical choice of target.}
If categorical fidelity is paramount, baseline (no acceptance) is optimal (total JS 0.00131). 
If a mild numeric gain is desired without large categorical cost, 0.10–0.15 reduces both tail-mass deltas and several $d$’s while keeping total JS small in absolute terms (e.g., 0.00289 at 0.10). 
Aggressive tightening (\(\tau \le 0.01\)) further improves medians but increases categorical drift and lower-tail shifts; loosening (\(\tau \ge 0.30\)) is not recommended here (JS grows to \(0.0083\)–\(0.0095\)).

\section{Multivariate Evaluation Extended Results}

\subsection{Protocol}
We quantify how well the synthetic data reproduce multivariate structure by comparing
pairwise association matrices computed on real and synthetic data for three association types:
(i) numerical--numerical (Pearson correlation, $\rho$), (ii) categorical--categorical
(Cram\'er's $V$), and (iii) categorical--numerical (correlation ratio, $\eta$).

\paragraph{How the synthetic datasets are produced.}
All $\tau$–variants are generated with the \textsc{HEOM--kNN} $\varepsilon_{\text{ANY}}$
\emph{rejection-with-replacement} sampler (Alg.~\ref{alg:heom-any}). In the HEOM-encoded space,
each real record $x_i$ defines a ``privacy ball'' of radius $r_i$ given by its 2-NN distance.
For any synthetic point $x$ we compute the signed margin
\[
  M(x) \;=\; \min_i \bigl(\,\lVert x-x_i\rVert_2^2 - r_i^2\,\bigr).
\]
A point is a \emph{violation} if $M(x)<0$ (it lies inside at least one privacy ball). For a
synthetic dataset $S$, the empirical violation rate is
\[
  \varepsilon_{\text{ANY}}(S)\;=\;\frac{1}{|S|}\sum_{x\in S}\mathbf{1}[M(x)<0].
\]
The sampler iteratively replaces the worst violating points until
$\varepsilon_{\text{ANY}}(S)$ falls below a user-specified \textbf{sampling threshold} $\tau$
(this is the parameter \texttt{min\_eps} in Alg.~\ref{alg:heom-any}). Smaller $\tau$ yields a
\emph{stricter} filter (fewer violations allowed, more aggressive replacement), while larger
$\tau$ is looser. We sweep
$\tau\in\{0.40,\,0.35,\,0.30,\,0.25,\,0.20,\,0.15,\,0.10,\,0.05,\,0.01,\,0.005\}$ and include a
\emph{baseline} obtained directly from the generator $G$ without rejection.

\paragraph{What we report for each $\tau$.}
For every dataset produced at threshold $\tau$ (and for the baseline), we compute:
\begin{itemize}
  \item \textbf{Frobenius distance} between the real and synthetic association matrices,
  \[
    d_F \;=\; \lVert A_{\text{real}} - A_{\text{synth}} \rVert_F,
    \quad\text{(smaller is better)}.
  \]
  \item \textbf{Spearman rank correlation} between the flattened off-diagonal entries,
  \[
    r_s \;=\; \mathrm{corr}_S\!\big(\mathrm{vec}_\triangle(A_{\text{real}}),\,
    \mathrm{vec}_\triangle(A_{\text{synth}})\big),
    \quad\text{(larger is better)}.
  \]
\end{itemize}


\subsection{Multivariate evaluation (Credit Card --- CTGAN)}
\label{app:credit-ctgan}

\subsubsection{Quantitative results}
\noindent\textit{Setup.} All variants are produced by the \textsc{HEOM--kNN}
$\varepsilon_{\mathrm{ANY}}$ \emph{rejection-with-replacement} sampler in
Alg.~\ref{alg:heom-any}. We vary the \textbf{sampling threshold} $\boldsymbol{\tau}$ (the
parameter \texttt{min\_eps}), which upper-bounds the empirical violation rate
$\varepsilon_{\mathrm{ANY}}(S)$; smaller $\tau$ is stricter (fewer violations allowed). The
\emph{baseline} is the raw CTGAN output (no rejection).

\begin{table}[t]
  \centering
  \caption{Frobenius distance between real and synthetic association matrices (Credit Card --- CTGAN
  with HEOM--kNN sampling). Lower is better.}
  \label{tab:credit-ctgan-frobenius}
  \small
  \resizebox{\columnwidth}{!}{%
  \begin{tabular}{
    l
    S[table-format=1.2]
    S[table-format=1.2]
    S[table-format=1.2]
    S[table-format=1.2]
    S[table-format=1.2]
    S[table-format=1.2]
    S[table-format=1.2]
    S[table-format=1.2]
    S[table-format=1.2]
    S[table-format=1.2]
    S[table-format=1.2]
  }
    \toprule
& \multicolumn{1}{c}{baseline}
& \multicolumn{1}{c}{$\tau=0.40$}
& \multicolumn{1}{c}{0.35}
& \multicolumn{1}{c}{0.30}
& \multicolumn{1}{c}{0.25}
& \multicolumn{1}{c}{0.20}
& \multicolumn{1}{c}{0.15}
& \multicolumn{1}{c}{0.10}
& \multicolumn{1}{c}{0.05}
& \multicolumn{1}{c}{0.01}
& \multicolumn{1}{c}{0.005} \\
\midrule
    Pearson $\rho$            & 0.9750 & 0.9630 & \bfseries 0.9524 & 0.9617 & 0.9663 & 0.9600 & 0.9758 & 0.9911 & 0.9811 & 0.9898 & \underline{\num{1.0366}} \\
    Cram\'er's $V$            & \underline{\num{0.7986}} & 0.7463 & 0.7478 & 0.7289 & 0.7271 & 0.7335 & 0.7595 & \bfseries 0.7125 & 0.7165 & 0.7146 & 0.7273 \\
    Correlation ratio $\eta$  & 0.6158 & 0.5806 & 0.5455 & 0.5701 & 0.5447 & \bfseries 0.5284 & 0.5520 & 0.5693 & 0.6010 & 0.6260 & \underline{\num{0.6423}} \\
    \bottomrule
  \end{tabular}}
\end{table}

\begin{table}[t]
  \centering
  \caption{Spearman correlation between real and synthetic association matrices (Credit Card --- CTGAN
  with HEOM--kNN sampling). Higher is better.}
  \label{tab:credit-ctgan-spearman}
  \small
  \resizebox{\columnwidth}{!}{%
  \begin{tabular}{
    l
    S[table-format=1.2]
    S[table-format=1.2]
    S[table-format=1.2]
    S[table-format=1.2]
    S[table-format=1.2]
    S[table-format=1.2]
    S[table-format=1.2]
    S[table-format=1.2]
    S[table-format=1.2]
    S[table-format=1.2]
    S[table-format=1.2]
  }
    \toprule
& \multicolumn{1}{c}{baseline}
& \multicolumn{1}{c}{$\tau=0.40$}
& \multicolumn{1}{c}{0.35}
& \multicolumn{1}{c}{0.30}
& \multicolumn{1}{c}{0.25}
& \multicolumn{1}{c}{0.20}
& \multicolumn{1}{c}{0.15}
& \multicolumn{1}{c}{0.10}
& \multicolumn{1}{c}{0.05}
& \multicolumn{1}{c}{0.01}
& \multicolumn{1}{c}{0.005} \\
\midrule
    Pearson $\rho$            & 0.793741 & 0.795461 & 0.787641 & 0.781940 & 0.785444 & 0.794633 & 0.803074 & 0.786781 & 0.803822 & \bfseries 0.807867 & 0.790269 \\
    Cram\'er's $V$            & 0.940711 & \bfseries 0.945455 & 0.942951 & 0.936891 & 0.939789 & 0.938867 & 0.935046 & 0.935178 & 0.932806 & \underline{\num{0.924506}} & 0.930698 \\
    Correlation ratio $\eta$  & 0.936962 & 0.936398 & 0.937303 & \underline{\num{0.933542}} & 0.938864 & 0.937329 & 0.940911 & \bfseries 0.942197  & 0.938698 & 0.936153 & 0.935589 \\
    \bottomrule
  \end{tabular}}
\end{table}

\paragraph{Key observations.}
\begin{itemize}
  \item \textbf{Pearson ($\rho$).} Frobenius distances range from \textbf{0.9524} at
        $\tau{=}0.35$ to \underline{\num{1.0366}} at $\tau{=}0.005$ (span $0.0842$).
  \item \textbf{Cram\'er's $V$.} Frobenius distances range from \textbf{0.7125} at $\tau{=}0.10$
        to \underline{\num{0.7986}} at baseline (span $0.0861$).
  \item \textbf{Correlation ratio ($\eta$).} Frobenius distances range from \textbf{0.5284} at
        $\tau{=}0.20$ to \underline{\num{0.6423}} at $\tau{=}0.005$ (span $0.1139$).
  \item \textbf{Spearman stability.} The rank ordering of pairwise dependencies is highly stable:
        $\rho$: $0.782$--$0.808$ (best at $\tau{=}0.01$);
        $V$: $0.924$--$0.945$ (best at $\tau{=}0.40$);
        $\eta$: $0.934$--$0.942$ (best at $\tau{=}0.10$).
\end{itemize}
Overall, CTGAN preserves the \emph{rank ordering} of dependencies very well---especially for
categorical and mixed-type relations---while the magnitude of the strongest continuous
correlations is slightly compressed at the smallest $\tau$. In these runs,
$\tau\!\in[0.20,\,0.35]$ offers the best trade-off.

\subsection{Multivariate evaluation (Credit Card --- TVAE)}
\label{app:credit-tvae}

\subsubsection{Quantitative results}
\noindent\textit{Setup.} All variants are produced by the \textsc{HEOM--kNN}
$\varepsilon_{\mathrm{ANY}}$ \emph{rejection-with-replacement} sampler in
Alg.~\ref{alg:heom-any}. We sweep the \textbf{sampling threshold} $\boldsymbol{\tau}$ (the
algorithm’s \texttt{min\_eps}), which upper-bounds the empirical violation rate
$\varepsilon_{\mathrm{ANY}}(S)$; smaller $\tau$ corresponds to a \emph{stricter} setting (fewer
violations). The \emph{baseline} is the raw TVAE output (no rejection).

\begin{table}[t]
  \centering
  \caption{Frobenius distance between real and synthetic association matrices (Credit Card --- TVAE
  with HEOM--kNN sampling). Lower is better.}
  \label{tab:credit-tvae-frobenius}
  \small
  \resizebox{\columnwidth}{!}{%
  \begin{tabular}{
    l
    S[table-format=1.2]
    S[table-format=1.2]
    S[table-format=1.2]
    S[table-format=1.2]
    S[table-format=1.2]
    S[table-format=1.2]
    S[table-format=1.2]
    S[table-format=1.2]
    S[table-format=1.2]
    S[table-format=1.2]
    S[table-format=1.2]
  }
    \toprule
    \toprule
& \multicolumn{1}{c}{baseline}
& \multicolumn{1}{c}{$\tau=0.40$}
& \multicolumn{1}{c}{0.35}
& \multicolumn{1}{c}{0.30}
& \multicolumn{1}{c}{0.25}
& \multicolumn{1}{c}{0.20}
& \multicolumn{1}{c}{0.15}
& \multicolumn{1}{c}{0.10}
& \multicolumn{1}{c}{0.05}
& \multicolumn{1}{c}{0.01}
& \multicolumn{1}{c}{0.005} \\
\midrule
    Pearson $\rho$            & \bfseries 0.8974 & 0.9126 & 0.9109 & 0.9195 & 0.9325 & 0.9547 & 0.9764 & 0.9962 & 1.0135 & 1.0289 & \underline{\num{1.0329}} \\
    Cram\'er's $V$            & 1.0010 & 0.9901 & \bfseries 0.9851 & 0.9935 & 1.0123 & 1.0247 & 1.0425 & 1.0500 & 1.0656 & 1.0767 & \underline{\num{1.0790}} \\
    Correlation ratio $\eta$  & \bfseries 0.7492 & 0.7762 & 0.7875 & 0.8078 & 0.8430 & 0.8830 & 0.9175 & 0.9599 & 1.0076 & 1.0391 & \underline{\num{1.0434}} \\
    \bottomrule
  \end{tabular}}
\end{table}

\begin{table}[t]
  \centering
  \caption{Spearman correlation between real and synthetic association matrices (Credit Card --- TVAE
  with HEOM--kNN sampling). Higher is better.}
  \label{tab:credit-tvae-spearman}
  \small
  \resizebox{\columnwidth}{!}{%
  \begin{tabular}{
    l
    S[table-format=1.2]
    S[table-format=1.2]
    S[table-format=1.2]
    S[table-format=1.2]
    S[table-format=1.2]
    S[table-format=1.2]
    S[table-format=1.2]
    S[table-format=1.2]
    S[table-format=1.2]
    S[table-format=1.2]
    S[table-format=1.2]
  }
    \toprule
& \multicolumn{1}{c}{baseline}
& \multicolumn{1}{c}{$\tau=0.40$}
& \multicolumn{1}{c}{0.35}
& \multicolumn{1}{c}{0.30}
& \multicolumn{1}{c}{0.25}
& \multicolumn{1}{c}{0.20}
& \multicolumn{1}{c}{0.15}
& \multicolumn{1}{c}{0.10}
& \multicolumn{1}{c}{0.05}
& \multicolumn{1}{c}{0.01}
& \multicolumn{1}{c}{0.005} \\
\midrule
    Pearson $\rho$            & \underline{\num{0.843319}} & 0.847157 & 0.851664 & 0.855439 & 0.861761 & 0.864548 & 0.866046 & 0.869581 & 0.874789 & \bfseries 0.875426 & 0.874757 \\
    Cram\'er's $V$            & \bfseries 0.889196 & 0.880632 & 0.884321 & 0.883267 & 0.875758 & 0.871805 & 0.865086 & 0.864032 & 0.854414 & 0.850329 & \underline{\num{0.846904}} \\
    Correlation ratio $\eta$  & 0.894260 & \bfseries 0.895187 & 0.893766 & 0.893984 & 0.890661 & 0.886585 & 0.886187 & 0.882425 & 0.876985 & 0.873569 & \underline{\num{0.873145}} \\
    \bottomrule
  \end{tabular}}
\end{table}

\paragraph{Key observations.}
\begin{itemize}
  \item \textbf{Pearson ($\rho$).} Frobenius distances increase as the sampler is tightened
        (smaller $\tau$), from \textbf{0.8974} at baseline to \underline{\num{1.0329}} at
        $\tau{=}0.005$ (span $\approx 0.1355$). The rank ordering of coefficients remains very
        stable, with Spearman correlations slightly increasing and peaking at \textbf{0.8754}
        at $\tau{=}0.01$ (range $\approx 0.0321$).
  \item \textbf{Cram\'er's $V$.} Frobenius distances follow a similar pattern (best
        \textbf{0.9851} at $\tau{=}0.35$, worst \underline{\num{1.0790}} at $\tau{=}0.005$; span
        $\approx 0.0939$), while Spearman correlations decline gradually from \textbf{0.8892}
        (baseline) to \underline{\num{0.8469}} at $\tau{=}0.005$.
  \item \textbf{Correlation ratio ($\eta$).} Frobenius distances rise nearly monotonically with
        stricter $\tau$ (\textbf{0.7492} $\rightarrow$ \underline{\num{1.0434}}; span
        $\approx 0.2942$), and the Spearman correlation degrades mildly (from
        \textbf{0.8952} at $\tau{=}0.40$ to \underline{\num{0.8731}} at $\tau{=}0.005$).
\end{itemize}
Overall, TVAE preserves the \emph{rank ordering} of dependencies well (all Spearman values
$\gtrsim 0.85$), while the magnitude of associations compresses as $\tau$ is tightened. In these
runs, moderate thresholds ($\tau\!\approx\!0.25$--$0.40$) provide a good utility trade-off.

\subsection{Multivariate evaluation (Adult --- CTGAN)}
\label{app:adult-ctgan}

\subsubsection{Quantitative results}
\noindent\textit{Setup.} All variants are produced by the \textsc{HEOM--kNN}
$\varepsilon_{\mathrm{ANY}}$ \emph{rejection-with-replacement} sampler in
Alg.~\ref{alg:heom-any}. We sweep the \textbf{sampling threshold} $\boldsymbol{\tau}$ (the
algorithm’s \texttt{min\_eps}), which upper-bounds the empirical violation rate
$\varepsilon_{\mathrm{ANY}}(S)$; smaller $\tau$ is a \emph{stricter} setting (fewer violations).
The \emph{baseline} is the raw CTGAN output without rejection.

\begin{table}[t]
  \centering
  \caption{Frobenius distance between real and synthetic association matrices (Adult --- CTGAN
  with HEOM--kNN sampling). Lower is better.}
  \label{tab:adult-ctgan-frobenius}
  \small
  \resizebox{\columnwidth}{!}{%
  \begin{tabular}{
    l
    S[table-format=1.2]
    S[table-format=1.2]
    S[table-format=1.2]
    S[table-format=1.2]
    S[table-format=1.2]
    S[table-format=1.2]
    S[table-format=1.2]
    S[table-format=1.2]
    S[table-format=1.2]
    S[table-format=1.2]
    S[table-format=1.2]
  }
    \toprule
& \multicolumn{1}{c}{baseline}
& \multicolumn{1}{c}{$\tau=0.40$}
& \multicolumn{1}{c}{0.35}
& \multicolumn{1}{c}{0.30}
& \multicolumn{1}{c}{0.25}
& \multicolumn{1}{c}{0.20}
& \multicolumn{1}{c}{0.15}
& \multicolumn{1}{c}{0.10}
& \multicolumn{1}{c}{0.05}
& \multicolumn{1}{c}{0.01}
& \multicolumn{1}{c}{0.005} \\
\midrule
    Pearson $\rho$            & 0.2148 & \underline{\num{0.2466}} & 0.2159 & 0.2144 & 0.2303 & 0.2190 & 0.2238 & \bfseries 0.2066 & 0.2117 & 0.2197 & 0.2181 \\
    Cram\'er's $V$            & 0.2967 & \bfseries 0.2924 & 0.2926 & 0.3013 & 0.3120 & 0.3196 & 0.3385 & 0.3590 & 0.3678 & 0.3934 & \underline{\num{0.3988}} \\
    Correlation ratio $\eta$  & 0.3561 & 0.3814 & 0.3584 & \bfseries 0.3524 & 0.3608 & 0.3716 & 0.3804 & 0.3865 & 0.3897 & \underline{\num{0.4181}} & 0.4048 \\
    \bottomrule
  \end{tabular}}
\end{table}

\begin{table}[t]
  \centering
  \caption{Spearman correlation between real and synthetic association matrices (Adult --- CTGAN
  with HEOM--kNN sampling). Higher is better.}
  \label{tab:adult-ctgan-spearman}
  \small
  \resizebox{\columnwidth}{!}{%
  \begin{tabular}{
    l
    S[table-format=1.2]
    S[table-format=1.2]
    S[table-format=1.2]
    S[table-format=1.2]
    S[table-format=1.2]
    S[table-format=1.2]
    S[table-format=1.2]
    S[table-format=1.2]
    S[table-format=1.2]
    S[table-format=1.2]
    S[table-format=1.2]
  }
    \toprule
& \multicolumn{1}{c}{baseline}
& \multicolumn{1}{c}{$\tau=0.40$}
& \multicolumn{1}{c}{0.35}
& \multicolumn{1}{c}{0.30}
& \multicolumn{1}{c}{0.25}
& \multicolumn{1}{c}{0.20}
& \multicolumn{1}{c}{0.15}
& \multicolumn{1}{c}{0.10}
& \multicolumn{1}{c}{0.05}
& \multicolumn{1}{c}{0.01}
& \multicolumn{1}{c}{0.005} \\
\midrule
    Pearson $\rho$            & \bfseries 0.810714 & 0.782143 & 0.771429 & 0.785714 & 0.767857 & 0.757143 & \underline{\num{0.742857}} & 0.760714 & 0.771429 & 0.778571 & 0.785714 \\
    Cram\'er's $V$            & 0.913256 & 0.923037 & \bfseries 0.927928 & 0.927413 & 0.925354 & 0.919434 & 0.893179 & 0.887773 & 0.876963 & 0.872587 & \underline{\num{0.866667}} \\
    Correlation ratio $\eta$  & 0.910806 & 0.902649 & 0.911340 & \bfseries 0.923461 & 0.907147 & 0.919726 & 0.916143 & 0.908748 & 0.907681 & \underline{\num{0.892510}} & 0.903945 \\
    \bottomrule
  \end{tabular}}
\end{table}

\paragraph{Key observations.}
\begin{itemize}
  \item \textbf{Pearson ($\rho$).} Frobenius distances are small across thresholds (minimum
        \textbf{0.2066} at $\tau{=}0.10$, maximum \underline{\num{0.2466}} at $\tau{=}0.40$; span
        $\approx 0.040$). Spearman correlations remain high, from a best of \textbf{0.8107} at
        baseline to a worst of \underline{\num{0.7429}} at $\tau{=}0.15$, indicating stable
        \emph{ordering} of numeric dependencies.
  \item \textbf{Cram\'er's $V$.} Frobenius distances increase as the sampler is tightened
        (smaller $\tau$), from \textbf{0.2924} at $\tau{=}0.40$ to \underline{\num{0.3988}} at
        $\tau{=}0.005$ (span $\approx 0.106$). Rank concordance declines gradually but remains
        high (Spearman from \textbf{0.9279} to \underline{\num{0.8667}}).
  \item \textbf{Correlation ratio ($\eta$).} Frobenius distance is lowest at $\tau{=}0.30$
        (\textbf{0.3524}) and largest at $\tau{=}0.01$ (\underline{\num{0.4181}}; span
        $\approx 0.066$). Spearman correlations peak at \textbf{0.9235} ($\tau{=}0.30$) and
        bottom at \underline{\num{0.8925}} ($\tau{=}0.01$).
\end{itemize}
Overall, Adult--CTGAN maintains the \emph{relative ranking} of associations well (all Spearman
values $\gtrsim 0.74$ for $\rho$, $\gtrsim 0.86$ for $V$, and $\gtrsim 0.89$ for $\eta$), with
moderate magnitude compression as $\tau$ decreases (stricter sampling).

\subsection{Multivariate evaluation (Adult --- TVAE)}
\label{app:adult-tvae}

\subsubsection{Quantitative results}
\noindent\textit{Setup.} All variants are produced by the \textsc{HEOM--kNN}
$\varepsilon_{\mathrm{ANY}}$ \emph{rejection-with-replacement} sampler in
Alg.~\ref{alg:heom-any}. We sweep the \textbf{sampling threshold} $\boldsymbol{\tau}$ (the
algorithm’s \texttt{min\_eps}), which upper-bounds the empirical violation rate
$\varepsilon_{\mathrm{ANY}}(S)$. Smaller $\tau$ is a \emph{stricter} setting (fewer violations
allowed). The \emph{baseline} is the raw TVAE output (no rejection).

\begin{table}[t]
  \centering
  \caption{Frobenius distance between real and synthetic association matrices (Adult --- TVAE
  with HEOM--kNN sampling). Lower is better.}
  \label{tab:adult-tvae-frobenius}
  \small
  \resizebox{\columnwidth}{!}{%
  \begin{tabular}{
    l
    S[table-format=1.2]
    S[table-format=1.2]
    S[table-format=1.2]
    S[table-format=1.2]
    S[table-format=1.2]
    S[table-format=1.2]
    S[table-format=1.2]
    S[table-format=1.2]
    S[table-format=1.2]
    S[table-format=1.2]
    S[table-format=1.2]
  }
    \toprule
& \multicolumn{1}{c}{baseline}
& \multicolumn{1}{c}{$\tau=0.40$}
& \multicolumn{1}{c}{0.35}
& \multicolumn{1}{c}{0.30}
& \multicolumn{1}{c}{0.25}
& \multicolumn{1}{c}{0.20}
& \multicolumn{1}{c}{0.15}
& \multicolumn{1}{c}{0.10}
& \multicolumn{1}{c}{0.05}
& \multicolumn{1}{c}{0.01}
& \multicolumn{1}{c}{0.005} \\
\midrule
    Pearson $\rho$            & \underline{\num{0.2005}} & \bfseries 0.1845 & 0.1893 & 0.1899 & 0.1936 & 0.1932 & 0.1922 & 0.1936 & 0.1954 & 0.1952 & 0.1956 \\
    Cram\'er's $V$            & 0.4572 & \bfseries 0.4468 & 0.4469 & 0.4523 & 0.4611 & 0.4665 & 0.4763 & 0.4885 & 0.5050 & 0.5185 & \underline{\num{0.5205}} \\
    Correlation ratio $\eta$  & 0.4690 & \bfseries 0.4518 & 0.4550 & 0.4576 & 0.4641 & 0.4690 & 0.4784 & 0.4895 & 0.5048 & 0.5181 & \underline{\num{0.5201}} \\
    \bottomrule
  \end{tabular}}
\end{table}

\begin{table}[t]
  \centering
  \caption{Spearman correlation between real and synthetic association matrices (Adult --- TVAE
  with HEOM--kNN sampling). Higher is better.}
  \label{tab:adult-tvae-spearman}
  \small
  \resizebox{\columnwidth}{!}{%
  \begin{tabular}{
    l
    S[table-format=1.2]
    S[table-format=1.2]
    S[table-format=1.2]
    S[table-format=1.2]
    S[table-format=1.2]
    S[table-format=1.2]
    S[table-format=1.2]
    S[table-format=1.2]
    S[table-format=1.2]
    S[table-format=1.2]
    S[table-format=1.2]
  }
    \toprule
    & \multicolumn{1}{c}{baseline}
    & \multicolumn{1}{c}{$\tau=0.40$}
    & \multicolumn{1}{c}{0.35}
    & \multicolumn{1}{c}{0.30}
    & \multicolumn{1}{c}{0.25}
    & \multicolumn{1}{c}{0.20}
    & \multicolumn{1}{c}{0.15}
    & \multicolumn{1}{c}{0.10}
    & \multicolumn{1}{c}{0.05}
    & \multicolumn{1}{c}{0.01}
    & \multicolumn{1}{c}{0.005} \\
    \midrule
    Pearson $\rho$            & \underline{\num{0.435714}} & 0.575000 & 0.575000 & 0.575000 & 0.578571 & 0.592857 & 0.621429 & \bfseries 0.635714 & 0.632143 & 0.607143 & 0.632143 \\
    Cram\'er's $V$            & 0.866152 & 0.893179 & \bfseries 0.896782 & 0.889060 & 0.892149 & 0.881081 & 0.875418 & 0.868211 & 0.865122 & 0.857915 & \underline{\num{0.857658}} \\
    Correlation ratio $\eta$  & \underline{\num{0.849133}} & 0.881609 & 0.881075 & 0.885649 & 0.886564 & 0.891748 & \bfseries 0.892281 & 0.883438 & 0.875586 & 0.861254 & 0.858967 \\
    \bottomrule
  \end{tabular}}
\end{table}

\paragraph{Key observations.}
\begin{itemize}
  \item \textbf{Pearson ($\rho$).} Frobenius distances are uniformly small; the best match occurs
        at $\tau{=}0.40$ (\textbf{0.1845}), and the largest at baseline (\underline{\num{0.2005}};
        span $\approx 0.016$). The \emph{ordering} of numeric dependencies improves as the sampler
        is tightened (smaller $\tau$), with Spearman correlations rising from \underline{\num{0.436}}
        (baseline) to a peak of \textbf{0.636} at $\tau{=}0.10$.
  \item \textbf{Cram\'er's $V$.} Frobenius distances increase monotonically as the sampler becomes
        stricter (from \textbf{0.4468} at $\tau{=}0.40$ to \underline{\num{0.5205}} at $\tau{=}0.005$;
        span $\approx 0.074$), while Spearman correlations remain high but decline gently
        (\textbf{0.897} $\rightarrow$ \underline{\num{0.858}}).
  \item \textbf{Correlation ratio ($\eta$).} Frobenius distances show a similar pattern (best
        \textbf{0.4518} at $\tau{=}0.40$; worst \underline{\num{0.5201}} at $\tau{=}0.005$; span
        $\approx 0.068$). Rank concordance stays strong throughout (peak \textbf{0.892} at
        $\tau{=}0.15$).
\end{itemize}
Overall, Adult--TVAE reproduces the \emph{relative ranking} of associations well for categorical
and mixed-type structure (all Spearman values $\gtrsim 0.86$), and improves the ordering of
numeric correlations as the sampler is tightened, albeit with mild magnitude compression at the
smallest $\tau$.

\subsection{Multivariate evaluation (Cardio --- CTGAN)}
\label{app:cardio-ctgan}

\subsubsection{Quantitative results}
\noindent\textit{Setup.} All variants are produced by the \textsc{HEOM--kNN}
$\varepsilon_{\mathrm{ANY}}$ \emph{rejection-with-replacement} sampler in
Alg.~\ref{alg:heom-any}. We sweep the \textbf{sampling threshold} $\boldsymbol{\tau}$ (the
algorithm’s \texttt{min\_eps}), which upper-bounds the empirical violation rate
$\varepsilon_{\mathrm{ANY}}(S)$. Smaller $\tau$ is a \emph{stricter} setting (fewer violations
allowed). The \emph{baseline} is the raw CTGAN output (no rejection).

\begin{table}[t]
  \centering
  \caption{Frobenius distance between real and synthetic association matrices (Cardio --- CTGAN
  with HEOM--kNN sampling). Lower is better.}
  \label{tab:cardio-ctgan-frobenius}
  \small
  \resizebox{\columnwidth}{!}{%
  \begin{tabular}{
    l
    S[table-format=1.2]
    S[table-format=1.2]
    S[table-format=1.2]
    S[table-format=1.2]
    S[table-format=1.2]
    S[table-format=1.2]
    S[table-format=1.2]
    S[table-format=1.2]
    S[table-format=1.2]
    S[table-format=1.2]
    S[table-format=1.2]
  }
    \toprule
    & \multicolumn{1}{c}{baseline}
    & \multicolumn{1}{c}{$\tau=0.40$}
    & \multicolumn{1}{c}{0.35}
    & \multicolumn{1}{c}{0.30}
    & \multicolumn{1}{c}{0.25}
    & \multicolumn{1}{c}{0.20}
    & \multicolumn{1}{c}{0.15}
    & \multicolumn{1}{c}{0.10}
    & \multicolumn{1}{c}{0.05}
    & \multicolumn{1}{c}{0.01}
    & \multicolumn{1}{c}{0.005} \\
    \midrule
    Pearson $\rho$            & \bfseries 0.5061 & 0.5282 & 0.5295 & 0.5250 & 0.5491 & 0.5461 & 0.5370 & 0.5436 & 0.5542 & \underline{\num{0.5550}} & 0.5497 \\
    Cram\'er's $V$            & 0.2621 & \underline{\num{0.2998}} & 0.2584 & 0.2370 & 0.2431 & 0.2234 & \bfseries 0.2123 & 0.2166 & 0.2163 & 0.2319 & 0.2393 \\
    Correlation ratio $\eta$  & \bfseries 0.3699 & 0.3967 & 0.3913 & 0.3909 & 0.3979 & \underline{\num{0.3989}} & 0.3983 & 0.3938 & 0.3930 & 0.3975 & 0.3953 \\
    \bottomrule
  \end{tabular}}
\end{table}

\begin{table}[t]
  \centering
  \caption{Spearman correlation between real and synthetic association matrices (Cardio --- CTGAN
  with HEOM--kNN sampling). Higher is better.}
  \label{tab:cardio-ctgan-spearman}
  \small
  \resizebox{\columnwidth}{!}{%
  \begin{tabular}{
    l
    S[table-format=1.2]
    S[table-format=1.2]
    S[table-format=1.2]
    S[table-format=1.2]
    S[table-format=1.2]
    S[table-format=1.2]
    S[table-format=1.2]
    S[table-format=1.2]
    S[table-format=1.2]
    S[table-format=1.2]
    S[table-format=1.2]
  }
    \toprule
    & \multicolumn{1}{c}{baseline}
    & \multicolumn{1}{c}{$\tau=0.40$}
    & \multicolumn{1}{c}{0.35}
    & \multicolumn{1}{c}{0.30}
    & \multicolumn{1}{c}{0.25}
    & \multicolumn{1}{c}{0.20}
    & \multicolumn{1}{c}{0.15}
    & \multicolumn{1}{c}{0.10}
    & \multicolumn{1}{c}{0.05}
    & \multicolumn{1}{c}{0.01}
    & \multicolumn{1}{c}{0.005} \\
    \midrule
    Pearson $\rho$            & \bfseries 0.418182 & 0.393939 & 0.296970 & 0.406061 & 0.224242 & 0.333333 & 0.406061 & 0.236364 & \underline{\num{0.212121}} & 0.212121 & 0.212121 \\
    Cram\'er's $V$            & \bfseries 0.810390 & \underline{\num{0.715584}} & 0.722078 & 0.724675 & 0.735065 & 0.774026 & 0.761039 & 0.728571 & 0.801299 & 0.738961 & 0.748052 \\
    Correlation ratio $\eta$  & 0.834734 & 0.848179 & \bfseries 0.853501 & 0.844818 & 0.836134 & 0.831092 & 0.848459 & 0.833894 & 0.845658 & \underline{\num{0.830812}} & 0.840056 \\
    \bottomrule
  \end{tabular}}
\end{table}

\paragraph{Key observations.}
\begin{itemize}
  \item \textbf{Pearson ($\rho$).} Frobenius distances vary moderately, from a minimum of
        \textbf{0.5061} at baseline to a maximum of \underline{\num{0.5550}} at $\tau{=}0.01$ (span
        $\approx 0.049$). Rank agreement for numeric correlations is comparatively low and
        degrades as the sampler is tightened (smaller $\tau$): Spearman falls from \textbf{0.418}
        to \underline{\num{0.212}}.
  \item \textbf{Cram\'er's $V$.} The best Frobenius distance is achieved around $\tau{=}0.15$
        (\textbf{0.2123}), and the worst at $\tau{=}0.40$ (\underline{\num{0.2998}}; span
        $\approx 0.088$). Despite this, the \emph{ordering} of categorical associations remains
        high and fairly stable (Spearman $0.716$--$0.810$).
  \item \textbf{Correlation ratio ($\eta$).} Frobenius distances are very similar across $\tau$
        settings (range $\approx 0.029$ around $0.37$--$0.40$), with consistently strong rank
        concordance (Spearman $0.831$--\textbf{0.854}).
\end{itemize}
Overall, Cardio--CTGAN preserves the relative ranking of categorical and mixed-type dependencies
well, while numeric correlation ordering is less robust as the sampler is tightened (smaller
$\tau$).

\subsection{Multivariate evaluation (Cardio --- TVAE)}
\label{app:cardio-tvae}

\subsubsection{Quantitative results}
\noindent\textit{Setup.} As above, all variants are produced by the \textsc{HEOM--kNN}
$\varepsilon_{\mathrm{ANY}}$ \emph{rejection-with-replacement} sampler in
Alg.~\ref{alg:heom-any}, sweeping the sampling threshold $\boldsymbol{\tau}$ over
$\{0.40,\ldots,0.005\}$. The \emph{baseline} is the raw TVAE output (no rejection).

\begin{table}[h]
  \centering
  \caption{Frobenius distance between real and synthetic association matrices (Cardio --- TVAE
  with HEOM--kNN sampling). Lower is better.}
  \label{tab:cardio-tvae-frobenius}
  \small
  \resizebox{\columnwidth}{!}{%
  \begin{tabular}{
    l
    S[table-format=1.2]
    S[table-format=1.2]
    S[table-format=1.2]
    S[table-format=1.2]
    S[table-format=1.2]
    S[table-format=1.2]
    S[table-format=1.2]
    S[table-format=1.2]
    S[table-format=1.2]
    S[table-format=1.2]
    S[table-format=1.2]
  }
    \toprule
    & \multicolumn{1}{c}{baseline}
    & \multicolumn{1}{c}{$\tau=0.40$}
    & \multicolumn{1}{c}{0.35}
    & \multicolumn{1}{c}{0.30}
    & \multicolumn{1}{c}{0.25}
    & \multicolumn{1}{c}{0.20}
    & \multicolumn{1}{c}{0.15}
    & \multicolumn{1}{c}{0.10}
    & \multicolumn{1}{c}{0.05}
    & \multicolumn{1}{c}{0.01}
    & \multicolumn{1}{c}{0.005} \\
    \midrule
    Pearson $\rho$            & \bfseries 0.2789 & 0.2890 & 0.2922 & 0.2928 & 0.2916 & \underline{\num{0.2939}} & 0.2914 & 0.2905 & 0.2891 & 0.2891 & 0.2877 \\
    Cram\'er’s $V$            & \underline{\num{0.2510}} & 0.2202 & \bfseries 0.2191 & 0.2276 & 0.2311 & 0.2372 & 0.2372 & 0.2387 & 0.2436 & 0.2471 & 0.2467 \\
    Correlation ratio $\eta$  & \bfseries 0.3473 & 0.3632 & 0.3611 & 0.3623 & 0.3633 & 0.3652 & 0.3679 & 0.3720 & 0.3771 & 0.3807 & \underline{\num{0.3814}} \\
    \bottomrule
  \end{tabular}}
\end{table}

\begin{table}[h]
  \centering
  \caption{Spearman correlation between real and synthetic association matrices (Cardio --- TVAE
  with HEOM--kNN sampling). Higher is better.}
  \label{tab:cardio-tvae-spearman}
  \small
  \resizebox{\columnwidth}{!}{%
  \begin{tabular}{
    l
    S[table-format=1.2]
    S[table-format=1.2]
    S[table-format=1.2]
    S[table-format=1.2]
    S[table-format=1.2]
    S[table-format=1.2]
    S[table-format=1.2]
    S[table-format=1.2]
    S[table-format=1.2]
    S[table-format=1.2]
    S[table-format=1.2]
  }
    \toprule
    & \multicolumn{1}{c}{baseline}
    & \multicolumn{1}{c}{$\tau=0.40$}
    & \multicolumn{1}{c}{0.35}
    & \multicolumn{1}{c}{0.30}
    & \multicolumn{1}{c}{0.25}
    & \multicolumn{1}{c}{0.20}
    & \multicolumn{1}{c}{0.15}
    & \multicolumn{1}{c}{0.10}
    & \multicolumn{1}{c}{0.05}
    & \multicolumn{1}{c}{0.01}
    & \multicolumn{1}{c}{0.005} \\
    \midrule
    Pearson $\rho$            & \bfseries 0.551515 & 0.539394 & 0.539394 & 0.539394 & 0.539394 & \underline{\num{0.515152}} & \underline{\num{0.515152}} & \underline{\num{0.515152}} & \underline{\num{0.515152}} & \underline{\num{0.515152}} & \underline{\num{0.515152}} \\
    Cram\'er’s $V$            & \bfseries 0.887013 & 0.835065 & 0.855844 & 0.832468 & 0.814286 & \underline{\num{0.764935}} & 0.772727 & 0.823377 & 0.835065 & 0.867532 & 0.867532 \\
    Correlation ratio $\eta$  & 0.728852 & \bfseries 0.741176 & 0.718487 & 0.694678 & 0.707283 & \underline{\num{0.689356}} & 0.693277 & 0.693838 & 0.702521 & 0.694398 & 0.698039 \\
    \bottomrule
  \end{tabular}}
\end{table}

\paragraph{Key observations.}
\begin{itemize}
  \item \textbf{Pearson ($\rho$).} Frobenius distances are tightly clustered, from a minimum of
        \textbf{0.2789} at baseline to a maximum of \underline{\num{0.2939}} at $\tau{=}0.20$ (span
        $\approx 0.015$). Rank agreement for numeric correlations is modest and decreases
        slightly as the sampler is tightened (Spearman from \textbf{0.552} to
        \underline{\num{0.515}}; span $\approx 0.036$).
  \item \textbf{Cram\'er’s $V$.} Frobenius distances improve over baseline under moderate
        thresholds (best \textbf{0.219} at $\tau{=}0.35$ vs.\ worst \underline{\num{0.251}} at
        baseline; span $\approx 0.032$), while rank concordance remains high
        (Spearman $0.765$--\textbf{0.887}).
  \item \textbf{Correlation ratio ($\eta$).} Frobenius distances gradually worsen as $\tau$ is
        tightened (from \textbf{0.3473} to \underline{\num{0.3814}}; span $\approx 0.034$), with
        Spearman correlations staying in a moderate band (\underline{\num{0.689}}--\textbf{0.741}).
\end{itemize}
Overall, Cardio--TVAE preserves the \emph{ordering} of categorical and mixed-type dependencies
well and keeps numeric distances stable across $\tau$, although numeric rank agreement is lower
than for categorical structure.

\section{Weighted HEOM--$k$NN $\varepsilon_{\mathrm{ANY}}$ Rejection-with-Replacement}\label{app:weighted-heom-any-knn}

\providecommand{\Enc}{\mathrm{ENCODE}}

\noindent\textbf{Overview.} This ablation extends the unweighted HEOM--$k$NN $\varepsilon_{\mathrm{ANY}}$ filter by reweighting coordinates according to column entropies and, optionally, by using general $k$-NN radii instead of fixed $2$-NN radii. It is a post-hoc, distance-based heuristic that reuses the \emph{ANY} risk definition and the same rejection-with-replacement protocol, while altering the geometry in which distances are measured and permitting different radius orders. The method \emph{does not} provide a formal $(\varepsilon,\delta)$-DP guarantee; it only controls an empirical proximity proxy. In practice it isolates dense regions in a mixed-type space and replaces candidates that encroach on any real record’s $k$-NN ball, reducing near-duplicates without retraining the generator.

\medskip
\noindent\textbf{Notation \& setup.} We use a mixed-type HEOM embedding: min--max scaling to $[0,1]$ for numeric columns $\mathcal{N}$, and one--hot encoding for categorical columns $\mathcal{C}$ with each dummy scaled by $1/\sqrt{2}$. Let $\Enc$ denote the fitted encoder, mapping $D \mapsto X\in\mathbb{R}^d$. For column-wise weights we set $w_c \propto 1/(H_c+\varepsilon)$, where $H_c$ is the empirical Shannon entropy (for numerics, from binned histograms), and expand them over the one--hot dummies to obtain $w_{\mathrm{dim}}\in\mathbb{R}^d$ (all dummies of a categorical share its $w_c$). Define $w_{\mathrm{scale}}=\sqrt{w_{\mathrm{dim}}}$ and the weighted Euclidean norm
\[
\lVert u - v \rVert_{w} := \bigl\lVert (u-v)\odot w_{\mathrm{scale}}\bigr\rVert_{2},
\]
so that $X_r^{w}=X_r\odot w_{\mathrm{scale}}$ and similarly for synthetic encodings $X_s^{w}$. Symbols follow Algorithm~\ref{alg:weighted-heom-any-knn}.

\medskip
\noindent\textbf{$k$-NN radii and ANY margin (weighted space).} For each real point $x_i$ (in the weighted space), let $r_i$ be the distance to its $k$-th nearest \emph{real} neighbor; set $R_{\max}=\max_i r_i$. For a candidate $x$, define the squared margin
\[
M(x) := \min_{i}\bigl(\,\lVert x-x_i\rVert_{w}^{2}-r_i^{2}\,\bigr).
\]
A candidate is unsafe iff $M(x)<0$. For a synthetic set of size $n$, the empirical ANY-risk is
\[
\varepsilon_{\mathrm{ANY}}=\frac{1}{n}\sum_{x\in X_s^{w}}\mathbf{1}\!\left[M(x)<0\right].
\]

\medskip
\noindent\textbf{Acceptance rule \& convergence intuition.} We reuse the rejection-with-replacement policy: at each iteration, propose $x'$ and accept it if it \emph{fixes a violation} ($M<0\!\to\!M'\ge 0$) or \emph{strictly increases a safe margin} ($M\ge 0$ and $M'>M$); otherwise keep the incumbent. Because a replacement never increases the indicator of violation for the replaced slot and may strictly improve margins, the empirical $\varepsilon_{\mathrm{ANY}}$ cannot increase upon accepted moves; otherwise the current safe margin is monotonically improved.

\medskip
\noindent\textbf{Limitations.} This is a heuristic filter with \emph{no} DP accounting; it only regulates a distance-based proxy. Under very tight thresholds it can inflate numeric tails or reweight rare categorical levels (“categorical drift”) as mass is pushed out of dense regions. Results may be sensitive to the weighting scheme and to the mixed-type distance chosen.

\begin{algorithm}[t]
\scriptsize
\DontPrintSemicolon
\caption{Weighted HEOM--$k$NN $\varepsilon_{\mathrm{ANY}}$ Rejection-with-Replacement (Ablation)}
\label{alg:weighted-heom-any-knn}
\KwIn{Real data $D_{\mathrm{real}}$ with numeric $\mathcal{N}$ and categorical $\mathcal{C}$; generator $G$ (\texttt{model.sample}); target \texttt{min\_eps}; sample size $n$; $k$-NN order $k$; optional per-column weights $w_c$.}
\KwOut{Synthetic dataset $S$ with $\varepsilon_{\mathrm{ANY}}<\texttt{min\_eps}$ (if feasible).}

\SetKwFunction{Encode}{EncodeAndWeight}
\SetKwFunction{Radii}{RadiiFromReal}
\SetKwFunction{Margin}{AnyMargin}
\SetKwProg{Fn}{Function}{}{}

\BlankLine
\Fn{\Encode{$D_{\mathrm{real}},\mathcal{N},\mathcal{C}$}}{
\textbf{HEOM encoder.} Fit min--max on $\mathcal{N}$; one--hot $\mathcal{C}$ with scale $1/\sqrt{2}$ to obtain encoder $\Enc: D\mapsto X\in\mathbb{R}^d$.\;
$X_r\leftarrow \Enc(D_{\mathrm{real}})$.\;
\textbf{Column weights.} Set $w_c\leftarrow 1/(H_c+\varepsilon)$.\;
\textbf{Expand \& scale.} Expand to $w_{\mathrm{dim}}\in\mathbb{R}^d$ (all dummies of a categorical share its $w_c$).\;
$w_{\mathrm{scale}}\leftarrow \sqrt{w_{\mathrm{dim}}}$; define $\lVert u-v\rVert_{w} := \lVert (u-v)\odot w_{\mathrm{scale}}\rVert_{2}$.\;
\Return{$\bigl(\Enc,\,w_{\mathrm{scale}},\,X_r^{w}{=}\,X_r\odot w_{\mathrm{scale}}\bigr)$}\;
}

\BlankLine
\Fn{\Radii{$X_r^{w},k$}}{
Build a Euclidean $k$NN index $\mathsf{NN}$ on $X_r^{w}$; $n_r\leftarrow |X_r^{w}|$; $k_{\mathrm{eff}}\leftarrow \min(\max(k,2),n_r)$.\;
For each $x_i\in X_r^{w}$, query the $k_{\mathrm{eff}}$-NN distance $r_i$ (the $k_{\mathrm{eff}}$-th).\;
$R_{\max}\leftarrow \max_i r_i$.\;
\Return{$\bigl(\{r_i\},\,R_{\max},\,\mathsf{NN},\,n_r\bigr)$}\;
}

\BlankLine
\Fn{\Margin{$x^{w},\mathsf{NN},R_{\max},\{r_i\},n_r$}}{
$m\leftarrow \min(64,n_r)$; $b\leftarrow +\infty$.\;
\While{True}{
Query $m$-NN of $x^{w}$ from $\mathsf{NN}$: distances $d_{1:m}$ and indices $i_{1:m}$ (sorted).\;
\lIf{$\exists j:\ d_j\le R_{\max}$}{$b\leftarrow \min\!\bigl(b,\ \min_{j:\, d_j\le R_{\max}}(d_j^2-r_{i_j}^2)\bigr)$}
\lElse{$b\leftarrow \min\!\bigl(b,\ d_1^2-R_{\max}^2\bigr)$ \tcp*[f]{fallback if no neighbor within $R_{\max}$}}
\lIf{$d_m\ge R_{\max}$ \textbf{or} $b\le d_m^2-R_{\max}^2$ \textbf{or} $m=n_r$}{\Return{$b$}}
$m\leftarrow \min(n_r,2m)$ \tcp*[f]{progressively widen the search}
}
}

\BlankLine
\textbf{Initialize.}
$(\Enc,w_{\mathrm{scale}},X_r^{w})\leftarrow \Encode(D_{\mathrm{real}},\mathcal{N},\mathcal{C})$.\;
$\bigl(\{r_i\},R_{\max},\mathsf{NN},n_r\bigr)\leftarrow \Radii(X_r^{w},k)$.\;
Draw $n$ samples $S\sim G$; $X_s^{w}\leftarrow \Enc(S)\odot w_{\mathrm{scale}}$.\;
For each $x^{w}\in X_s^{w}$, compute $M(x^{w})\leftarrow \Margin(x^{w},\mathsf{NN},R_{\max},\{r_i\},n_r)$.\;
$\displaystyle \varepsilon_{\mathrm{ANY}}\leftarrow \frac{1}{n}\sum_{x^{w}\in X_s^{w}}\mathbf{1}\!\left[M(x^{w})<0\right]$.\;

\BlankLine
\textbf{Iterate.}
\While{$\varepsilon_{\mathrm{ANY}}\ge \texttt{min\_eps}$}{
$j^\star\leftarrow \arg\min_j M\bigl(X_s^{w}[j]\bigr)$ \tcp*[f]{current worst}\;
Draw $s'\sim G$; $x'^{w}\leftarrow \Enc(s')\odot w_{\mathrm{scale}}$; $M(x'^{w})\leftarrow \Margin(x'^{w},\mathsf{NN},R_{\max},\{r_i\},n_r)$.\;
\If(\tcp*[f]{accept if it fixes a violation or improves a safe margin}){$\bigl(M(X_s^{w}[j^\star])<0\ \wedge\ M(x'^{w})\ge 0\bigr)\ \lor\ \bigl(M(X_s^{w}[j^\star])\ge 0\ \wedge\ M(x'^{w})>M(X_s^{w}[j^\star])\bigr)$}{
$S[j^\star]\leftarrow s'$; $X_s^{w}[j^\star]\leftarrow x'^{w}$; $M(X_s^{w}[j^\star])\leftarrow M(x'^{w})$.\;
$\displaystyle \varepsilon_{\mathrm{ANY}}\leftarrow \frac{1}{n}\sum_{x^{w}\in X_s^{w}}\mathbf{1}\!\left[M(x^{w})<0\right]$.\;
}
}
\Return{$S$}\;

\end{algorithm}

\begin{table}[t]
\centering
\small
\caption{AIA (Classification) on \textit{Credit} with \textbf{CTGAN} at fixed $\varepsilon_{\mathrm{ANY}}=0.01$. Scores averaged over sensitive attributes. $\Delta$ vs \textit{No-filter}.}
\label{tab:aia_cls_credit_CTGAN}
\begin{tabular}{lrr@{\,}lr@{\,}l}
\toprule
Setting & Accuracy & Weighted F1 & $\Delta$ Accuracy & $\Delta$ Weighted F1 \\
\midrule
Real & 0.564 & 0.439 & -- & -- \\
No-filter & 0.525 & 0.441 & +0.000 & +0.000 \\
V0 & 0.549 & 0.434 & +0.024 & -0.007 \\
V1 & 0.546 & 0.436 & +0.021 & -0.005 \\
V2 & 0.550 & 0.425 & +0.025 & -0.016 \\
\bottomrule
\end{tabular}
\end{table} 

\begin{table}[t]
\centering
\small
\caption{AIA (Classification) on \textit{Credit} with \textbf{TVAE} at fixed $\varepsilon_{\mathrm{ANY}}=0.01$. Scores averaged over sensitive attributes. $\Delta$ vs \textit{No-filter}.}
\label{tab:aia_cls_credit_TVAE}
\begin{tabular}{lrr@{\,}lr@{\,}l}
\toprule
Setting & Accuracy & Weighted F1 & $\Delta$ Accuracy & $\Delta$ Weighted F1 \\
\midrule
Real & 0.564 & 0.439 & -- & -- \\
No-filter & 0.480 & 0.391 & +0.000 & +0.000 \\
V0 & 0.492 & 0.414 & +0.012 & +0.023 \\
V1 & 0.439 & 0.375 & -0.041 & -0.017 \\
V2 & 0.514 & 0.424 & +0.034 & +0.032 \\
\bottomrule
\end{tabular}
\end{table} 

\begin{table}[t]
\centering
\small
\caption{AIA (Classification) on \textit{Adult} with \textbf{CTGAN} at fixed $\varepsilon_{\mathrm{ANY}}=0.01$. Scores averaged over sensitive attributes. $\Delta$ vs \textit{No-filter}.}
\label{tab:aia_cls_adult_CTGAN}
\begin{tabular}{lrr@{\,}lr@{\,}l}
\toprule
Setting & Accuracy & Weighted F1 & $\Delta$ Accuracy & $\Delta$ Weighted F1 \\
\midrule
Real & 0.845 & 0.826 & -- & -- \\
No-filter & 0.817 & 0.809 & +0.000 & +0.000 \\
V0 & 0.809 & 0.807 & -0.007 & -0.002 \\
V1 & 0.805 & 0.802 & -0.012 & -0.007 \\
V2 & 0.821 & 0.811 & +0.004 & +0.002 \\
\bottomrule
\end{tabular}
\end{table} 

\begin{table}[t]
\centering
\small
\caption{AIA (Classification) on \textit{Adult} with \textbf{TVAE} at fixed $\varepsilon_{\mathrm{ANY}}=0.01$. Scores averaged over sensitive attributes. $\Delta$ vs \textit{No-filter}.}
\label{tab:aia_cls_adult_TVAE}
\begin{tabular}{lrr@{\,}lr@{\,}l}
\toprule
Setting & Accuracy & Weighted F1 & $\Delta$ Accuracy & $\Delta$ Weighted F1 \\
\midrule
Real & 0.845 & 0.826 & -- & -- \\
No-filter & 0.842 & 0.821 & +0.000 & +0.000 \\
V0 & 0.819 & 0.808 & -0.022 & -0.013 \\
V1 & 0.831 & 0.815 & -0.011 & -0.006 \\
V2 & 0.833 & 0.816 & -0.008 & -0.005 \\
\bottomrule
\end{tabular}
\end{table} 

\begin{table}[t]
\centering
\small
\caption{AIA (Classification) on \textit{Cardio} with \textbf{CTGAN} at fixed $\varepsilon_{\mathrm{ANY}}=0.01$. Scores averaged over sensitive attributes. $\Delta$ vs \textit{No-filter}.}
\label{tab:aia_cls_cardio_CTGAN}
\begin{tabular}{lrr@{\,}lr@{\,}l}
\toprule
Setting & Accuracy & Weighted F1 & $\Delta$ Accuracy & $\Delta$ Weighted F1 \\
\midrule
Real & 0.730 & 0.733 & -- & -- \\
No-filter & 0.732 & 0.730 & +0.000 & +0.000 \\
V0 & 0.696 & 0.714 & -0.035 & -0.016 \\
V1 & 0.702 & 0.717 & -0.030 & -0.013 \\
V2 & 0.688 & 0.702 & -0.044 & -0.028 \\
\bottomrule
\end{tabular}
\end{table} 

\begin{table}[t]
\centering
\small
\caption{AIA (Classification) on \textit{Cardio} with \textbf{TVAE} at fixed $\varepsilon_{\mathrm{ANY}}=0.01$. Scores averaged over sensitive attributes. $\Delta$ vs \textit{No-filter}.}
\label{tab:aia_cls_cardio_TVAE}
\begin{tabular}{lrr@{\,}lr@{\,}l}
\toprule
Setting & Accuracy & Weighted F1 & $\Delta$ Accuracy & $\Delta$ Weighted F1 \\
\midrule
Real & 0.730 & 0.733 & -- & -- \\
No-filter & 0.755 & 0.738 & +0.000 & +0.000 \\
V0 & 0.720 & 0.723 & -0.036 & -0.015 \\
V1 & 0.723 & 0.724 & -0.033 & -0.014 \\
V2 & 0.718 & 0.717 & -0.037 & -0.021 \\
\bottomrule
\end{tabular}
\end{table}

\begin{table}[t]
\centering
\small
\caption{AIA (Regression) on \textit{Credit} with \textbf{CTGAN} at fixed $\varepsilon_{\mathrm{ANY}}=0.01$. Values averaged over targets (if multiple). $\Delta$ vs \textit{No-filter}.}
\label{tab:aia_reg_credit_CTGAN}
\begin{tabular}{lrr@{\,}lr@{\,}l}
\toprule
Setting & $R^2$ & RMSE & $\Delta$ $R^2$ & $\Delta$ RMSE \\
\midrule
Real & 0.022 & 46964.032 & -- & -- \\
No-filter & 0.008 & 47548.453 & +0.000 & +0.000 \\
V0 & 0.000 & 47860.076 & -0.008 & +311.623 \\
V1 & 0.002 & 47811.231 & -0.006 & +262.777 \\
V2 & -0.043 & 49231.207 & -0.051 & +1682.754 \\
\bottomrule
\end{tabular}
\end{table} 

\begin{table}[t]
\centering
\small
\caption{AIA (Regression) on \textit{Credit} with \textbf{TVAE} at fixed $\varepsilon_{\mathrm{ANY}}=0.01$. Values averaged over targets (if multiple). $\Delta$ vs \textit{No-filter}.}
\label{tab:aia_reg_credit_TVAE}
\begin{tabular}{lrr@{\,}lr@{\,}l}
\toprule
Setting & $R^2$ & RMSE & $\Delta$ $R^2$ & $\Delta$ RMSE \\
\midrule
Real & 0.022 & 46964.032 & -- & -- \\
No-filter & -0.002 & 47887.983 & +0.000 & +0.000 \\
V0 & -0.013 & 47950.474 & -0.011 & +62.491 \\
V1 & -0.012 & 47918.299 & -0.010 & +30.316 \\
V2 & -0.058 & 48926.967 & -0.057 & +1038.984 \\
\bottomrule
\end{tabular}
\end{table} 

\begin{table}[t]
\centering
\small
\caption{AIA (Regression) on \textit{Adult} with \textbf{CTGAN} at fixed $\varepsilon_{\mathrm{ANY}}=0.01$. Values averaged over targets (if multiple). $\Delta$ vs \textit{No-filter}.}
\label{tab:aia_reg_adult_CTGAN}
\begin{tabular}{lrr@{\,}lr@{\,}l}
\toprule
Setting & $R^2$ & RMSE & $\Delta$ $R^2$ & $\Delta$ RMSE \\
\midrule
Real & 0.088 & 2594.217 & -- & -- \\
No-filter & 0.051 & 2647.314 & +0.000 & +0.000 \\
V0 & 0.030 & 2648.363 & -0.020 & +1.049 \\
V1 & 0.053 & 2642.525 & +0.002 & -4.788 \\
V2 & 0.056 & 2642.457 & +0.005 & -4.856 \\
\bottomrule
\end{tabular}
\end{table} 

\begin{table}[t]
\centering
\small
\caption{AIA (Regression) on \textit{Adult} with \textbf{TVAE} at fixed $\varepsilon_{\mathrm{ANY}}=0.01$. Values averaged over targets (if multiple). $\Delta$ vs \textit{No-filter}.}
\label{tab:aia_reg_adult_TVAE}
\begin{tabular}{lrr@{\,}lr@{\,}l}
\toprule
Setting & $R^2$ & RMSE & $\Delta$ $R^2$ & $\Delta$ RMSE \\
\midrule
Real & 0.088 & 2594.217 & -- & -- \\
No-filter & 0.037 & 2674.862 & +0.000 & +0.000 \\
V0 & 0.012 & 2672.861 & -0.025 & -2.001 \\
V1 & 0.038 & 2668.798 & +0.002 & -6.064 \\
V2 & 0.034 & 2668.404 & -0.002 & -6.458 \\
\bottomrule
\end{tabular}
\end{table} 

\begin{table}[t]
\centering
\small
\caption{AIA (Regression) on \textit{Cardio} with \textbf{CTGAN} at fixed $\varepsilon_{\mathrm{ANY}}=0.01$. Values averaged over targets (if multiple). $\Delta$ vs \textit{No-filter}.}
\label{tab:aia_reg_cardio_CTGAN}
\begin{tabular}{lrr@{\,}lr@{\,}l}
\toprule
Setting & $R^2$ & RMSE & $\Delta$ $R^2$ & $\Delta$ RMSE \\
\midrule
Real & 0.001 & 157.308 & -- & -- \\
No-filter & 0.001 & 157.327 & +0.000 & +0.000 \\
V0 & -0.000 & 157.434 & -0.001 & +0.107 \\
V1 & 0.000 & 157.380 & -0.001 & +0.054 \\
V2 & -1.206 & 229.904 & -1.207 & +72.577 \\
\bottomrule
\end{tabular}
\end{table} 

\begin{table}[t]
\centering
\small
\caption{AIA (Regression) on \textit{Cardio} with \textbf{TVAE} at fixed $\varepsilon_{\mathrm{ANY}}=0.01$. Values averaged over targets (if multiple). $\Delta$ vs \textit{No-filter}.}
\label{tab:aia_reg_cardio_TVAE}
\begin{tabular}{lrr@{\,}lr@{\,}l}
\toprule
Setting & $R^2$ & RMSE & $\Delta$ $R^2$ & $\Delta$ RMSE \\
\midrule
Real & 0.001 & 157.308 & -- & -- \\
No-filter & 0.000 & 157.378 & +0.000 & +0.000 \\
V0 & 0.001 & 157.324 & +0.001 & -0.054 \\
V1 & 0.001 & 157.346 & +0.000 & -0.032 \\
V2 & -0.352 & 183.549 & -0.352 & +26.171 \\
\bottomrule
\end{tabular}
\end{table}

\end{document}